\documentclass[5p,times,twocolumn]{elsarticle}

\usepackage{ecrc}

\volume{}

\firstpage{1}

\journalname{}

\runauth{}

\jid{procs}

\jnltitlelogo{}

\usepackage{amssymb}
\usepackage{color}
\usepackage{times}
\usepackage{epsfig}
\usepackage{amsmath}
\usepackage{amssymb}
\usepackage{multirow}
\usepackage{multicol}
\usepackage{diagbox}
\usepackage{algorithm}
\usepackage{algorithmic}
\usepackage{booktabs}
\usepackage{caption}
\usepackage{makecell}
\usepackage{CJKutf8}
\usepackage{color}
\usepackage{float}
\usepackage{graphicx}
\usepackage{stfloats}
\usepackage{lipsum}
\usepackage{lineno}
\usepackage{fnlineno}

\usepackage[figuresright]{rotating}

\begin{document}
	\begin{frontmatter}
		
		
		\dochead{}
		
		\title{Unpaired Quad-Path Cycle Consistent Adversarial Networks for Single Image Defogging}


		\author[label1,label2]{Wei Liu\corref{cor1}}
		\ead{josephson870921@gmail.com}
		\author[label1,label2]{Cheng Chen}
		\author[label1,label2]{Rui Jiang}
		\author[label1,label2]{Tao Lu}
		\ead{lutxyl@gmail.com}
		\author[label3]{Zixiang Xiong}
		\ead{zx@ece.tamu.edu}
		
		\address[label1]{ School of Computer Science and Engineering, Wuhan Institute of Technology, Wuhan 430205, China}
		\address[label2]{Hubei Key Laboratory of Intelligent Robot, Wuhan Institute of Technology, Wuhan 430205, China}
		\address[label3]{Department of Electrical and Computer Engineering, Texas A\&M University, College Station, TX 77843 USA}
		
		\cortext[cor1]{Corresponding author.}

		\begin{abstract}
			Adversarial learning-based image defogging methods have been extensively studied in computer vision due to their remarkable performance. However, most existing methods have limited defogging capabilities for real cases because they are trained on the paired clear and synthesized foggy images of the same scenes. In addition, they have limitations in preserving vivid color and rich textual details in defogging. To address these issues, we develop a novel generative adversarial network, called quad-path cycle consistent adversarial network (QPC-Net), for single image defogging. QPC-Net consists of a Fog2Fogfree block and a Fogfree2Fog block. In each block, there are three learning-based modules, namely, fog removal, color-texture recovery, and fog synthetic, which sequentially compose dual-path that constrain each other to generate high quality images. Specifically, the color-texture recovery model is designed to exploit the self-similarity of texture and structure information by learning the holistic channel-spatial feature correlations between the foggy image with its several derived images. Moreover, in the fog synthetic module, we utilize the atmospheric scattering model to guide it to improve the generative quality by focusing on an atmospheric light optimization with a novel sky segmentation network. Extensive experiments on both synthetic and real-world datasets show that QPC-Net outperforms state-of-the-art defogging methods in terms of quantitative accuracy and subjective visual quality.
		\end{abstract}
		
		\begin{keyword}
			Image defogging, quad-path cycle consistent adversarial,  Fog2Fogfree block, sky segmentation network.
		\end{keyword}

	\end{frontmatter}
\section{Introduction}
\label{introduction}
In outdoor natural scenes, the captured digital images often experience quality degradation due to bad weather conditions, such as fog, smoke and haze. The resulting foggy images suffer from low contrast, distorted colors, and severe texture information loss, which adversely impact most existing computer vision applications such as image alignment \cite{zhou2020alignment}, object tracking \cite{shen2019visual}, and remote sensing \cite{han2014object}. It is thus necessary to design effective single image defogging algorithms to restore the content, color, and texture details from the foggy images. Existing image defogging methods can be roughly divided into three classes: prior-based methods \cite{He2011PAMI}, \cite{hazingline2016CVPR}, \cite{zhu2015TIPCAP}, \cite{IDE2021ide}, \cite{huedisparity2010ACCV}, fusion-based methods \cite{fusiondehazing2013TIP}, \cite{galdran2018image}, \cite{gao2020fusion} \cite{son2017near}, and learning-based methods \cite{dehazenet2016TIP}, \cite{GFN2018CVPR}, \cite{cycle-dehazing}, \cite{zhu2019tcyb}, \cite{zhang2021dual}. Most prior-based methods remove the fog from an image by using or improving an atmospheric scattering model (ASM) \cite{ASM}. Although these methods work well for some scenes, they often fail to handle other scenes that do not meet the predetermined prior assumptions. For instance, a dark-channel prior does not work well for restoring the sky or white building since there is no dark channel in these areas. Traditional fusion-based defogging methods usually need to combine several derived inputs from the original foggy image to fuse a fine-weight map to remove the fog. The limitation of these methods is that the derived inputs fail to reflect the inherent correlation between the depth information and the foggy image, leading to poor defogging performance when the fog density is high.

In recent years, many learning-based defogging methods have been proposed to address the disadvantages of prior- and fusion-based methods by leveraging the powerful feature extraction ability and spatial mapping capacities of convolutional neural networks (CNN). These methods can be grouped into two categories: paired and unpaired defogging methods: for former requires pairs of fog and fog-free images in network training, whereas the latter does not. In addition, paired defogging mainly includes ASM based defogging networks (ASMDN) \cite{dehazenet2016TIP} \cite{MSCNN2016ECCV} \cite{pang2018visual} \cite{zhu2019tcyb} \cite{AODNET2017ICCV} and end-to-end defogging networks (EDN) \cite{GFN2018CVPR} \cite{wang2018aipnet}  \cite{pix2pixdehazing} \cite{zhang2021dual}. In ASM, transmission and atmospheric light are two most important parameters. The accuracy of their estimation greatly affects the quality of the recovered image, with estimation inaccuracy resulting in undesired artifacts such as color distortion, halo, and over enhancement. ASMDN exploits CNN to estimate these two parameters before using ASM for defogging. However, ASMDN still suffers if transmission approximations are inaccurate due to the colors of the objects in the scene are inherently similar to the atmospheric lights. Thus, EDN was proposed to directly remove the fog form foggy images by using CNN without ASM. EDN has gradually become the mainstream algorithm for defrogging. Although ASMDN and EDN have rectified some shortcomings of traditional methods (\textit{e.g.,} prior-based ones), training their networks require a large number of fog-fogfree image pairs, which are difficult to obtain in practice. Therefore, for a clear image, ASM is usually adopted to synthesize the corresponding foggy image. This synthesis method is simple and efficient, but the distribution of the generated fog is relatively uniform, whereas the atmosphere is heterogeneous. CNN based models trained with this synthesized dataset are not best suited for the foggy images in real-world scenes.  A suitable learning-based defogging method not only needs to have good performance, but also fine texture details and high fidelity. 
	
\begin{figure*}[t]\scriptsize
	\vspace{-2.0mm}
	\begin{center}
		\begin{tabular}{@{}c@{}}
			\includegraphics[width = 0.72\textwidth,height=0.33\textheight]{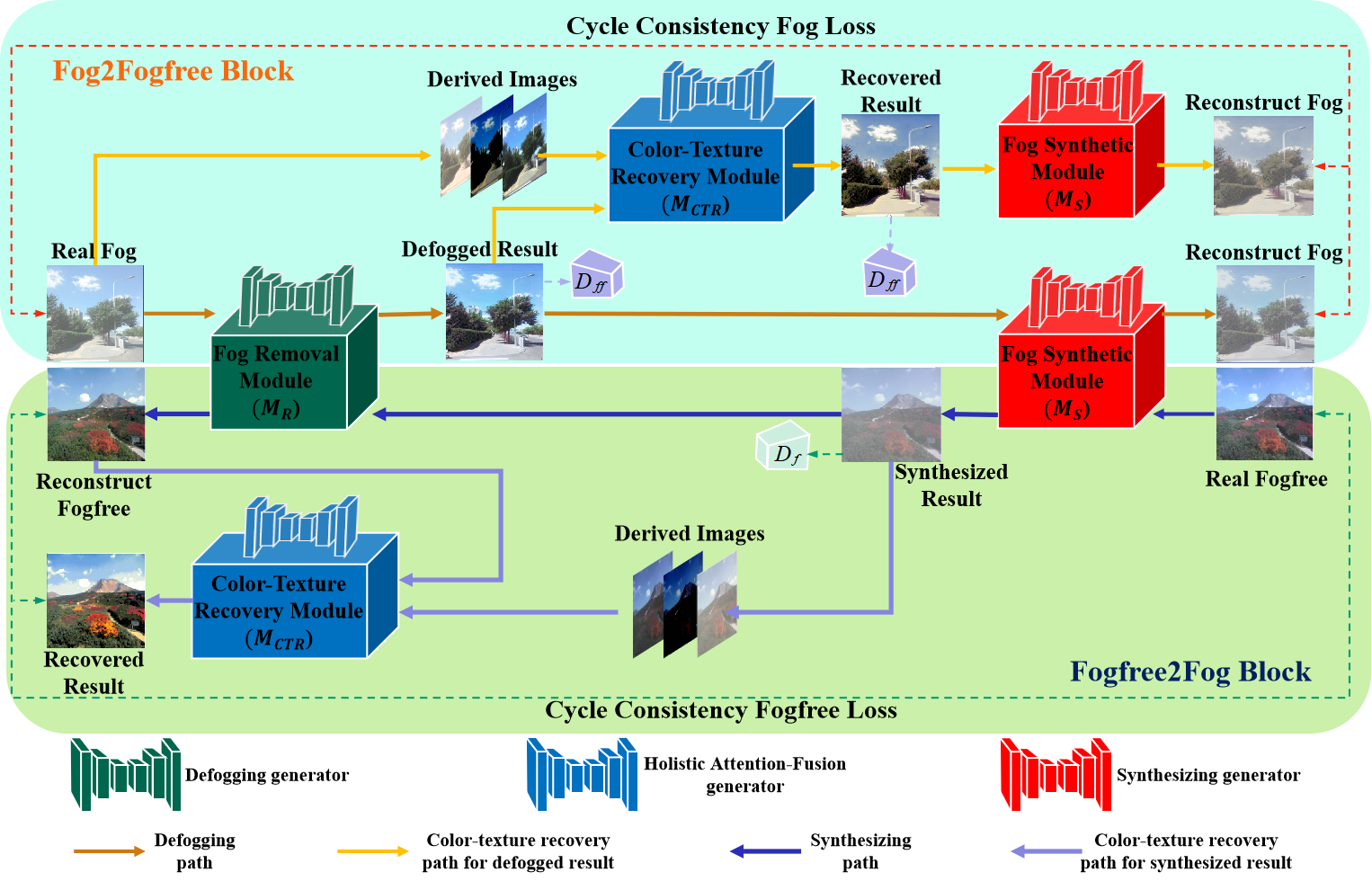}
		\end{tabular}
	\end{center}
	\vspace{-0.5cm}
	\caption{
		The overall architecture of the proposed QPC-Net.  It consists of two main components: 1) a Fog2Fogfree Block; 2) a Fogfree2Fog Block. Each block has a dual-path mapping mechanism. Inside the Fog2Fogfree block, the bottom sequential units constitute a defogging path to map foggy images to fogfree images, and the top sequential units constitute a color-texture recovery path for defogged result by fusion several derived inputs. Similarly, inside the Fogfree2Fog block, the top sequential units constitute a synthesizing path to map fogfree images to foggy images, and the bottom sequential units constitute a color-texture recovery path for synthesized results.
	}
	\vspace{-5.0mm}
	\label{architctures}
\end{figure*}

Unpaired defogging methods \cite{yang2018towards} \cite{cycle-dehazing} \cite{liu2020end} \cite{zhao2021refinednet} have been proposed to address the above issues because they are trained with unpaired real-world foggy and clear images. However, the defogged images generated through these methods usually have low contrast and quality. In this work, we propose a novel unpaired defogging method, called quad-path cycle consistent adversarial network (QPC-Net), for improved performance. QPC-Net shares a similar structure with CycleGAN that it also includes two transformation blocks that we call Fog2Fogfree and Fogfree2Fog blocks. In each block, there are three modules: a Fog Removal Module (FRM), a Fog Synthetic Module (FSM), and a Color-Texture Recovery Module (CTRM), as shown in Fig. \ref{architctures}, which constitute dual-path mappings. Inside the Fog2Fogfree block, a defogging path is constructed by the FRM$\rightarrow$FSM to map the image from the fog to fogfree domain, and a color-texture recovery path for defogged result is constructed by the CTRM$\rightarrow$FSM to improve image details. Similarly, inside the Fogfree2Fog block, the FSM\textbf{$\rightarrow$}FRM constitutes a synthesizing path to map the image from the fogfree to fog domain, the FRM$\rightarrow$CTRM forms a color-texture recovery path for synthesized result to further enhance image details. Specifically, in the CTRM, we devise a Holistic Attention-Fusion generator, which concatenates the fog image and its several derived features to exploit their correlations through a Channel Attention and Spatial Attention network to retrieve more contextual information (as shown in Fig. \ref{attention-structure}). This allows us to better capture the global information to constrain the defog network to generate result with more texture details. Furthermore, we utilize the ASM to guide the FSM to improve the generative quality by focusing on an atmospheric light optimization with a pre-train sky segmentation model. A large number of qualitative and quantitative comparisons against state-of-the-art defogging approaches have been performed to show that QPC-Net improves the defogged image quality and gives more image details.

The remainder of this paper is organized as follows. Section \ref{Related Work} gives a literature review; Section \ref{Proposed Model} covers QPC-Net; Section \ref{experiment} provides implementation details, experiment results, and qualitative and quantitative comparisons; and Section \ref{concluding} concludes the paper.
\section{Related Work}
\label{Related Work}
In this section, we only review those learning-based works, including paired and unpaired defogging methods, that are closely related to ours.  Before introducing them, it helps to start with the ASM to better understand the formation of foggy images to catch on the design ideas of related defogging methods.  

\subsection{Atmospheric Scattering Model}
Most existing prior- and learning-based defogging algorithms are based on the ASM \cite{McCartney}\cite{NarasimhanandNayar1}\cite{NarasimhanandNayar2}. It can be formulated as:
\begin{equation}
		I(x) = J(x)T(x)+A[1-T(x)]
		\label{adm}
\end{equation}
where  \textit{x} denotes the position of a pixel, \textit{I(x)} is the observed foggy image, \textit{J(x)} represents the fog-free image, \textit{T(x)} is the transmission map and \textit{A} is the atmospheric light. Unfortunately, in this ill-conditioned equation, both \textit{T(x)} and \textit{A} are unknown in practice. Thus, if we want to recover \textit{J} from \textit{I}, we need to first determine \textit{A} and \textit{T}. Most state-of-the-art methods either assume prior knowledge of them or use CNN to estimate them before attempting to recover the fog-free image \textit{J(x)} according (\ref{adm}). Moreover, in practice, \textit{T(x)} represents the medium transmission map, which is related to the depth of the scene via
\begin{equation}
	t(x) = e^{-\beta d(x)}
	\label{depth}
\end{equation}
where $\beta$ stands for the atmosphere scattering parameter and \textit{d} is the depth of the scene. 
\subsection{Paired defogging methods}
\label{SM}
As mentioned before, learning-based paired defogging methods can be mainly divided into two groups: ASMDN and EDN.

\textbf{Defogging via ASMDN.} These methods focus on estimating the transmission map \textit{T(x)} and atmospheric light \textit{A} through CNN before recovering the foggy image by using ASM. Cai \textit{et al.} \cite{dehazenet2016TIP} proposed a deep CNN based method to extract the multi-features from foggy image to optimize \textit{T(x)} with a nonlinear activation function. Similarly, a multiscale network was designed by Ren \textit{et al.} \cite{MSCNN2016ECCV}  to estimate \textit{T(x)} and \textit{A}. Pang \textit{et al.} \cite{pang2018visual} designed a unified network to jointly estimating \textit{T(x)}, \textit{A} and fog-free images. In \cite{zhu2019tcyb}, Zhu \textit{et al.} proposed a DehazeGAN, which combines a dense coarse-scale network with a fine-scale network to extract multiscale features from the foggy image, and they designed estimators for \textit{T(x)} and \textit{A} from these features. Different from most estimation methods, Li \textit{et al.} \cite{AODNET2017ICCV} reformulated (\ref{adm}) to combine \textit{T(x)} and \textit{A} as an integrated variable before proposing a light-weight CNN to generate the clean image with this new formulation. Compared with the traditional prior-based methods, ASMDN obtains better estimates of \textit{T(x)} and \textit{A}. However, artifacts will be present in defogged images when these two parameters are estimated incorrectly.   

\textbf{Defogging via EDN.} These methods directly remove the fog from foggy images to generate fog-free images through CNNs without using ASM. Based on the derived features of an input foggy image, Ren \textit{et al.} \cite{GFN2018CVPR} employed an end-to-end trainable gated fusion network for single image defogging that can effectively learn the inherent correlation between fog-related features and the transmission maps. Wang \textit{et al.} \cite{wang2018aipnet} proposed an atmospheric illumination prior in which the luminance channel in the YCrCb color space is mainly disturbed by the atmospheric light in foggy weather. Based on this prior, they adopt an end-to-end multiscale trainable model to restore the Y channel of a foggy image. In \cite{zhang2021dual}, the authors presented a Dual-Path Recurrent network, which has two parallel branches to recover the image by simultaneously learning the characteristics of the basic content and details of foggy images. Qu \textit{et al.} \cite{pix2pixdehazing} treated image defogging as an image-to-image translation problem and proposed an enhanced pix2pix defogging network. This method constructs a multi-resolution generative adversarial network and an enhancer to translate a foggy image to a clear one.

\subsection{Unpaired defogging methods}
Although paired learning-based defogging methods have achieved remarkable results, they require fog/fogfree image pair to train the network. In practice, it is difficult to synthesize fog around distant objects because the depth information is hard to estimate. Moreover, it is a time-consuming and labor-intensive task to synthesize a large number of foggy images with different fog concentrations. Therefore, unpaired learning-based defogging methods are better suited for real-world defogging tasks. These methods not only learn the domain correlation between the foggy and clear images in the natural scenes, but also effectively reduce the cost of data preparation. Generative Adversarial Networks (GNA) \cite{Goodfellow2020} as a representative unpaired technique, which has been proved to be superior in the field of image generation and restoration \cite{Isola2017} \cite{zhu2017unpaired} \cite{Hu2019}. For defogging, Yang \textit{et al.} \cite{yang2018towards} first proposed a disentangled dehazing network consists of three generators and a multi-scale discriminator to produce defogged results from the foggy images. By introducing a perceptual loss function, Engin \textit{et al.} \cite{cycle-dehazing} employed an enhanced CycleGAN \cite{zhu2017unpaired} to directly generate defogging results. Inspired by \cite{zhu2017unpaired}, we proposed in one of our previous works \cite{liu2020end} a network with a two-stage transformation path based on Cyclegan to map foggy images to the fog-free domain; the network in \cite{liu2020end} consists of two generators, two enhancers, and a discriminator. In \cite{zhao2021refinednet}, Zhao proposed a two-stage RefineDNet to combine the merits of prior- and learning-based approaches to achieve both visibility restoration and realness improvement. The methods above typically tend to generate low-quality defogged images with loss of texture detail and low contrast. Our proposed method focuses on unpaired images for training, it uses quad-path mappings to improve the defogging generator performance.

\section{Proposed Model}
\label{Proposed Model}
\subsection{Overview} 
\label{overview}
As illustrated in Fig. \ref{architctures}, our proposed QPC-Net consists of a Fog2Fogfree block and a Fogfree2Fog block. In these two blocks,  there are three modules, including Fog Removal Module (FRM, $M_R$), Fog Synthetic Module (FSM, $M_S$) and Color-Texture Recovery Module (CTRM, $M_{CTR}$). As can be seen from Fig. \ref{architctures}, different modules are sequentially combined to form quad-path mappings, which are defogging path (composed of $M_R$ and $M_S$), synthesizing path (composed of $M_S$ and $M_R$), color-texture recovery path for defogged result (composed of $M_{CTR}$ and $M_S$) and color-texture recovery path for synthesized result (composed of $M_R$ and $M_{CTR}$). Therefore, our method forms a game learning process of confrontation by these quad paths to achieve the mapping transformation from the foggy images to the fog-free images domain. Moreover, the functions of $M_R$, $M_S$ and $M_{CTR}$ are mainly completed by our proposed Defogging generator, Synthesizing generator and Holistic Attention-Fusion generator respectively. 

\textit{For Fog2Fogfree block}, its dual-path consists of a defogging path and a color-texture recovery path for defogged result. Given an input real-world foggy image $I_{rf}$, we first obtain the defogged image $I_{df}^{rf}$ by 
\begin{equation}
	\begin{aligned}
		I_{df}^{rf}=M_{R}(I_{rf})
	\end{aligned}
\end{equation}

Then, $I_{df}^{rf}$ is fed to fog synthetic module to reconstruct the first foggy image $I_{rcf}^{1}$,
\begin{equation}
	\begin{aligned}
		I_{rcf}^{1}=M_S(M_{R}(I_{rf}))=M_S(	I_{df}^{rf})
	\end{aligned}
\end{equation}

The above process we call it defogging path. To further improve the performance of $M_R$, the color-texture recovery path for defogged result is proposed. This path first takes the defogged image $I_{df}^{rf}$ and several derived images of the $I_{rf}$ as the input, and generated a recovered result $I_{rr}^{rf}$,
\begin{equation}
	I_{rr}^{rf}=M_{CTR}(I_{df}^{rf},I_{d}^{rf})
\end{equation}
where $I_{d}^{rf}$ denotes the derived images of $I_{rf}$. In this paper, we select three derived images for $I_{rf}$. Then, we put this result again into fog synthetic module to reconstruct the second foggy image $I_{rcf}^{2}$ by
\begin{equation}
	I_{rcf}^{2}=M_S(M_{CTR}(I_{df}^{rf},I_{d}^{rf}))=M_S(I_{rr}^{rf})
\end{equation}

Finally, we can obtain two cycle consistency loss functions for foggy images among $I_{rf}$, $I_{rcf}^{1}$, and $I_{rcf}^{2}$.  Note that, the defogged result obtained by fog removal module is our goal. The color-texture recovery path for defogged result is to constrain the fog removal module to generate results with better color fidelity and texture details.

\textit{For Fogfree2Fog block}, there still has a dual-path, which consists of a synthesizing path and a color-texture recovery path for synthesized result. Given an input real-world fogfree image $I_{rff}$, the synthesizing path is first fed it to fog synthetic module to generate a synthesized result $I_{sf}$, and then use $I_{sf}$ as the input of the fog removal module to generate the reconstruct fogfree image $I_{rcff}$,
\begin{equation}
	\begin{cases}
		I_{sf} =  M_S(I_{rff}) \\
		I_{rcff} = M_R(I_{sf})
	\end{cases}
\end{equation}

For color-texture recovery path for synthesized result, we first take the reconstruct fogfree image $I_{rcff}$ and the derived images $I_{d}^{sf}$ produced by $I_{sf}$ as input. Then we recovery this result by using color-texture recovery module.
\begin{equation}
	I_{rr}^{sf} = M_{CTR}(I_{rcff},I_{d}^{sf})
\end{equation}
where $I_{rr}^{sf}$ is the recovered image for synthesized result $I_{sf}$. Similarly, two cycle consistency loss functions for fogfree images are constructed among $I_{rff}$, $I_{rcff}$, and $I_{rr}^{sf}$.

\subsection{Network Architecture} 
\label{Network Architecture}
As discussed in \ref{overview}, each module has a major generator to perform its corresponding functions (as shown in Fig. \ref{architctures}), which are Defogging generator for $M_R$, Synthesizing generator for $M_S$, and Holistic Attention-Fusion generator for $M_{CTR}$. Moreover, another major component of our QPC-Net is the discriminator.
\begin{figure}[H]\scriptsize
	\begin{center}
		\vspace{-0.2cm}
		\begin{tabular}{@{}c@{}}
			\includegraphics[width = 0.49\textwidth]{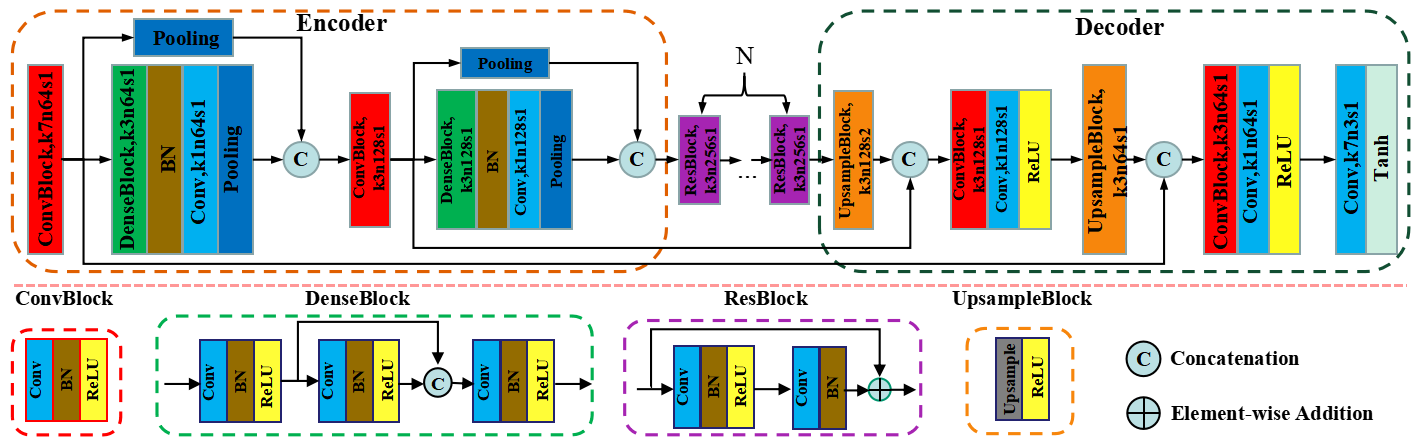}\\
		\end{tabular}
	\end{center}
	\vspace{-0.5cm}
	\caption{The architecture of the defogging generator.}
	\vspace{-4mm}
	\label{defog-structure}
\end{figure}

\textbf{\textit{Defogging generator:}} As can be seen from Fig. \ref{defog-structure}, we show the architecture details of the defogging generator. Inspired by \cite{dong2020fd}, we design a densely-residual connected encoder-decoder for removed the fog from the foggy image. Compared with the structure of densely connect alone, introducing the residually connect block in our architecture, the generator can not only better learn the correlation information from shallow layers to deep layers, but also can better preserve the image content and texture details for defogged results. As shown in Fig. \ref{defog-structure}, the encoder contains two convolution blocks(ConvBlocks) and two dense blocks (DenseBlocks), which including a series of convolutional, concatenation, batch normalization(BN), pooling layers and ReLU layers. After the encoder, six residual blocks (ResBlocks) are connected. For encoder, given a foggy image $I_{f}$, we first extract the shallow feature $\mathcal{F}_{C}^{1}\epsilon\mathbb{R}^{64\times H\times W}$ by the first convolution block:
\begin{equation}	
	\left\{
	\begin{aligned} 
		\mathcal{F}_{C}^{1} &= \mathcal{C}_{7\times 7}(I_{f}),\\
		\mathcal{C}_{k\times k}(.) &= \delta (\eta (C_{k\times k}(.)))
	\end{aligned}
	\right.
\end{equation}
where $\mathcal{C}_{k\times k}(.)$ denotes the ConvBlock operation, including convolution operation $C_{k\times k}$ where kernel size is $k\times k$, batch normalization function $\eta$, and rectified linear unit (ReLU) activation function $\delta$. Then, we fed $\mathcal{F}_{C}^{1}$ to the first DenseBlock to distill the first encoding feature $\mathcal{F}_{en}^{1}\epsilon\mathbb{R}^{128\times {H/2}\times {W/2}}$,
\begin{equation}
	\setlength\abovedisplayskip{6pt}
	\setlength\belowdisplayskip{6pt}
	\begin{split}
		\begin{matrix} 
			\mathcal{F}_{en}^{1}=[\mathcal{P}(\mathcal{F}_{C}^{1}),\mathcal{P}(C_{1\times 1}(\eta(\mathcal{D}_{3\times 3}(\mathcal{F}_{C}^{1}))))]
		\end{matrix}
	\end{split}
\end{equation}
where $\mathcal{D}_{k\times k}(.)$ is the DenseBlock operation, which has three ConvBlocks (as shown in Fig. \ref{defog-structure}). $[.]$ represents concatenation operation, $\mathcal{P}$ is the pooling operation. Finally, $\mathcal{F}_{en}^{1}$ is fed to the second ConvBlock and DenseBlock to obtain the final encoding feature $\mathcal{F}_{en}^{2}\epsilon\mathbb{R}^{256\times {H/4}\times {W/4}}$,
\begin{equation}
	\setlength\abovedisplayskip{6pt}
	\setlength\belowdisplayskip{6pt}
	\begin{split}
		\begin{matrix} 
			\mathcal{F}_{en}^{2}=[\mathcal{P}(\mathcal{F}_{C}^{2}),\mathcal{P}(C_{1\times 1}(\eta(\mathcal{D}_{3\times 3}(\mathcal{F}_{C}^{2}))))]
		\end{matrix}
	\end{split}
\end{equation}
where $	\mathcal{F}_{C}^{2}=\mathcal{C}_{3\times 3}(\mathcal{F}_{en}^{1})$. Moreover, we know that increasing the network depth can improve the representational ability of the network. However, to avoid the problem of gradient disappearance caused by increasing the network depth, we use the residual block to further refine the encoding feature $\mathcal{F}_{en}^{2}$. The residual features extracted from the ResBlock is expressed as follows:
\begin{equation}
	\begin{aligned} 
		\mathcal{F}_{re}^{i}= \mathcal{R}_{3\times 3}^{i}(\mathcal{F}_{re}^{i-1}), i \in [1,N] 
	\end{aligned}	
	\label{res}
\end{equation}
where $ \mathcal{F}_{re}^{0} =  \mathcal{F}_{en}^{2}$, $\mathcal{R}_{k\times k}^{i}(.)$ represents the i-th ResBlock operation, N is the number of the ResBlock. In this work, N=6. 

For decoder, as shown in Fig. \ref{defog-structure}, it contains two upsample blocks (UpsampleBlocks) and two convolution blocks(ConvBlocks). Inside decoder, to preserve the image content and recover the details from the input, we concate the shallow features. In our decoder, after each upsampling, its output is spliced with that of the convolution block in the encoding stage, and then the convolution block in the decoding stage is used to refine the splicing result to obtain the decoding feature. The first decoding feature $\mathcal{F}_{de}^{1}\epsilon\mathbb{R}^{128\times {H/2}\times {W/2}}$ is formed by
\begin{equation}
	\begin{split}
		\begin{matrix} 
			\mathcal{F}_{de}^{1}= \delta(C_{1 \times 1}(\mathcal{C}_{3\times3}([\mathcal{U}_{3\times 3}(\mathcal{F}_{re}^{N}),\mathcal{F}_{C}^{2}])))
		\end{matrix}
	\end{split}
\end{equation}
where $\mathcal{U}_{k\times k}(.)$ is the UpsampleBlocks operation where kernel size is  $k\times k$. Similarly, the second decoding feature $\mathcal{F}_{de}^{2}\epsilon\mathbb{R}^{64\times {H}\times {W}}$ is getted by
\begin{equation}
	\begin{aligned}
		\mathcal{F}_{de}^{2}= \delta(C_{1 \times 1}(\mathcal{C}_{3\times3}([\mathcal{U}_{3\times 3}(\mathcal{F}_{de}^{1}),\mathcal{F}_{C}^{1}])))
	\end{aligned}
\end{equation}

Then, through a convolution operation and a Tanh activation function, we obtain the defogged result $I_{df}\epsilon\mathbb{R}^{3\times {H}\times {W}}$.
\begin{equation}
	\begin{aligned}
		I_{df}= \tau(C_{7\times7}(\mathcal{F}_{de}^{2}))
	\end{aligned}
\end{equation}
where $\tau$ denotes the Tanh activation function.
\begin{figure}[H]\scriptsize
	\begin{center}
		\vspace{-0.2cm}
		\begin{tabular}{@{}c@{}}
			\includegraphics[width = 0.49\textwidth]{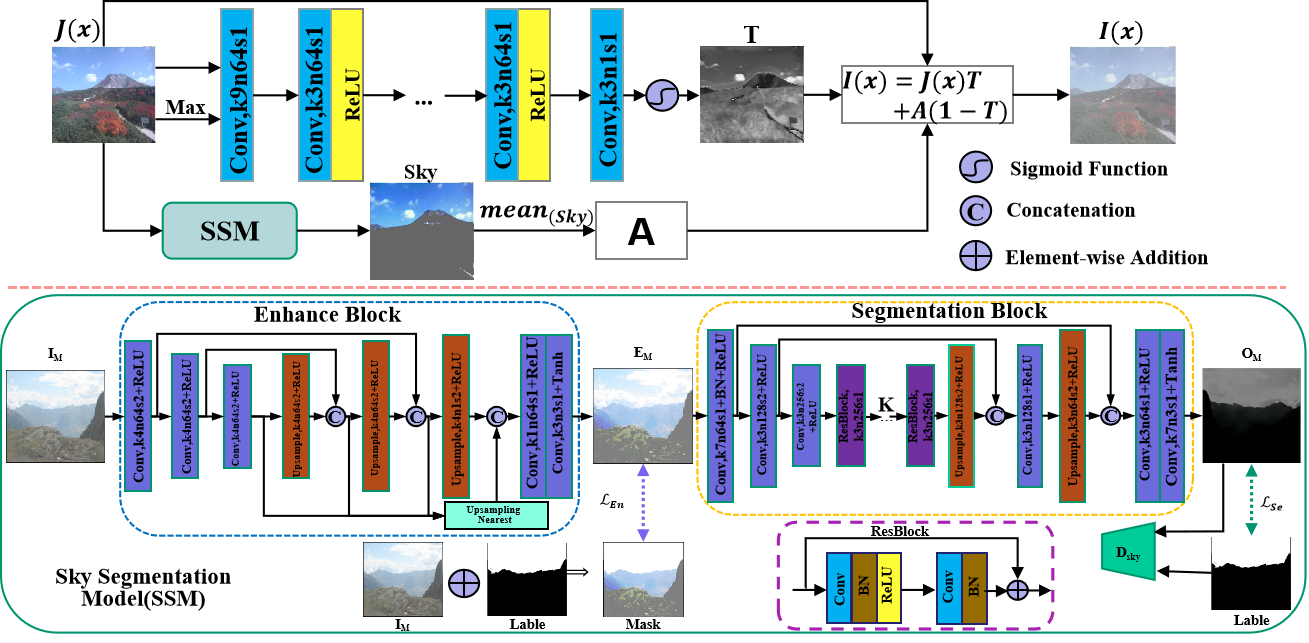}
		\end{tabular}
	\end{center}
	\vspace{-0.5cm}
	\caption{The architecture of the synthesizing generator.}
	\vspace{-3mm}
	\label{synthetic-structure}
\end{figure}

\textbf{\textit{Synthesizing generator:}} In practice, due to the diversity and complexity of fog distribution, it is very difficult to directly use CNN to synthesis the foggy image. Thus, as shown in Fig. \ref{synthetic-structure}, the ASM is embedded in our synthesizing generator. We design two CNN-based models to optimize the transmission $T$ and atmospheric light $A$ respectively. Especially for atmospheric light $A$, we propose a sky segmentation model (SSM, as shown in Fig. \ref{synthetic-structure}). To obtain finer  transmission $T$, we use 5 convolution layers. As shown in Fig. \ref{synthetic-structure}, we first take a clear image ($J$) and the maximum pixel value in its channel as the input, and then extract the roughly feature by using a convolution operation. After this operation, we use three convolutional neural networks with ReLU function to refine the rough feature. Finally, the transmission $T$ is generated by a convolution network with a sigmoid function. 
\begin{equation}
	\begin{aligned}
		T=\sigma(C_{3\times3}(\mathcal{M}(\mathcal{M}(\mathcal{M}(C_{9\times9}([J,Max(J)])))))
	\end{aligned}
\end{equation}
where $\mathcal{M}$ denotes a $3 \times 3$ convolution neural network with a ReLU function, $\sigma$ denotes the sigmoid function. $Max(.)$ means to take the maximum pixel value for a image. 
\begin{figure}[hb]\footnotesize
	\centering
	\begin{center}
		\vspace{-0.2cm}
		\begin{tabular}{@{}ccccc@{}}
			\rotatebox{90}{  \quad Foggy} & \hspace{-0.5cm} 
			\rotatebox{90}{ \quad images} & \hspace{-0.5cm} 
			\includegraphics[width = 0.15\textwidth]{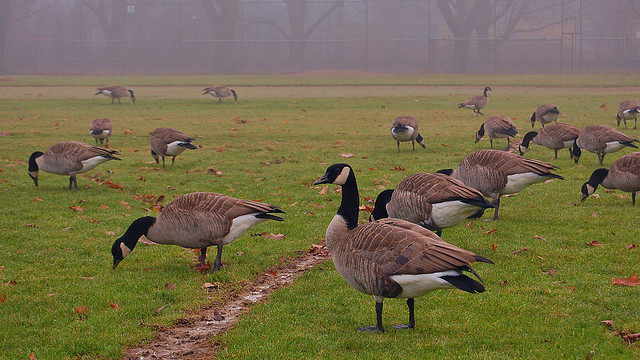} & \hspace{-0.5cm} 
			\includegraphics[width = 0.15\textwidth]{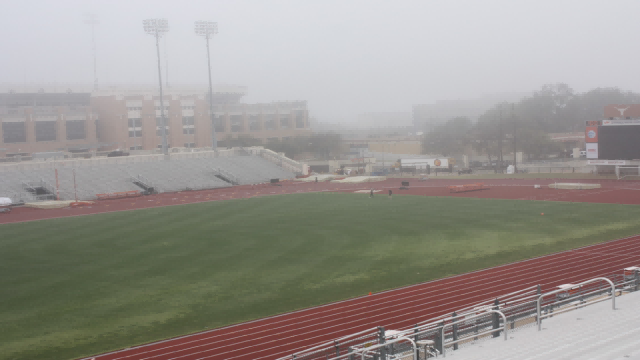} & \hspace{-0.5cm} 
			\includegraphics[width = 0.15\textwidth]{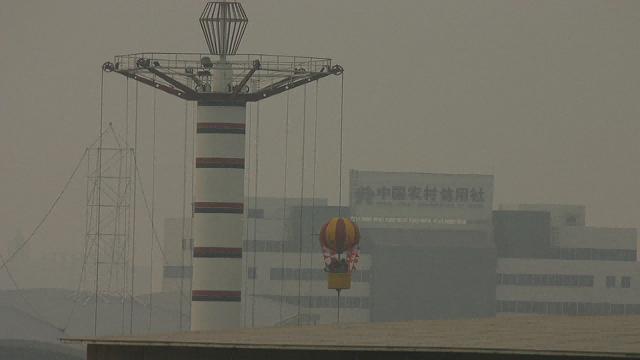} \\
			
			\rotatebox{90}{Transmission} & \hspace{-0.5cm} 
			\rotatebox{90}{ \quad maps} & \hspace{-0.5cm} 
			\includegraphics[width = 0.15\textwidth]{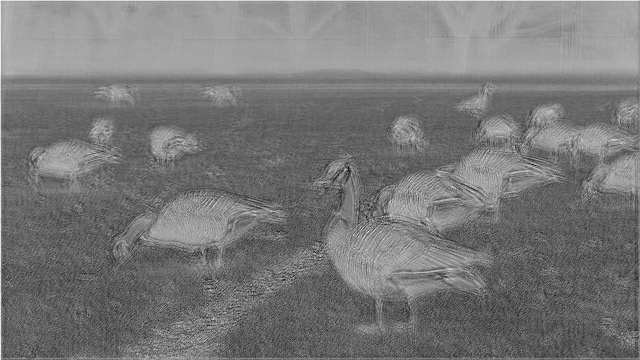} & \hspace{-0.5cm} 
			\includegraphics[width = 0.15\textwidth]{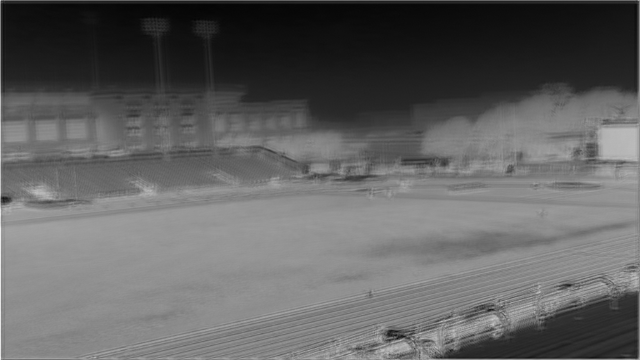} & \hspace{-0.5cm} 
			\includegraphics[width = 0.15\textwidth]{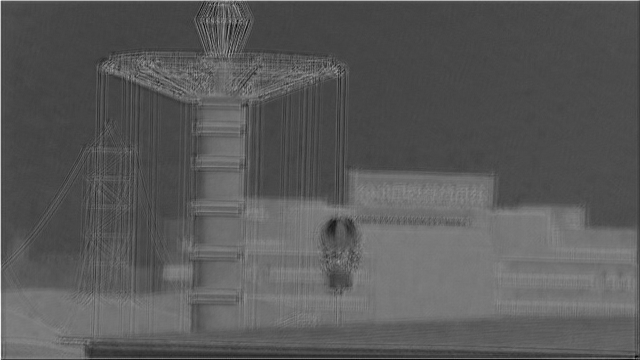} \\
			
			\rotatebox{90}{ Defogged} &\hspace{-0.6cm}
			\rotatebox{90}{\quad results}& \hspace{-0.5cm} 
			\includegraphics[width = 0.15\textwidth]{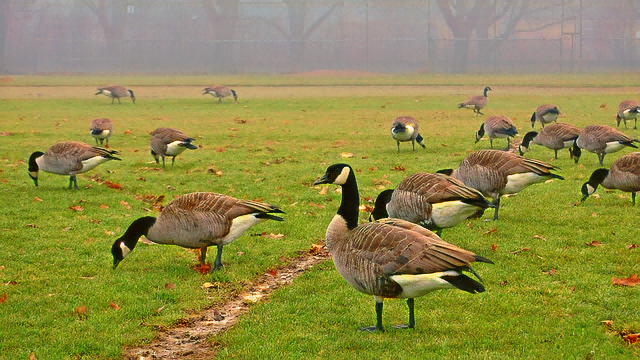} & \hspace{-0.5cm} 
			\includegraphics[width = 0.15\textwidth]{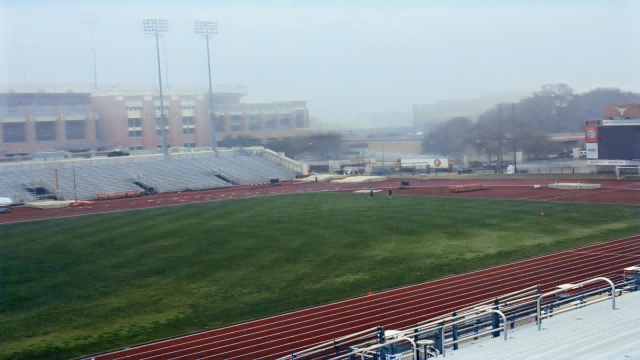} & \hspace{-0.5cm} 
			\includegraphics[width = 0.15\textwidth]{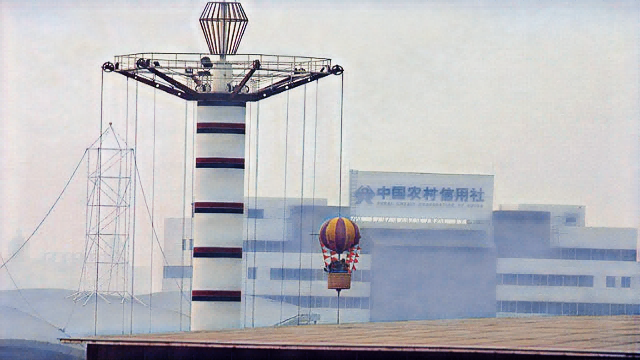}
		\end{tabular}
	\end{center}
	\vspace{-0.5cm}
	\caption{Defogged results and corresponding transmission maps through our QPC-Net on real-world outdoor foggy images.}
	\vspace{-0.3cm}
	\label{Transmission-map}
\end{figure}
Fig. \ref{Transmission-map} illustrates three transmission maps generated by our proposed network. It can be seen that each transmission map can effectively estimate the depth information of the scene and preserve richer texture details.

To optimize $A$, in our previous work \cite{liu2020end}, we propose a sky prior. For the foggy image with sky region, we assume that the depth of the sky is regarded as infinity, and define it as the intensity of pixels in the area of maximum fog density. Then, the average value of the sky region is described as $A$. For more details about this derivation process, please refer to our previous work  \cite{liu2020end}. Therefore, how to divide the sky region from the foggy image is crucial. Instead of using the existing technology \cite{he2010fast} \cite{closed2007TPAMI} \cite{otsu1979threshold} to segment the image to obtain the sky region, we propose a CNN-based sky segmentation model (SSM). 

As shown in Fig. \ref{synthetic-structure}, SSM consists of an Ehnace Block and a Segmentation Block. Each block adopts the encoder-decoder network. For an outdoor image, we observed that the boundary between sky area and non-sky area was often blurred. Thus, we first enhance the outdoor image to improve the edge information by using an enhance network, which can sharpen this boundary and label the sky region. Inside enhance block, we also use the dense connection to improve the texture details of the extracted features. For an input outdoor image $I_{M}$, the output of each layer in encoding stage is $E_{e}^{i}$, and $D_{e}^{j}$ is the output of each layer in decoding stage. The encoding process is defined as:
\begin{equation}
	\begin{aligned} 
		{E}_{e}^{i}= \delta(C_{4 \times 4}({E}_{e}^{i-1}))
	\end{aligned}	
\end{equation}
where ${E}_{e}^{0}=I_{M}$. The decoding process is defined as:
\begin{equation}
	\left\{\begin{aligned} 
		{D}_{e}^{1} = &\mathcal{U}_{4\times 4}({E}_{e}^{m}),\\
		{D}_{e}^{k+1} = &\mathcal{U}_{4\times 4}([{D}_{e}^{k},{E}_{e}^{m-k}]),k \in [1,n-1]\\
	\end{aligned}\right.
\end{equation}
where i and j are layer indices ($i=1,...,m$, $j=1,...,n$, $m$ and $n$ are the number of layers). In this work, $m=n=3$. Finally, the enhanced image $E_{M}$ is generated by concatenating all encoding and decoding features followed by two convolution layers.
\begin{figure}[h]\scriptsize
	\begin{center}
		\vspace{-0.2cm}
		\begin{tabular}{@{}ccc@{}}
			\includegraphics[width = 0.15\textwidth,height=0.08\textheight]{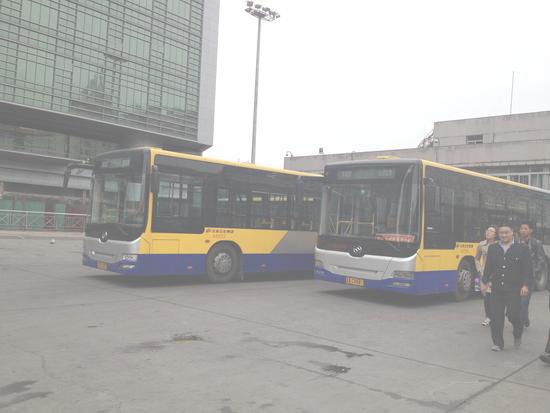} & \hspace{-0.4cm} 
			\includegraphics[width = 0.15\textwidth,height=0.08\textheight]{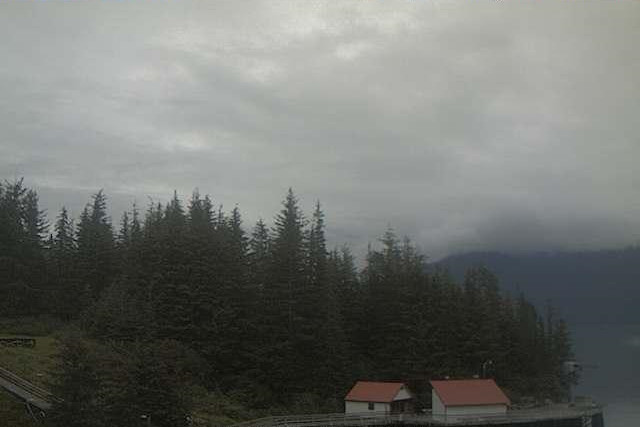} & \hspace{-0.4cm} 
			\includegraphics[width = 0.15\textwidth,height=0.08\textheight]{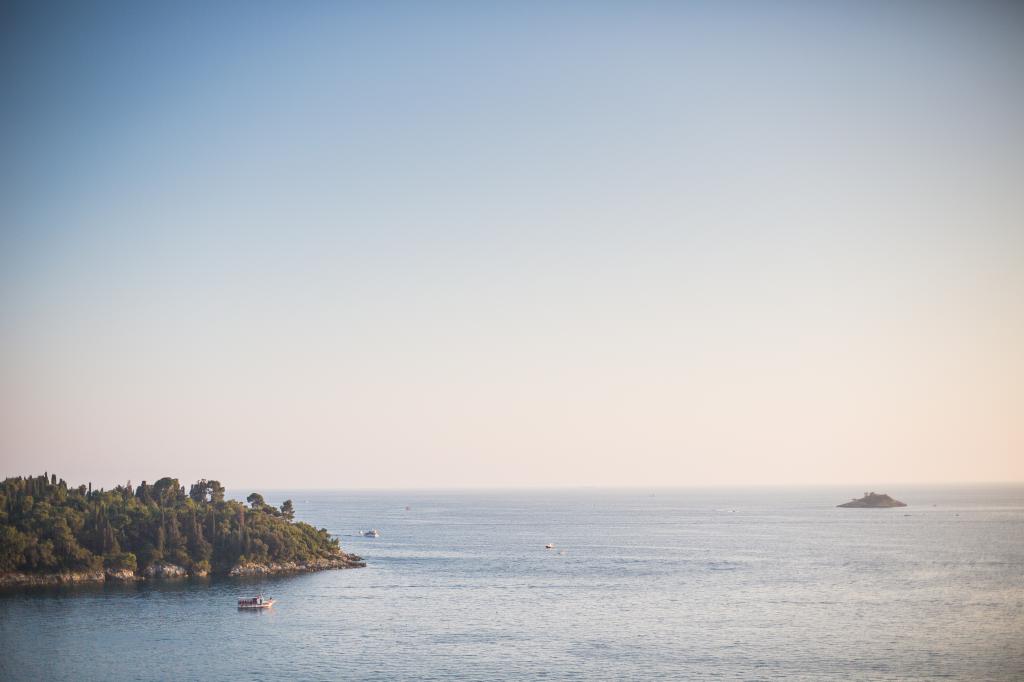} \\
			
			\includegraphics[width = 0.15\textwidth,height=0.08\textheight]{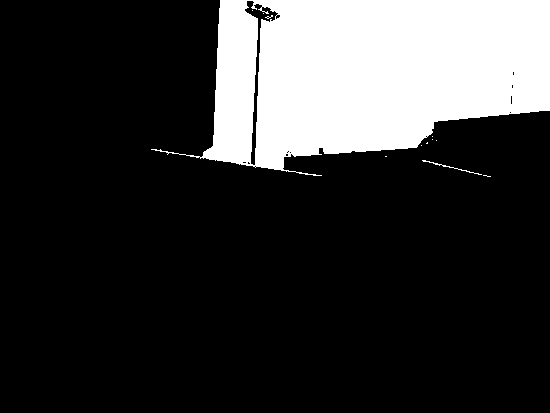} & \hspace{-0.4cm} 
			\includegraphics[width = 0.15\textwidth,height=0.08\textheight]{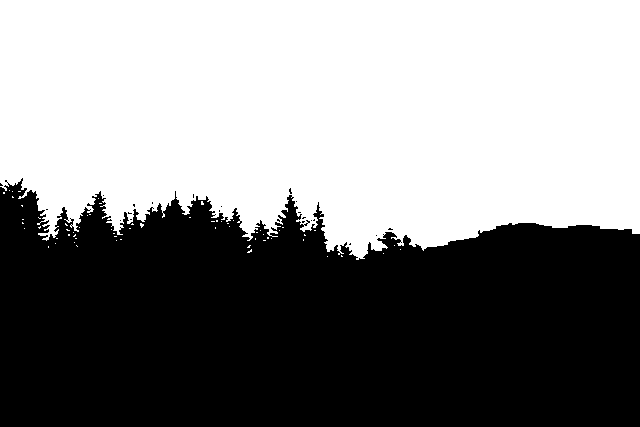} & \hspace{-0.4cm} 
			\includegraphics[width = 0.15\textwidth,height=0.08\textheight]{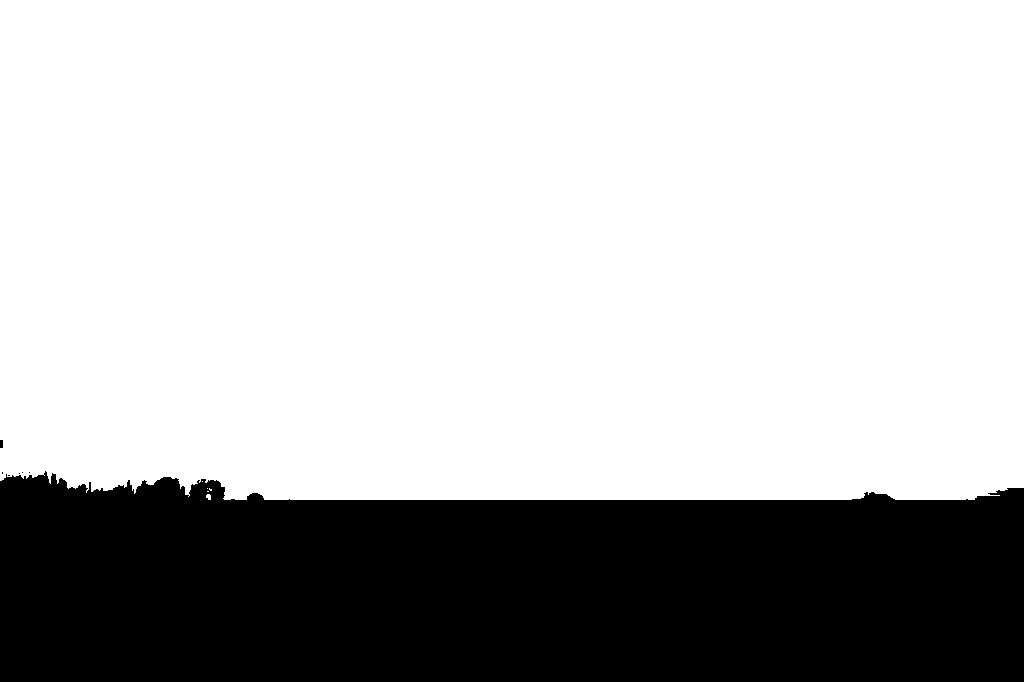} \\
			
			city & \hspace{-0.4cm}
			landscape & \hspace{-0.4cm}
			water \\
		\end{tabular}
	\end{center}
	\vspace{-0.5cm}
	\caption{The samples from FSID. The images in the top row are foggy images in different scenes. The bottom row represents the labeled sky region for the above images.}
	\vspace{-0.3cm}
	\label{sky-sample}
\end{figure}

In segmentation block, we add residual blocks in addition to encoding network and decoding network (as shown in Fig. \ref{synthetic-structure}). Similarly, we defined the outputs of each layer in encoding, decoding, and residual network as $E_{s}^{i1}$, $D_{s}^{j1}$, and $R_{s}^{k}$ respectively ($i1=1,...,m_{1}$; $j1=1...,n_{1}$; $k=1,...,K$; $m_{1}$, $n_{1}$ and $K$ are the number of layers). In this block, $m_{1}=3$, $n_{1}=2$, and $K=6$. Therefore, the encoding process in segmentation block is formulated as:
\begin{equation}
	{E}_{s}^{1} = \delta(\eta(C_{7\times 7}(E_{M})), {E}_{s}^{2} = \mathcal{M}({E}_{s}^{1}), {E}_{s}^{3} = \mathcal{M}({E}_{s}^{2})
\end{equation}

For $R_{s}^{k}$, we use the equation \ref{res} to obtain. Note that, $ \mathcal{F}_{re}^{0} =  {E}_{s}^{3}$. Then, the decoding process is formulated as:
\begin{equation}
	\begin{split}
		{D}_{s}^{1} = \mathcal{U}_{3\times 3}({R}_{s}^{6}),
		{D}_{s}^{2} = \mathcal{U}_{3\times 3}(\mathcal{M}([{D}_{s}^{1},{E}_{s}^{2}]))
	\end{split}
\end{equation}

After the above process, the final output segmentation image ($O_M \in [-1,1]$) is defined as:
\begin{equation}
	\begin{split}
		O_{M}= \tau(C_{7\times7}(\mathcal{M}([{D}_{s}^{2},{E}_{s}^{1}]))
	\end{split}
\end{equation}

Thus, the atmospheric light $A$ is calculated as follows:
\begin{equation}
	\begin{split}
		A = mean(O_{M} \odot I_{M})
	\end{split}
\end{equation}
where $mean(.)$ denotes an average filter, $\odot$ denotes the element-wise product. However, due to the small capacity of our GPU, we used a pre-trained SSM to segment the sky region to improve the training efficiency of our  QPC-Net. Please refer to our online address (https://github.com/ChengChen-ai/Sky-Segmentation) for details about our SSM. Moreover, we proposed a foggy-sky image dataset (FSID) for sky segmentation research. It contains 900 natural outdoor scenes, including traffic scenes, landscape, water scenes, etc. (as shown in Fig. \ref{sky-sample}). In addition, similar to \cite{liu2020end}, if the image has no sky region, we followed by the dark channel \cite{He2011PAMI} that  picked the maximum of the top 0.1 percent brightest pixels in the image as $A$.
\begin{figure}[H]\scriptsize
	\begin{center}
		\vspace{-0.2cm}
		\begin{tabular}{@{}c@{}}
			\includegraphics[width = 0.49\textwidth]{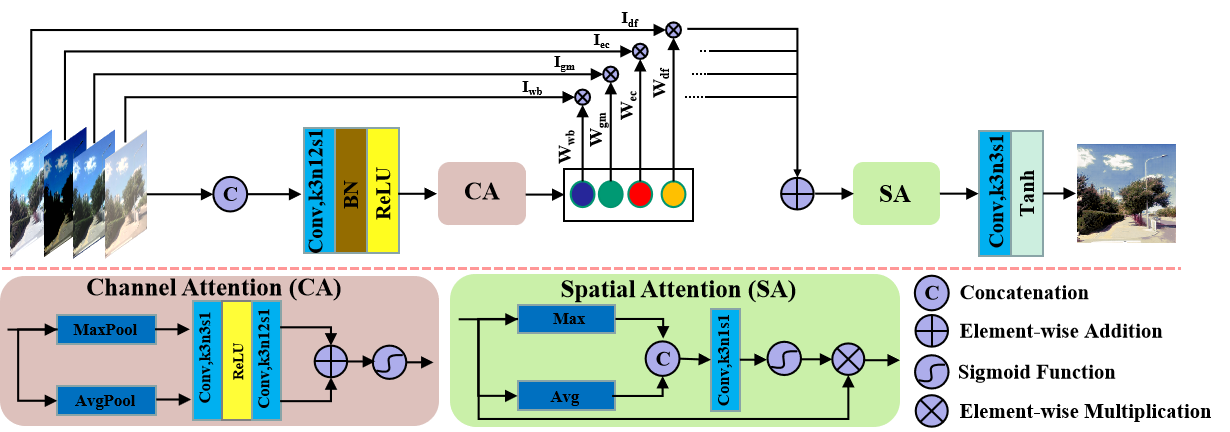}
		\end{tabular}
	\end{center}
	\vspace{-0.5cm}
	\caption{The architecture of Holistic Attention-Fusion generator.}
	\vspace{-3mm}
	\label{attention-structure}
\end{figure}

\textbf{\textit{Holistic Attention-Fusion generator:}}  To further improve the ability of transformation to map fog images to a fogfree image domain through the defogging generator, the output of $M_R$ and three derived images of the foggy image are input into a color-texture recovery network to refine correlated information for generating a better-recovered image with a color coherence and exactness of detail. We refer to this network as Holistic Attention-Fusion generator, as shown in Fig. \ref{attention-structure}. 
Inspired by \cite{GFN2018CVPR}, we also use the gated fusion strategy to recover the color and visibility of the defogged image with three derived inputs of the foggy image, such as white balanced image $I_{wb}$, contrast enhanced image $I_{ce}$, and gamma corrected image $I_{gc}$. 

In past studies, white balance has been found to have the potential to restore color in scenes, which is intended to eliminate chromatic aberrations caused by atmospheric colors. In this paper, we use the technology based on the gray world hypothesis \cite{GFN2018CVPR} \cite{wb2001} to obtain the white balance feature map of foggy images. The formula is as follows:
\begin{equation}
	\begin{split}
		\left\{\begin{matrix} I(R^{'})=I(R)*k_r\\ I(G^{'})=I(G)*k_g\\ I(B^{'})=I(B)*k_b \end{matrix}\right.
	\end{split}
\end{equation}
\begin{equation}
	\begin{split}
		I_{wb}=[I(R^{'}),I(G^{'}),I(B^{'})]
	\end{split}
\end{equation}
where, $k_r$, $k_g$ and $k_b$ represent the gain coefficients of R, G, and B channels respectively. $I_{wb}$ represents the result of a white balance operation.

To further improve the contrast of the results, we introduce the following two derived inputs. For $I_{ce}$, we exploit the Ancuti \cite{fusiondehazing2013TIP} proposed method to improve the global visibility of haze areas in foggy images.
\begin{equation}
	\begin{aligned}
		I_{ce}=\mu (I-\overline{I})
	\end{aligned}
	\label{ce}
\end{equation}
where $I$ is the foggy input, $\overline{I}$ is the average luminance value of $I$. $\mu$ is a linearly increase factor, $\mu=2(0.5 +\overline{I})$. However, it can be seen from (\ref{ce}) that as the value of $\overline{I}$ increases, $(I-\overline{I})$ may be negative, resulting in the dark area in foggy image tending to black after contrast enhancement. To address this problem, we adopt gamma correction:
\begin{equation}
	\begin{aligned}
		I_{gc}=\alpha I^{\gamma }
	\end{aligned}
\end{equation}

As discussed in \cite{GFN2018CVPR}, we also set $\alpha=1$ ,and the decoding gamma correction $\gamma=2.5$. Through gamma correction, we not only enhanced the visibility of the original foggy image, but also effectively avoided the appearance of the severe dark area.

After obtaining these derived inputs, how to fuse them together to optimize the defogged result is the key of this paper. In \cite{GFN2018CVPR}, however, Ren  \textit{et al.} ignored the influence of channel features and spatial features of each layer in the network on confidence maps respectively, resulting in the loss of partial texture details. Thus, we design a simple but effective method combining channel attention and spatial attention to fuse the confidence maps of each input to recover a higher quality image from foggy image, which can not only learns the hierarchical-specific knowledge from all preceding layers but also reduces the feature redundancy effectively. As shown in Fig. \ref{attention-structure}, we first concatenated each input image according to the channel direction to obtain the concatenation feature map $\mathcal{F}_{ct}$:
\begin{equation}
	\setlength\abovedisplayskip{6pt}
	\begin{split}
		\mathcal{F}_{ct}=[I_{df},I_{ec},I_{gm},I_{wb}]
	\end{split}
\end{equation}
where $I_{df}$ is the output of $M_R$. Then, a layer of convolutional network is used to initialize the global spatial information of the concatenation feature map to highlight the effective information of each channel. Thereafter, we use a channel attention network to assign weights to different channels of $\mathcal{F}_{ct})$. For channel attention network, we first use global average pooling and maximum pooling to compress the global spatial features of channels into two different spatial feature expressions: $g_{avg}^{c} \epsilon \mathbb{R}^{C\times 1\times 1}$ and $g_{max}^{c} \epsilon \mathbb{R}^{C\times 1\times 1}$. Then, we generate the channel-attention map $W_{c}\epsilon \mathbb{R}^{C\times 1\times 1}$ through a shared network:
\begin{equation}
	\begin{split}
		W_{c}&=\sigma (C_{3\times3}(\delta (C_{3\times3}(g_{avg}^{c})))\\
		&+C_{3\times3}(\delta (C_{3\times3}(g_{max}^{c})))), c \in \{df, wb, ec, gm\}
	\end{split}
\end{equation}
where $W_{df}$ is the weight of $I_{df}$, $W_{wb}$ is the white balanced weight, $W_{ec}$ is the contrast enhanced weight, and $W_{gm}$ is the gamma correction weight. Finally, we get the preliminary fusion feature $\mathcal{F}_{ct}^{*}$ through the following formula:
\begin{equation}
	\begin{aligned}
		\mathcal{F}_{ct}^{*} &=W_{c}\cdot \mathcal{F}_{ct} =(W_{df},W_{ec},W_{gm},W_{wb})\cdot (I_{df},I_{ec},I_{gm},I_{wb})\\& =W_{df}\odot I_{df} + W_{ec}\odot I_{ec} + W_{gm}\odot I_{gm} + W_{wb}\odot I_{wb}
	\end{aligned}
\end{equation}
where $\cdot$ denotes inner-vector product. To further uptake of channel dimension information, we propose a spatial attention network, which can contain responses from all dimensions of the feature
map $\mathcal{F}_{ct}^{*}$. It pays more attention to the location features information by assigning greater weight to high-frequency signals and dense fog areas. Along the channel direction, we calculate the mean value $g_{avg}^{s} \epsilon \mathbb{R}^{1\times H\times W}$ and maximum value $g_{max}^{s} \epsilon \mathbb{R}^{1\times H\times W}$ of $\mathcal{F}_{ct}^{*}$ in the spatial dimension to effectively highlight fog relevant feature to preserve more texture details for generated result. Then we concatenate these two spatial features and put them into a convolution layer and a sigmoid function to obtain the weight $W_{s}\epsilon \mathbb{R}^{1\times H\times W}$ of $\mathcal{F}_{ct}^{*}$:
\begin{equation}
	\begin{aligned}
		W_{s}=\sigma (C_{3\times3}([g_{avg}^{s},g_{max}^{s}]))
	\end{aligned}
\end{equation}

The final attention-fusion feature is calculated as follows:
\begin{equation}
	\begin{aligned}
		\mathcal{F}_{fusion}=W_{s} \odot \mathcal{F}_{ct}^{*}
	\end{aligned}
\end{equation}

\begin{figure*}[bp]\footnotesize
	\begin{center}
		\vspace{-0.2cm}
		\resizebox{\textwidth}{!}{
			\begin{tabular}{@{}cccccccc@{}}
				
				\includegraphics[width = 0.19\textwidth,height=0.165\textheight]{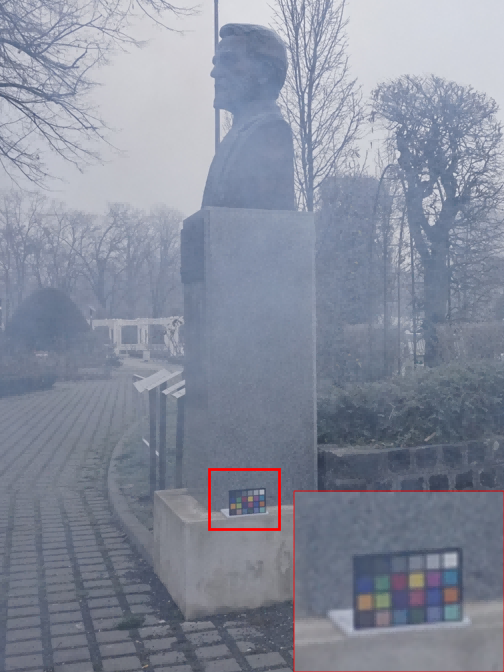} & \hspace{-0.45cm}
				\includegraphics[width = 0.19\textwidth,height=0.165\textheight]{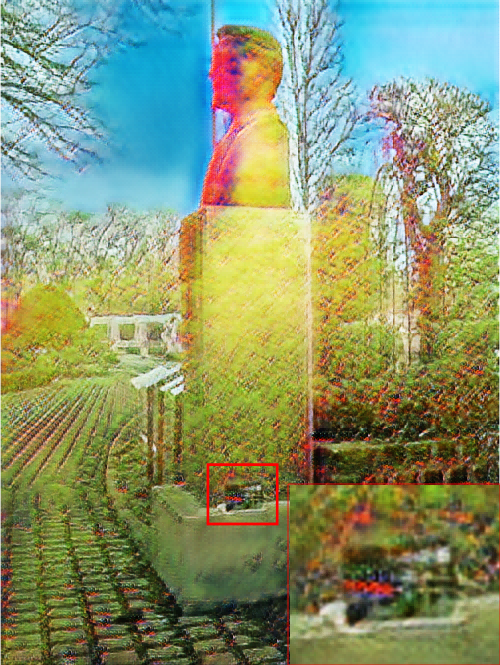} & \hspace{-0.45cm}
				\includegraphics[width = 0.19\textwidth,height=0.165\textheight]{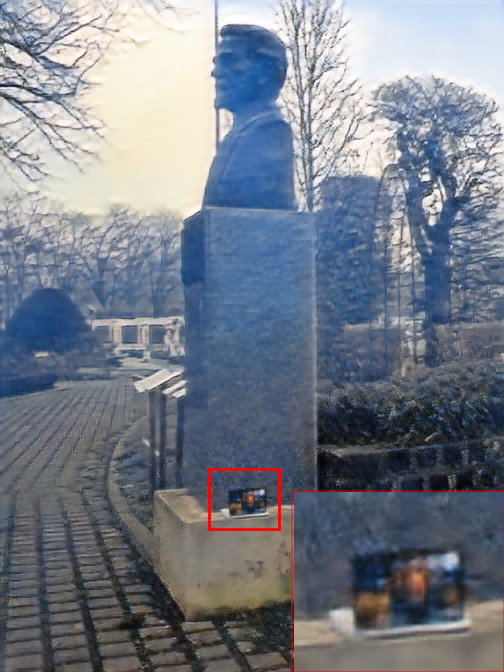} & \hspace{-0.45cm}
				\includegraphics[width = 0.19\textwidth,height=0.165\textheight]{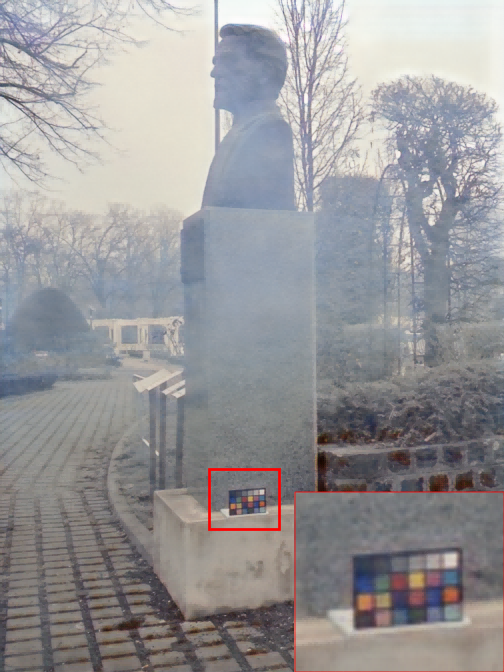} & \hspace{-0.45cm}
				\includegraphics[width = 0.19\textwidth,height=0.165\textheight]{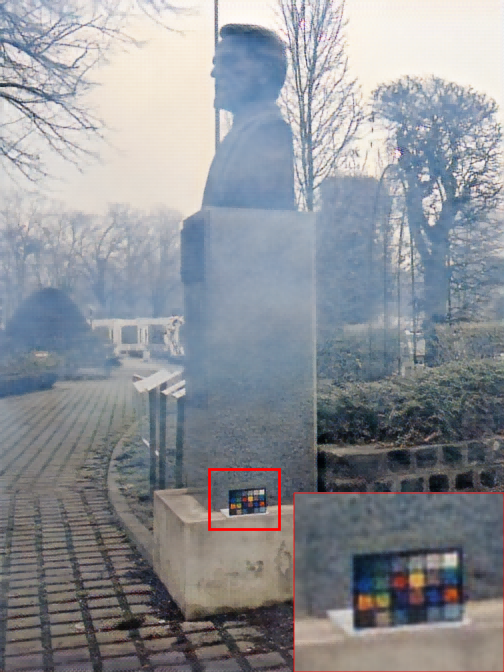} 
				& \hspace{-0.45cm}
				\includegraphics[width = 0.19\textwidth,height=0.165\textheight]{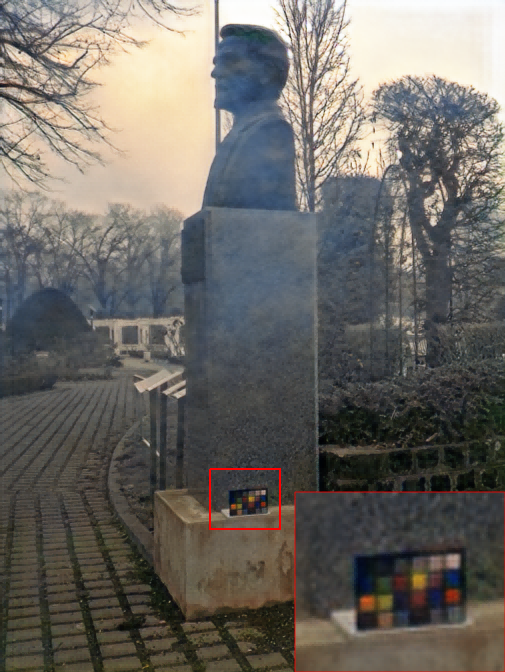} & \hspace{-0.45cm}
				\includegraphics[width = 0.19\textwidth,height=0.165\textheight]{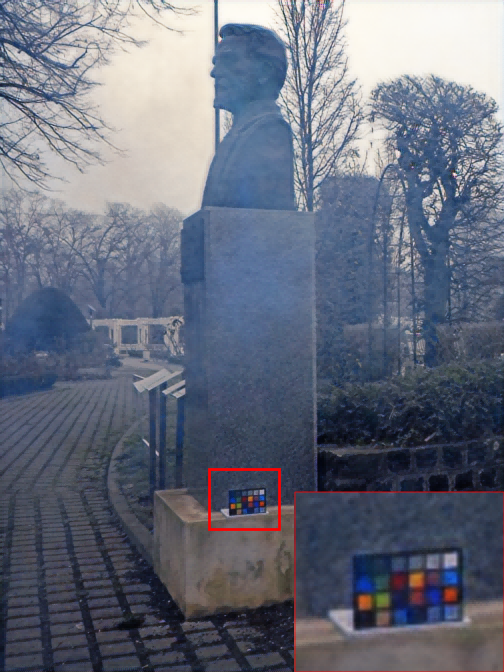}
				& \hspace{-0.45cm}
				\includegraphics[width = 0.19\textwidth,height=0.165\textheight]{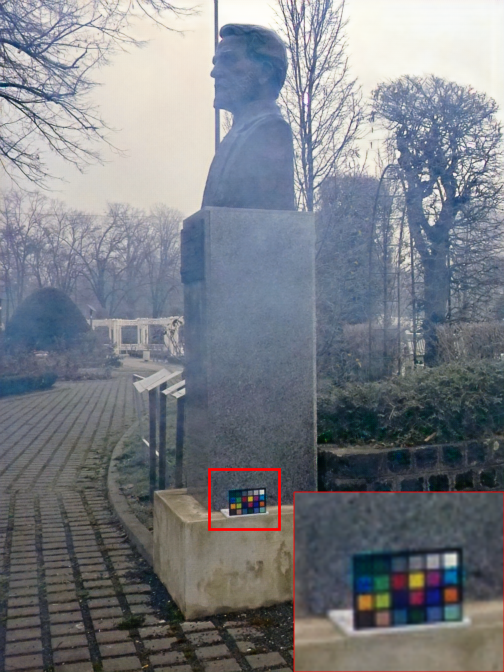} \\
				
				NIQE=3.3808
				&\hspace{-0.45cm} NIQE=3.1257 
				&\hspace{-0.45cm} NIQE=2.7277 
				&\hspace{-0.45cm} NIQE=2.1934
				&\hspace{-0.45cm} NIQE=2.4156 
				&\hspace{-0.45cm} NIQE=2.2274 
				&\hspace{-0.45cm} NIQE=2.4728 
				&\hspace{-0.45cm} NIQE=2.1456 \\
				
				\includegraphics[width = 0.19\textwidth]{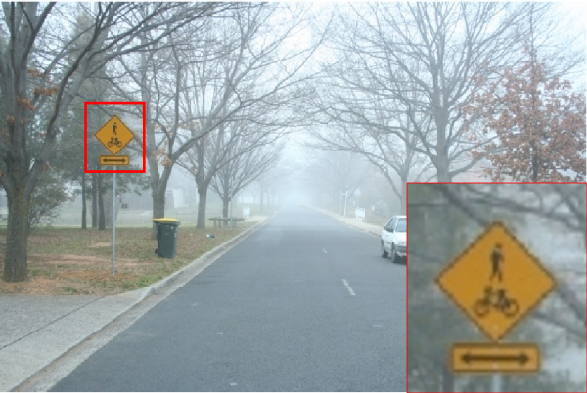} & \hspace{-0.45cm}
				\includegraphics[width = 0.19\textwidth]{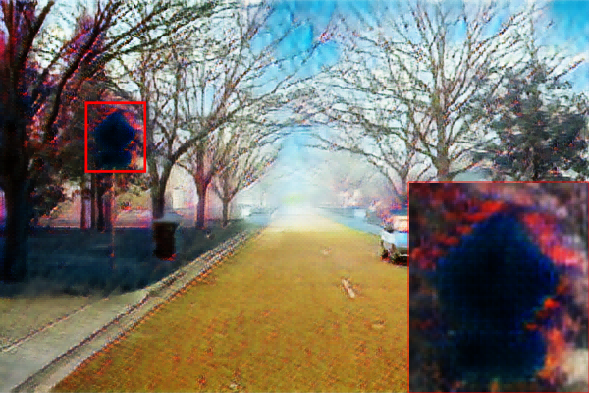} & \hspace{-0.45cm}
				\includegraphics[width = 0.19\textwidth]{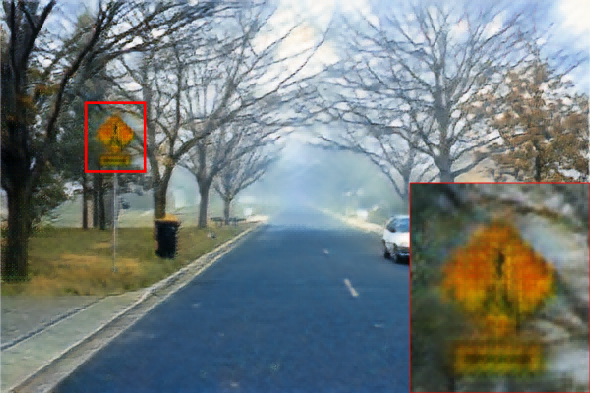} & \hspace{-0.45cm}
				\includegraphics[width = 0.19\textwidth]{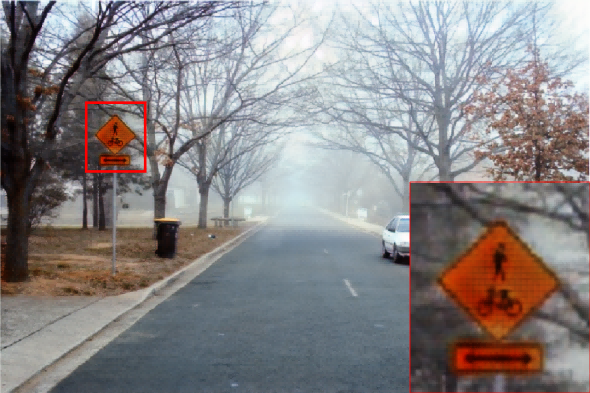} & \hspace{-0.45cm}
				\includegraphics[width = 0.19\textwidth]{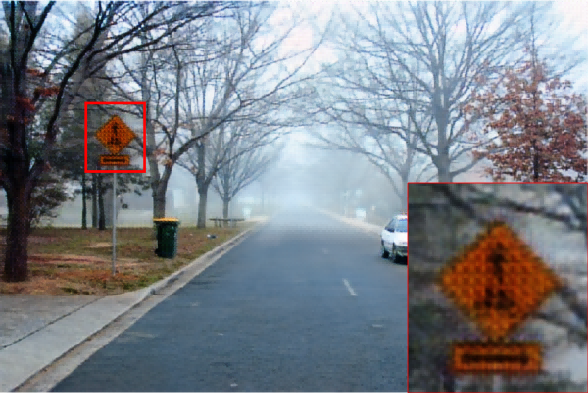} 
				& \hspace{-0.45cm}
				\includegraphics[width = 0.19\textwidth]{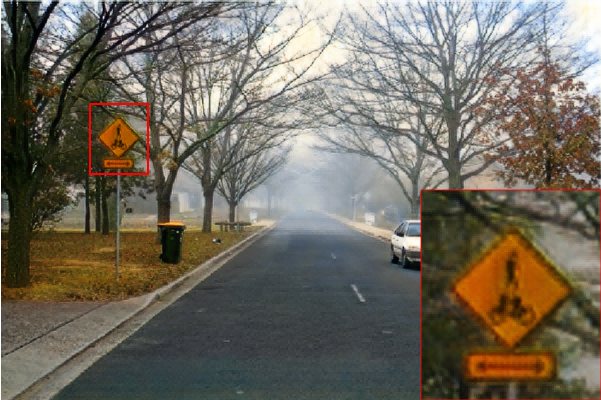} & \hspace{-0.45cm}
				\includegraphics[width = 0.19\textwidth]{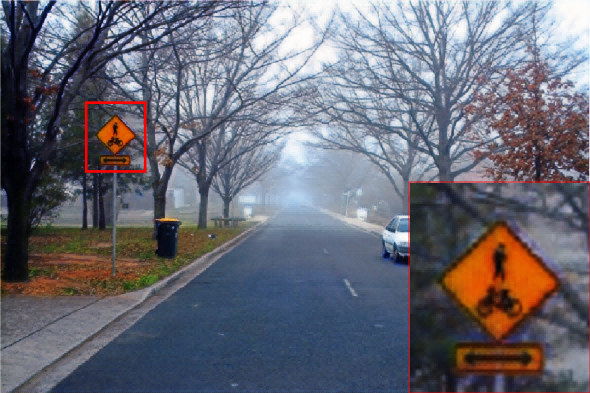}
				& \hspace{-0.45cm}
				\includegraphics[width = 0.19\textwidth]{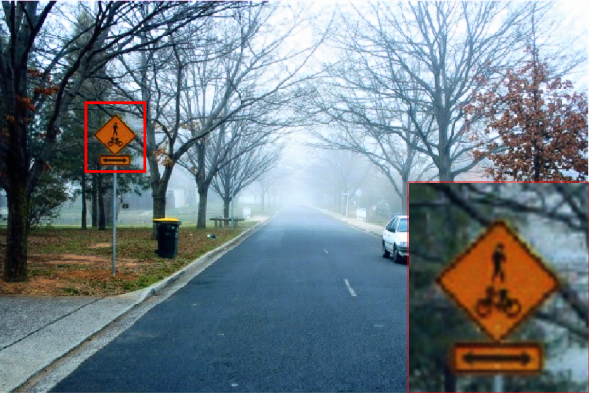} \\
				
				NIQE=2.6351 
				&\hspace{-0.45cm} NIQE=3.1334 
				&\hspace{-0.45cm} NIQE=2.7820 
				&\hspace{-0.45cm} NIQE=2.6722
				&\hspace{-0.45cm} NIQE=2.6277 
				&\hspace{-0.45cm} NIQE=2.2740 
				&\hspace{-0.45cm} NIQE=2.4565
				&\hspace{-0.45cm} NIQE=2.1156 \\

				(a) Foggy Image & \hspace{-0.45cm}
				(b) CycleGAN \cite{zhu2017unpaired} & \hspace{-0.45cm}
				(c) Cycle-dehaze \cite{cycle-dehazing} & \hspace{-0.45cm}
				(d) w/o VGG & \hspace{-0.45cm}
				(e) w/o DC & \hspace{-0.45cm}
				(f) w/o ASM & \hspace{-0.45cm}
				(g) w/o HAG & \hspace{-0.45cm}
				(h) QPC-Net 
				\end{tabular}}
	\end{center}
	\vspace{-0.5cm}
	\caption{Comparison results of different configurations in visual effects on real-world outdoor images.}
	\vspace{-0.1cm}
	\label{ablation_dehaze}
\end{figure*}

\begin{figure}[H]\scriptsize
	\begin{center}
		\vspace{-0.2cm}
		\begin{tabular}{@{}c@{}}
			\includegraphics[width = 0.5\textwidth]{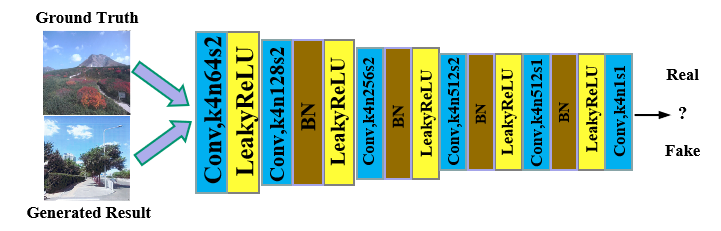}
		\end{tabular}
	\end{center}
	\vspace{-0.5cm}
	\caption{The architcture of Discriminator.}
	\vspace{-3mm}
	\label{Discriminate-structure}
\end{figure}

\textbf{\textit{Discriminator:}} As shown in Fig. \ref{architctures}, we designed two discriminator networks and named them as $D_{ff}$ and $D_f$. For $D_{ff}$, it is used to distinguish the generated results of $M_{R}$ and $M_{CTR}$ from the input clear images. For $D_f$, it is used to distinguish between the generated results from $M_{S}$ and the input foggy images. As can be seen in Fig. \ref{Discriminate-structure}, our discriminator consists of 6 convolutional blocks. The LeakyReLU function and Batch Normalization (BN) are used alternately. From the beginning to the end, the number of convolutional filters at each layer is 64, 128, 256, 512, 512 and 1. The stride of the first four layers is 2, and the stride of the last two layers is 1. Moreover, all filters have the same convolution kernel size, which are $4\times4$.
\subsection{Loss Function}
To train our proposed holistic attention-fusion adversarial defogging model, our loss function consists of three terms: 1) adversarial loss $\mathcal{L} _{R_{adv}}$, 	$\mathcal{L} _{CTR_{adv}}$, $\mathcal{L} _{S_{adv}}$; 2) cycle-consistency loss $\mathcal{L}_{c1}$, $\mathcal{L}_{c2}$; 3) perceptual loss $\mathcal{L}_{Per}$. The total loss of our network is presented as:
\begin{equation}
	\begin{aligned}
		\mathcal{L}_{Total} = &\lambda _1\mathcal{L} _{R_{adv}}+\lambda _2\mathcal{L} _{CTR_{adv}}+\lambda _3\mathcal{L} _{S_{adv}}+\\
		&\lambda _4\mathcal{L}_{c1}+
		\lambda _5\mathcal{L}_{c2}+\lambda _6\mathcal{L}_{Per}
	\end{aligned}
	\label{total}
\end{equation}
where $\lambda _1$, $\lambda _2$, $\lambda _3$, $\lambda _4$, $\lambda _5$ and $\lambda _6$ are the
positive weights, which are used to balance the importance of the corresponding loss.

\textbf{\textit{Adversarial Loss:}} We considered three losses of adversarial learning including $\mathcal{L} _{R_{adv}}$, $\mathcal{L} _{CTR_{adv}}$ and $\mathcal{L} _{S_{adv}}$, where $\mathcal{L} _{R_{adv}}$ and $\mathcal{L} _{CTR_{adv}}$ fooled the discriminator $D_{ff}$ by encouraging Defogging generator and Holistic Attention-Fusion generator to recover high-quality clean image, and $\mathcal{L} _{S_{adv}}$ aims to fool the $D_{f}$ using Synthesizing generator to synthesize a more realistic foggy image. These adversarial losses are defined as:
\begin{equation}
	\begin{aligned}
		\mathcal{L} _{R_{adv}}(M_{R},D_{ff})=&\underset{y_r\sim Y_R}{\mathbb{E}}\left[logD_{ff}(y_r)\right]+\\
		&\underset{x_r\sim X_R}{\mathbb{E}}\left[log(1-D_{ff}(M_{R}(x_r)))\right]
	\end{aligned}
\end{equation}
\begin{equation}
	\begin{aligned}
		\mathcal{L} _{CTR_{adv}}(M_{R},M_{CTR},D_{ff})=\underset{y_r\sim Y_R}{\mathbb{E}}\left[logD_{ff}(y_r)\right]+&\\
		\underset{x_r\sim X_R}{\mathbb{E}}\left[log(1-D_{ff}(M_{CTR}(M_{R}(x_r),I_{d}^{x_r})))\right]&
	\end{aligned}
\end{equation}
\begin{equation}
	\begin{aligned}
		\mathcal{L} _{S_{adv}}(M_{S},D_{f})=&\underset{x_r\sim X_R}{\mathbb{E}}\left[logD_{f}(x_r)\right]+\\
		&\underset{y_r\sim Y_R}{\mathbb{E}}\left[log(1-D_{f}(M_{S}(y_r)))\right]
	\end{aligned}
\end{equation}
where $X_R$ represents the real-world fog image domain and $Y_R$ refers to the real-world clear image domain; $x_r$ and $y_r$ represent the input image from $X_R$ and $Y_R$ respectively. $M_{R}(.)$, $M_{CTR}(.)$ and $M_{S}(.)$ represent the results generated by Fog Removal Module, Color-Texture Recovery Module and Fog Synthetic Module respectively; $I_{d}^{x_r}$ denotes the derived images of $x_r$.

\textbf{\textit{Cycle-consistency Loss:}} We applied a mean-square-error loss to make the final output of each path in the both Fog2Fogfree and Fogfree2Fog block close to the corresponding initial input. For Fog2Fogfree block, we obtained cycle-consistency loss function for foggy images among $I_{rf}$, $I^1_{rcf}$, and $I^2_{rcf}$:
\begin{equation}
	\begin{aligned}
		\mathcal{L}_{c1}=\left \|I_{rf}-I^1_{rcf} \right \|_{2}^{2}
		+\left \|I_{rf}-I^2_{rcf} \right \|_{2}^{2}
	\end{aligned}
\end{equation}
where $I_{rf}$ is the input of real-world foggy image, $I^1_{rcf}$ and $I^2_{rcf}$ represent the output of the defogging path and the color-texture recovery path respectively. For Fogfree2Fog block, cycle-consistency loss function for fogfree images is
constructed among $I_{rff}$, $I_{rcff}$, and $I^{sf}_{rr}$:
\begin{equation}
	\begin{aligned}
		\mathcal{L}_{c2}=\left \|I_{rff}-I_{rcff} \right \|_{2}^{2}
		+\left \|I_{rff}-I^{sf}_{rr} \right \|_{2}^{2}
	\end{aligned}
\end{equation}
where $I_{rff}$ is the input of real-world clean image, $I_{rcff}$ and $I^{sf}_{rr}$ denote the output of the synthesizing path and the color-texture recovery path respectively. 

\textbf{\textit{Perceptual Loss:}} To better extract more texture details from foggy images, we introduce a perceptual loss function, which is constructed with the pretrained VGG19 \cite{Simonyan2014} to compare the original input image with the reconstructed image in the feature space. This objective is defined as:
\begin{equation}
	\begin{aligned}
		\mathcal{L}_{Per}&=\left \|\phi _i(I_{rf})-\phi _i(I^1_{rcf}) \right \|_{2}^{2}
		+\left \|\phi _i(I_{rf})-\phi _i(I^2_{rcf}) \right \|_{2}^{2}
		\\&+\left \|\phi _i(I_{rff})-\phi _i(I_{rcff}) \right \|_{2}^{2}
		+\left \|\phi _i(I_{rff})-\phi _i(I^{sf}_{rr}) \right \|_{2}^{2}
	\end{aligned}
	\label{perceptual}
\end{equation}
where $\phi _i$ denotes the feature map in $i^{th}$ layer of VGG19.
\section{Experiments and Discussion}
\label{experiment}
\subsection{Implementation Setting and Details}
1) \textbf{\textit{Evaluation methods}} 

In this section, we compare the defogged performance of our proposed approach with that of eleven state-of-the-art methods. Among them, some are trained using paired data, including the DehazeNet \cite{dehazenet2016TIP} (TIP 2016), a densely connected pyramid dehazing network (DCPDN) \cite{zhang2018densely} (CVPR 2018), an enhanced pix2pix dehazing network (EPDN) \cite{pix2pixdehazing} (CVPR 2019), a gated context aggregation network (GCANet) \cite{chen2019gated} (WACV 2019), a domain adaptation for image dehazing (DA-Dehaze) \cite{shao2020domain} (CVPR 2020), a multi-scale boosted dehazing network (MSBDN) \cite{dong2020multi} (CVPR 2020)  and a principled synthetic-real dehazing (PSD) \cite{chen2021psd} (CVPR 2021). While other methods use unpaired data for training, including the Cycle-dehaze \cite{cycle-dehazing} (CVPR Workshops 2018), an image dehazing and exposure (IDE) \cite{IDE2021ide} (TIP 2021), a weakly supervised dehazing Refinement framework (RefineDNet) \cite{zhao2021refinednet} (TIP 2021) and a dehazing via decomposing transmission map into density and depth ($D^4$)  \cite{yang2022self} (CVPR 2022).

2) \textbf{\textit{Implementation details}} 

Due to our proposed QPC-Net uses unpaired training, we randomly choose both outdoor synthetic and real-world unpaired hazy images from the RESIDE dataset \cite{RESIDE}. The dataset is widely used, which is divided into five subsets: Indoor Training Set (ITS), Outdoor Training Set (OTS), Synthetic Object Testing Set (SOTS), Unannotated real Hazy Images (URHI) and real Task-driven Testing Set (RTTS). We train the network by randomly selecting 10000 synthetic foggy images from the OTS and 8000 real foggy images from the URHI and RTTS.

Our QPC-Net is implemented by PyTorch 1.8.0 with one NVIDIA GeForce GTX 3060 GPU. During training, we resize all the images to $256 \times 256 $ and normalize the pixel values to [-1,1]. The models are trained using ADAM optimizer with exponential decay rates $\beta$ equal to 0.999, meanwhile the learning rate and batchsize are initialised to $1 \times 10^{-4}$ and 2, respectively. We set the total number of epoch to 15 for 270000 iterations. The perceptual loss in (\ref{perceptual}) is set using the latent features of the 2nd and 5th layers from the fixed pre-trained VGG19, and empirically initialise default values of $\lambda _1$, $\lambda _2$, $\lambda _3$, $\lambda _4$, $\lambda _5$ and $\lambda _6$ in (\ref{total}) are set to 10, 10, 10, 5, 5, 1 respectively.
\begin{figure*}[tp]\tiny
	\begin{center}
		\vspace{-0.2cm}
		\resizebox{\textwidth}{!}{
			\begin{tabular}{@{}ccccccc@{}}
				
				\includegraphics[width = 0.09\textwidth,height=0.05\textheight]{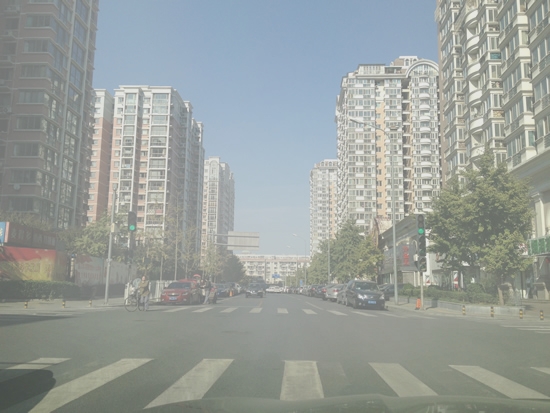} & \hspace{-0.45cm}
				\includegraphics[width = 0.09\textwidth,height=0.05\textheight]{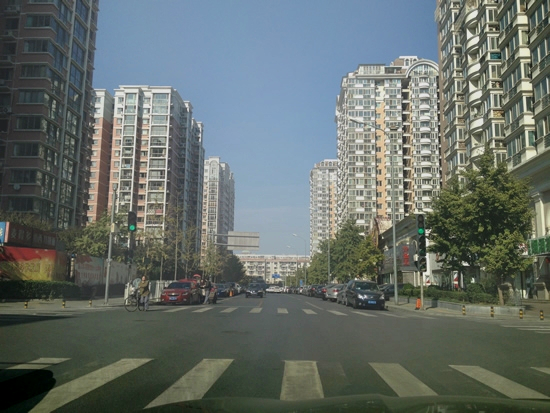} & \hspace{-0.45cm}
				\includegraphics[width = 0.09\textwidth,height=0.05\textheight]{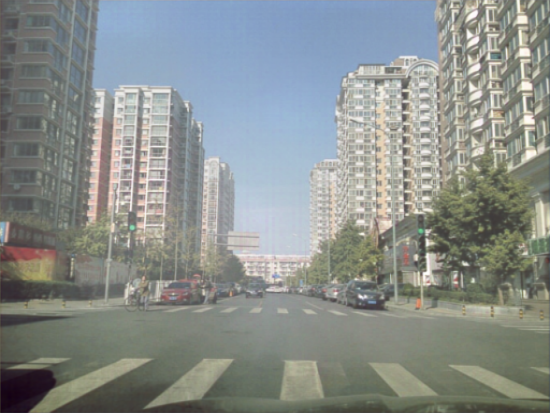} & \hspace{-0.45cm}
				\includegraphics[width = 0.09\textwidth,height=0.05\textheight]{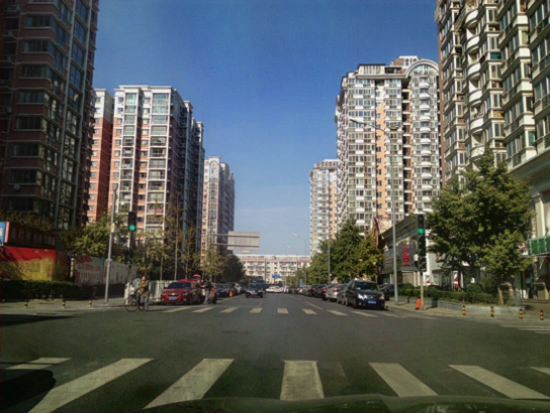} & \hspace{-0.45cm}
				\includegraphics[width = 0.09\textwidth,height=0.05\textheight]{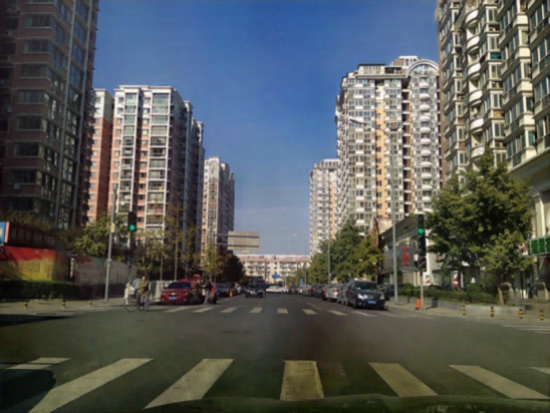} & \hspace{-0.45cm}
				\includegraphics[width = 0.09\textwidth,height=0.05\textheight]{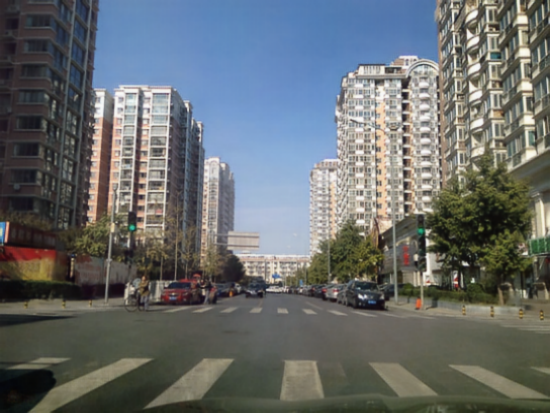} & \hspace{-0.45cm}
				\includegraphics[width = 0.09\textwidth,height=0.05\textheight]{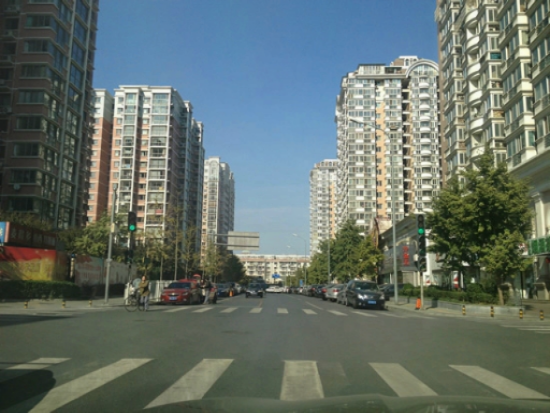} \\
				
				PSNR/SSIM
				& \hspace{-0.45cm} 23.607/0.948
				& \hspace{-0.45cm} 16.322/0.846
				& \hspace{-0.45cm} 21.731/0.927
				& \hspace{-0.45cm} 20.313/0.917
				& \hspace{-0.45cm} 27.105/0.948
				& \hspace{-0.45cm} \textbf{28.996}/0.967\\
				
				\includegraphics[width = 0.09\textwidth,height=0.07\textheight]{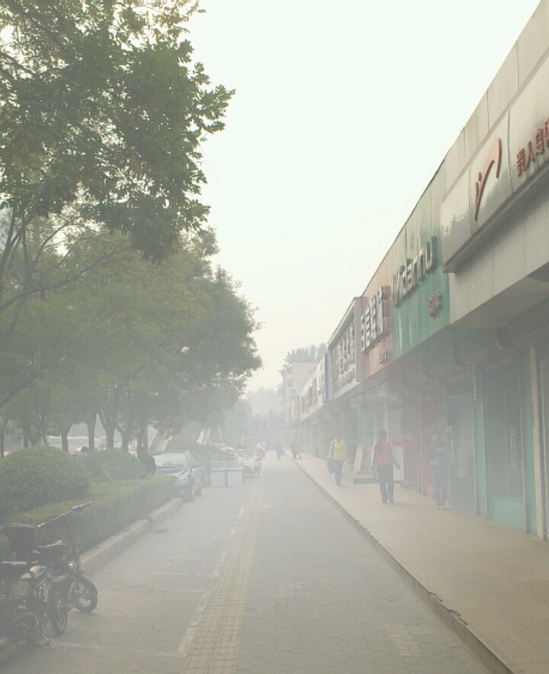} & \hspace{-0.45cm}
				\includegraphics[width = 0.09\textwidth,height=0.07\textheight]{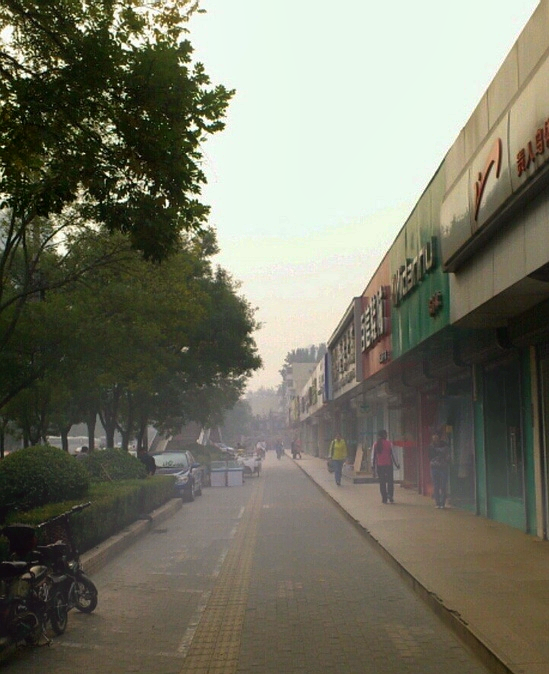} & \hspace{-0.45cm}
				\includegraphics[width = 0.09\textwidth,height=0.07\textheight]{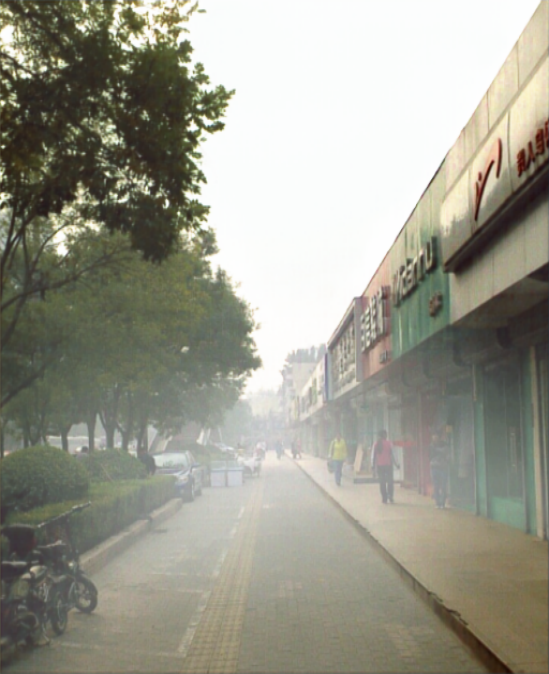} & \hspace{-0.45cm}
				\includegraphics[width = 0.09\textwidth,height=0.07\textheight]{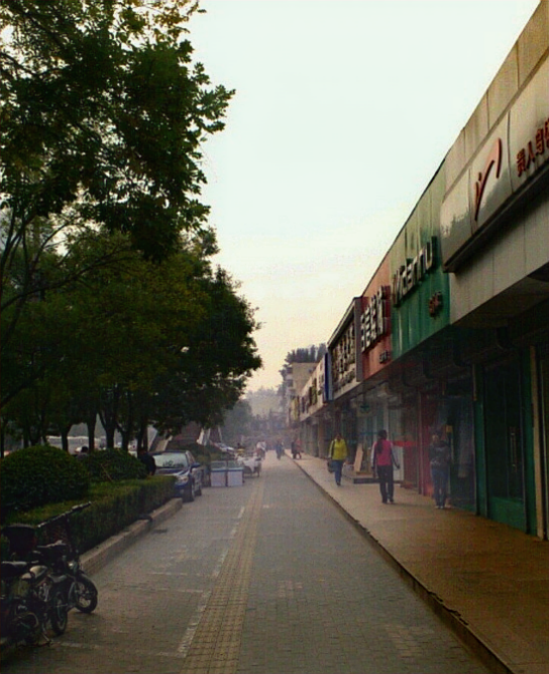} & \hspace{-0.45cm}
				\includegraphics[width = 0.09\textwidth,height=0.07\textheight]{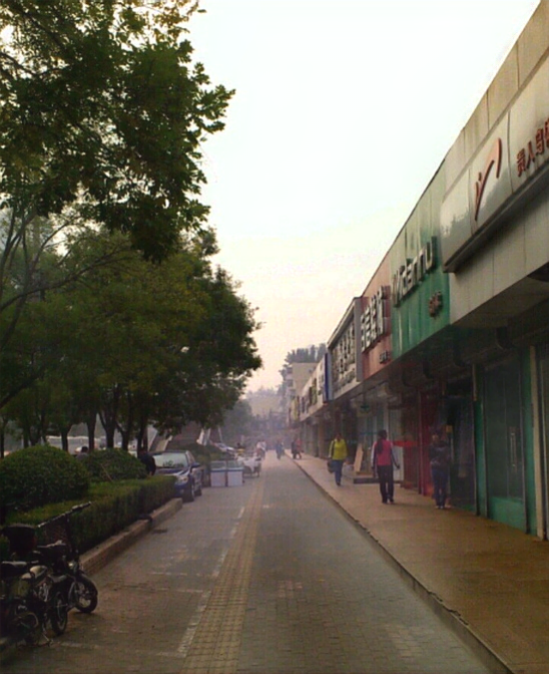} & \hspace{-0.45cm}
				\includegraphics[width = 0.09\textwidth,height=0.07\textheight]{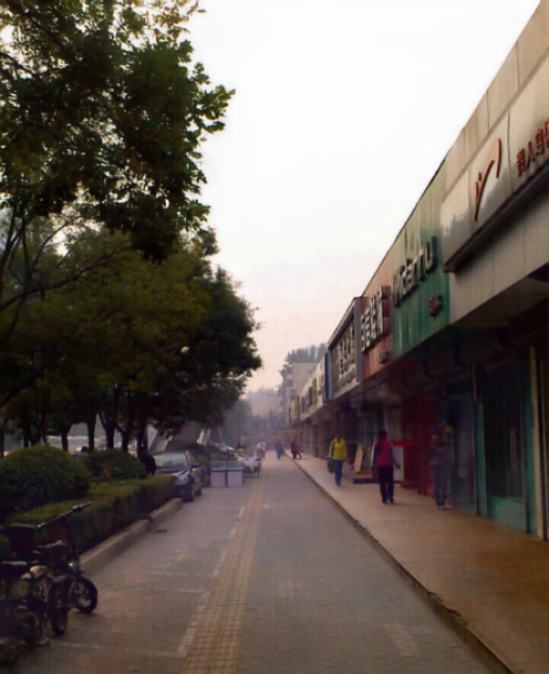} & \hspace{-0.45cm}
				\includegraphics[width = 0.09\textwidth,height=0.07\textheight]{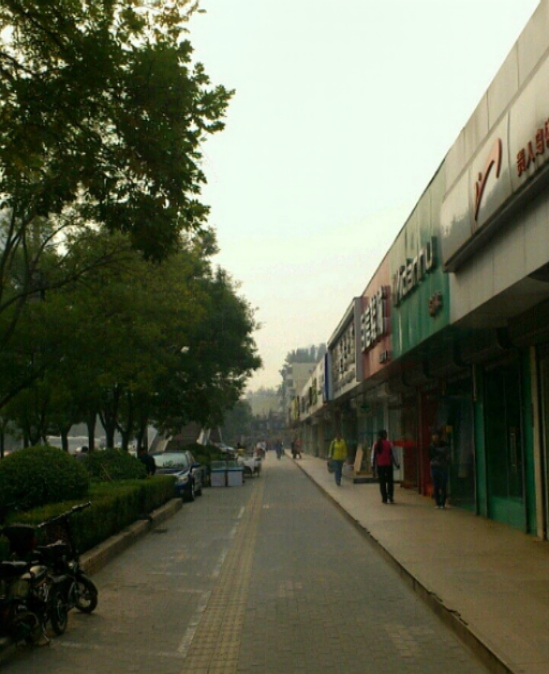} \\
				
				PSNR/SSIM
				& \hspace{-0.45cm} 19.946/0.857
				& \hspace{-0.45cm} 11.740/0.688
				& \hspace{-0.45cm} 22.921/0.888
				& \hspace{-0.45cm} 22.969/0.873
				& \hspace{-0.45cm} 23.867/0.867
				& \hspace{-0.45cm} 29.833/0.935\\
				
				(a) Input & \hspace{-0.45cm}
				(b) DehazeNet \cite{dehazenet2016TIP} & \hspace{-0.45cm}
				(c) DCPDN \cite{zhang2018densely} & \hspace{-0.45cm}
				(d) EPDN \cite{pix2pixdehazing} & \hspace{-0.45cm}
				(e) GCANet \cite{chen2019gated}& \hspace{-0.45cm}
				(f) DA-Dehaze \cite{shao2020domain}& \hspace{-0.45cm}
				(g) MSBDN \cite{dong2020multi}\\
				
				\includegraphics[width = 0.09\textwidth,height=0.05\textheight]{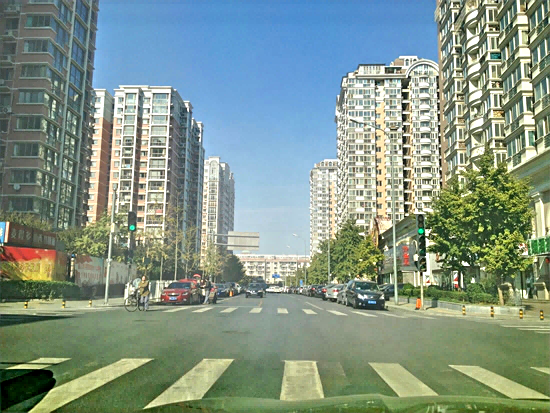} & \hspace{-0.45cm}
				\includegraphics[width = 0.09\textwidth,height=0.05\textheight]{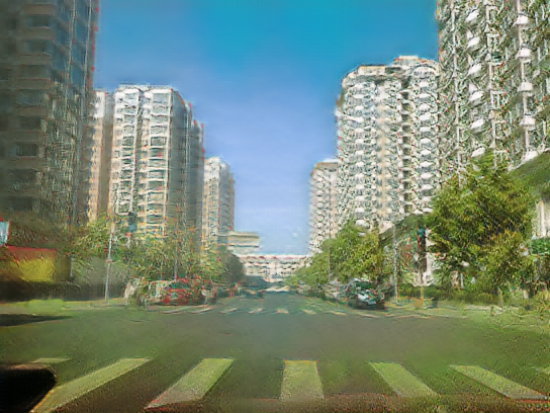} & \hspace{-0.45cm}
				\includegraphics[width = 0.09\textwidth,height=0.05\textheight]{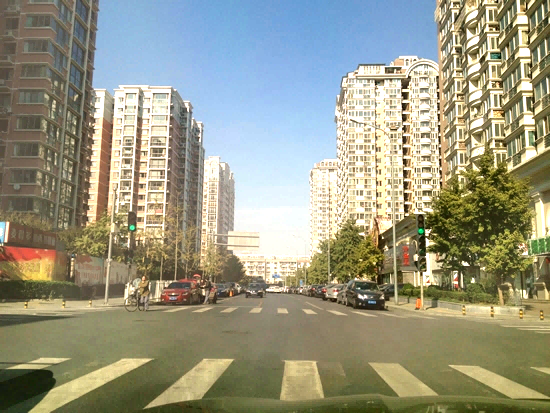} & \hspace{-0.45cm}
				\includegraphics[width = 0.09\textwidth,height=0.05\textheight]{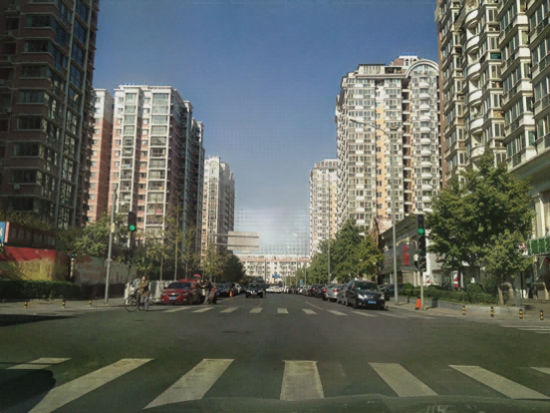} & \hspace{-0.45cm}
				\includegraphics[width = 0.09\textwidth,height=0.05\textheight]{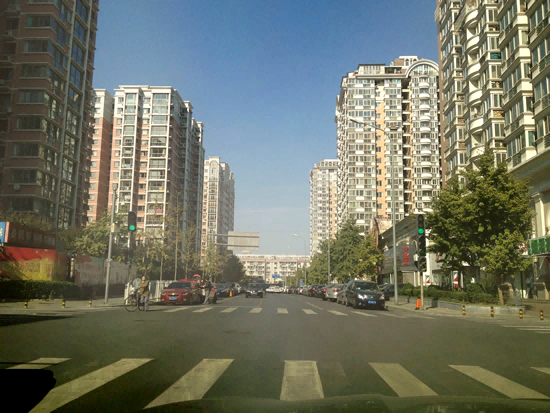} & \hspace{-0.45cm}
				\includegraphics[width = 0.09\textwidth,height=0.05\textheight]{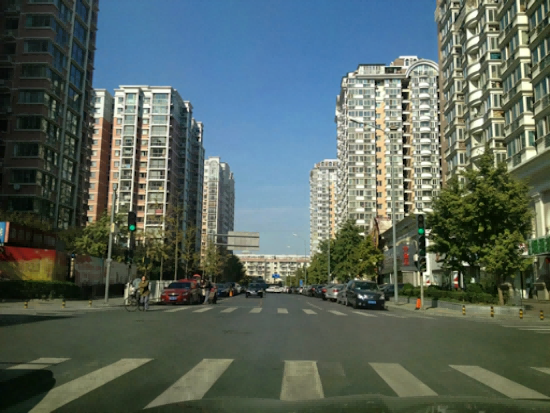} & \hspace{-0.45cm}
				\includegraphics[width = 0.09\textwidth,height=0.05\textheight]{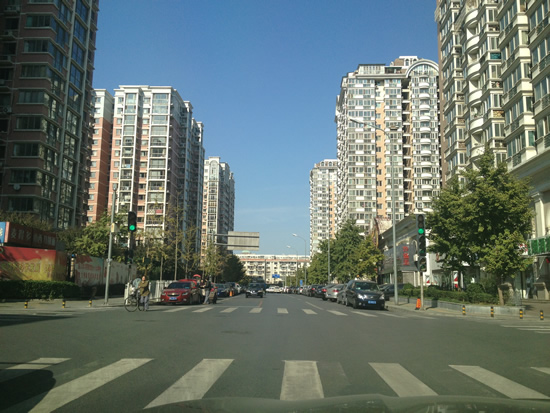} \\
				
				16.876/0.851
				& \hspace{-0.45cm} 18.315/0.661
				& \hspace{-0.45cm} 15.161/0.865
				& \hspace{-0.45cm} 23.965/0.951
				& \hspace{-0.45cm} 23.348/0.938
				& \hspace{-0.45cm} 28.576/\textbf{0.968}
				& \hspace{-0.45cm} $+\infty$/1\\
				
				\includegraphics[width = 0.09\textwidth,height=0.07\textheight]{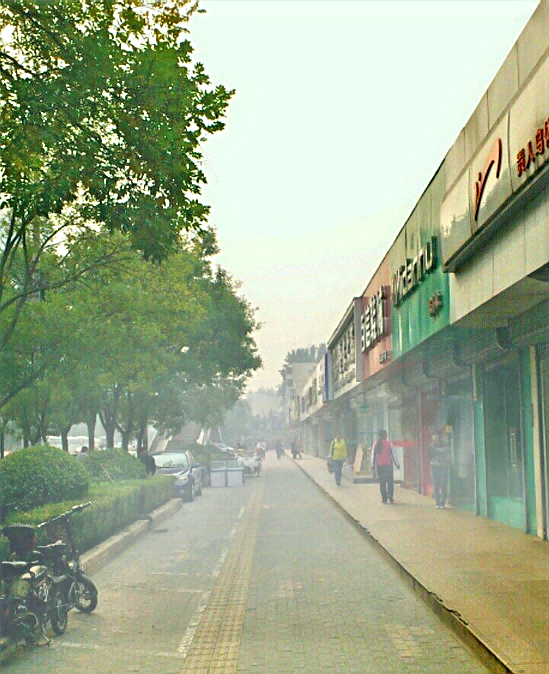} & \hspace{-0.45cm}
				\includegraphics[width = 0.09\textwidth,height=0.07\textheight]{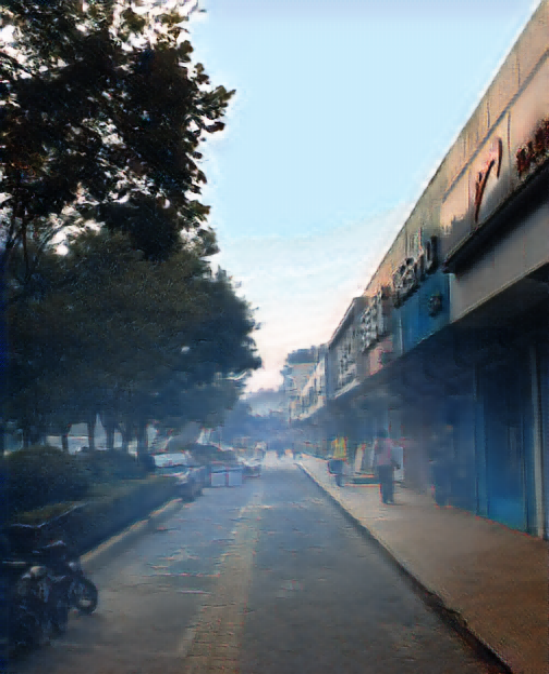} & \hspace{-0.45cm}
				\includegraphics[width = 0.09\textwidth,height=0.07\textheight]{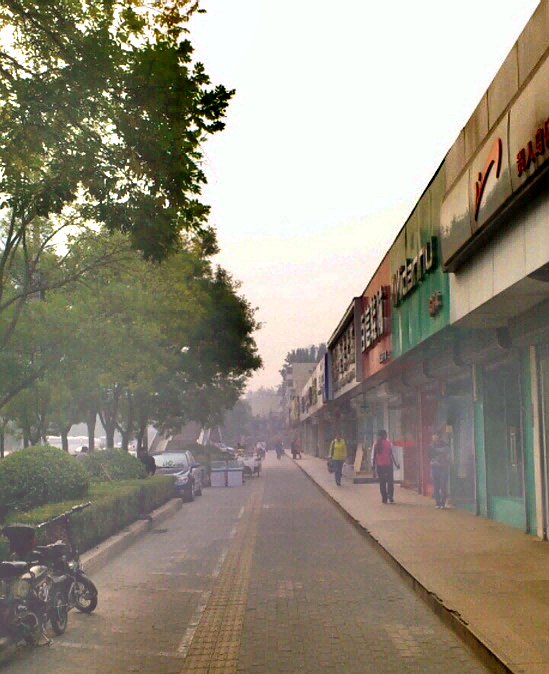} & \hspace{-0.45cm}
				\includegraphics[width = 0.09\textwidth,height=0.07\textheight]{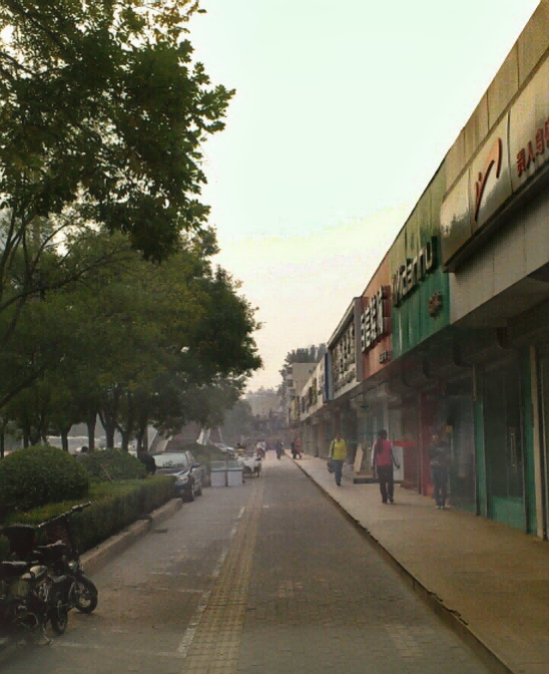} & \hspace{-0.45cm}
				\includegraphics[width = 0.09\textwidth,height=0.07\textheight]{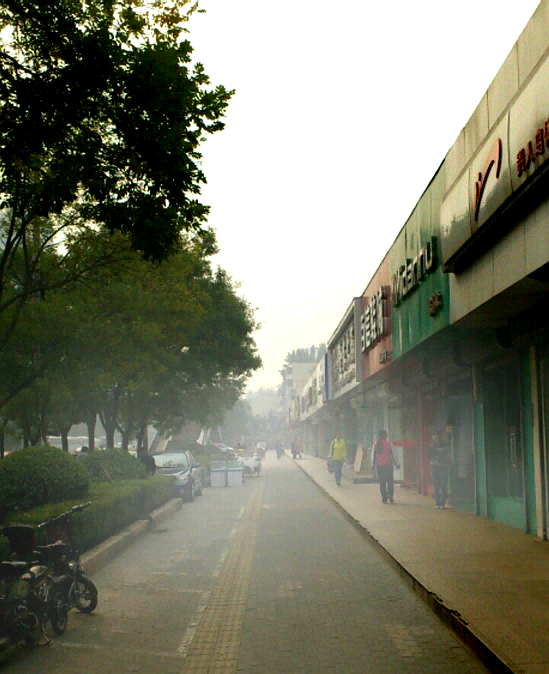} & \hspace{-0.45cm}
				\includegraphics[width = 0.09\textwidth,height=0.07\textheight]{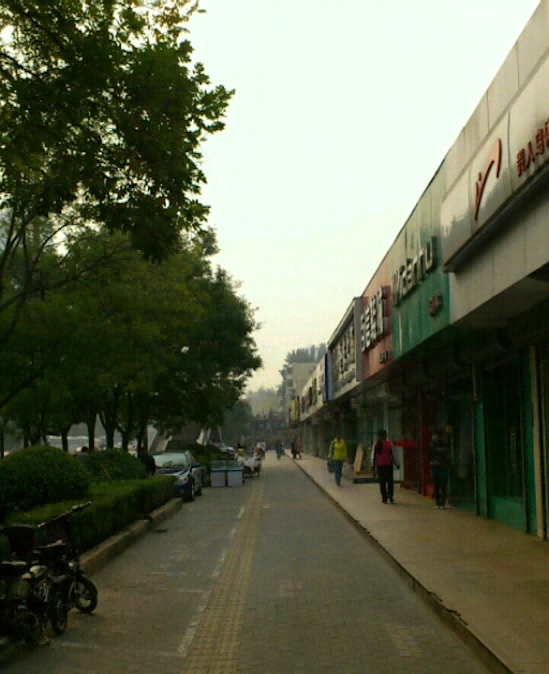} & \hspace{-0.45cm}
				\includegraphics[width = 0.09\textwidth,height=0.07\textheight]{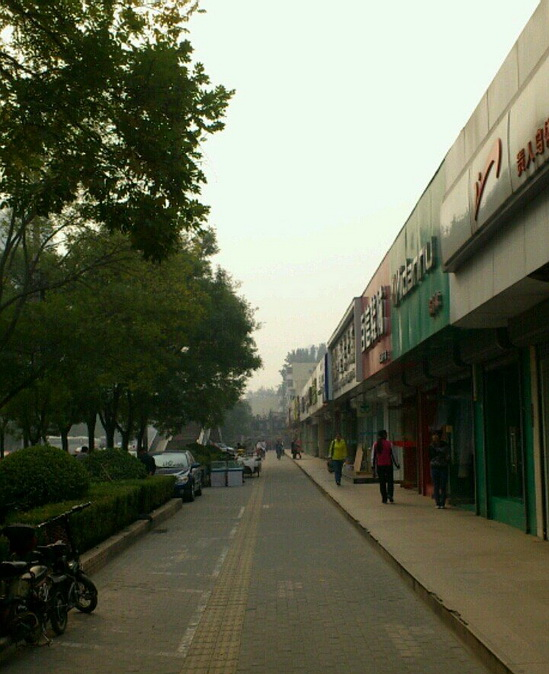} \\
				
				10.739/0.644
				& \hspace{-0.45cm} 17.196/0.679
				& \hspace{-0.45cm} 13.407/0.718
				& \hspace{-0.45cm} 21.202/0.850
				& \hspace{-0.45cm} 15.906/0.749
				& \hspace{-0.45cm} \textbf{30.516/0.955}
				& \hspace{-0.45cm} $+\infty$/1\\
				
				(h) PSD \cite{chen2021psd}& \hspace{-0.45cm}
				(i) Cycle-dehaze \cite{cycle-dehazing}& \hspace{-0.45cm}
				(j) IDE \cite{IDE2021ide}& \hspace{-0.45cm}
				(k) RefineDNet \cite{zhao2021refinednet}& \hspace{-0.45cm}
				(l) $D^{4}$ \cite{yang2022self}& \hspace{-0.45cm}
				(m) QPC-Net& \hspace{-0.45cm}
				(n) GT
				
		\end{tabular}}
	\end{center}
	\vspace{-0.5cm}
	\caption{Qualitative comparison between the proposed QPC-Net and the state-of-the-art methods on the SOTS dataset.}
	\vspace{-0.2cm}
	\label{synthesis1_dehaze}
\end{figure*}
\begin{table}[htbp]\scriptsize
	\begin{center}
		\captionsetup{justification=centering}
		\caption{\\Q{\footnotesize UANTITATIVE} PSNR {\footnotesize AND} SSIM R{\footnotesize ESULTS} {\footnotesize ON} SOTS O{\footnotesize UTDOOR} D{\footnotesize ATASET} U{\footnotesize SING} D{\footnotesize IFFERENT} C{\footnotesize ONFIGURATIONS}}
		\resizebox{0.85\columnwidth}{!}{
			\begin{tabular}{ccc}
				\toprule
				& PSNR \cite{huynh2008scope}  & SSIM \cite{wang2004image} \\
				\midrule
				CycleGAN \cite{zhu2017unpaired} & 14.7115 & 0.5024 \\
				Cycle-dehaze \cite{cycle-dehazing} & 20.9274 & 0.7431\\
				w/o VGG  & 21.2083 & 0.9073 \\
				w/o DC & 22.5896 & 0.8795\\
				w/o ASM & 23.2111 & 0.8467\\
				w/o HAG & 25.7661 & 0.9331\\
				QPC-Net & \textbf{29.3991} & \textbf{0.9646}\\
				\bottomrule
		\end{tabular}}
		\label{ablation-tb}
		\vspace{-0.8cm}
	\end{center}
\end{table}

3) \textbf{\textit{Evaluation datasets and metrics}}

We evaluate the proposed method on two synthetic datasets (SOTS, HazeRD) and two real-world datasets (O-HAZE, LIVE). For the synthetic datasets, SOTS comes from the RESIDE dataset, which contains 500 indoor scenes and 500 outdoor scenes. HazeRD synthesized by Zhang \textit{et al.} \cite{HAZERD2017ICIP} which contains 75 synthetic outdoor foggy images of different fog concentrations. The real-world foggy dataset O-HAZE \cite{O-HAZE} used in the NTIRE 2018 image dehazing challenge. It contains 45 outdoor real-world hazy images, and the LIVE dataset provided by Choi \textit{et al.} \cite{referencelessdefogging2015TIP} which contains 500 outdoor real-world foggy images.

To evaluate the performance of our method, we utilize reference indicators and non-reference indicators as the evaluativ metrics. Like most defogging papers, the Structure Similarity (SSIM) \cite{wang2004image}, the Peak Signal to Noise Ratio (PSNR) \cite{huynh2008scope} and the Learned Perceptual Image Patch Similarity (LPIPS) \cite{zhang2018unreasonable} are usually used as criteria to evaluate the quantitative performance of defogging methods. In addition, as the corresponding ground truth of real-world foggy images is extremely difficult to available, the performance of defogged results on these real-world images was assessed by no-parameter models including Fog Aware Density Evaluator (FADE) \cite{referencelessdefogging2015TIP}, the Blind/Referenceless Image Spatial Quality Evaluator (BRISQUE) \cite{MSCN2012TIP}, Natural Image Quality Evaluator (NIQE) \cite{mittal2012making} and Bilnd Assessment based on Visibility Enhancement (BAVE) \cite{hautiere2011blind}. Among them, FADE predicts the visibility of a single foggy image to perceive the fog density. The lower the value of the FADE, the stronger the defogged ability. BRISQUE uses natural scene statistics to quantify losses of \text{"naturalness"} in the image due to the distortions. The score of the BRISQUE ranges from 0 to 100, the closer to 0 the better the defogged quality. NIQE is a space domain natural scene statistics-based quality aware collection for predicting the quality of defogged image. The lower the value of the NIQE, the better the defogged effect. For BAVE, we consider the evaluation indicator $\overline{\gamma}$, which represents the mean ratio of the gradients at visible edges. The higher the value of $\overline{\gamma}$, the better the defogged performance.

\subsection{Ablation Studies}
To demonstrate the effectiveness of our QPC-Net architecture, we perform a series of ablation analyses on the different modules of our proposed QPC-Net. We mainly consider the following four factors: 1) \textit{w/o VGG:} our full model without the VGG perceptual losses; 2) \textit{w/o DC:} our full model without the proposed densely-residual connected encoder-decoder; 3) \textit{w/o ASM:} our full model without the atmospheric scattering model; 4) \textit{w/o HAG:} our full model without the holistic attention-fusion generator. Moreover, we also compare our method to CycleGAN \cite{zhu2017unpaired} and Cycle-dehaze \cite{cycle-dehazing}.
\begin{figure*}[htp]\tiny
	\begin{center}
		\vspace{-0.2cm}
		\resizebox{\textwidth}{!}{
			\begin{tabular}{ccccccc}
				
				\includegraphics[width = 0.085\textwidth,height=0.045\textheight]{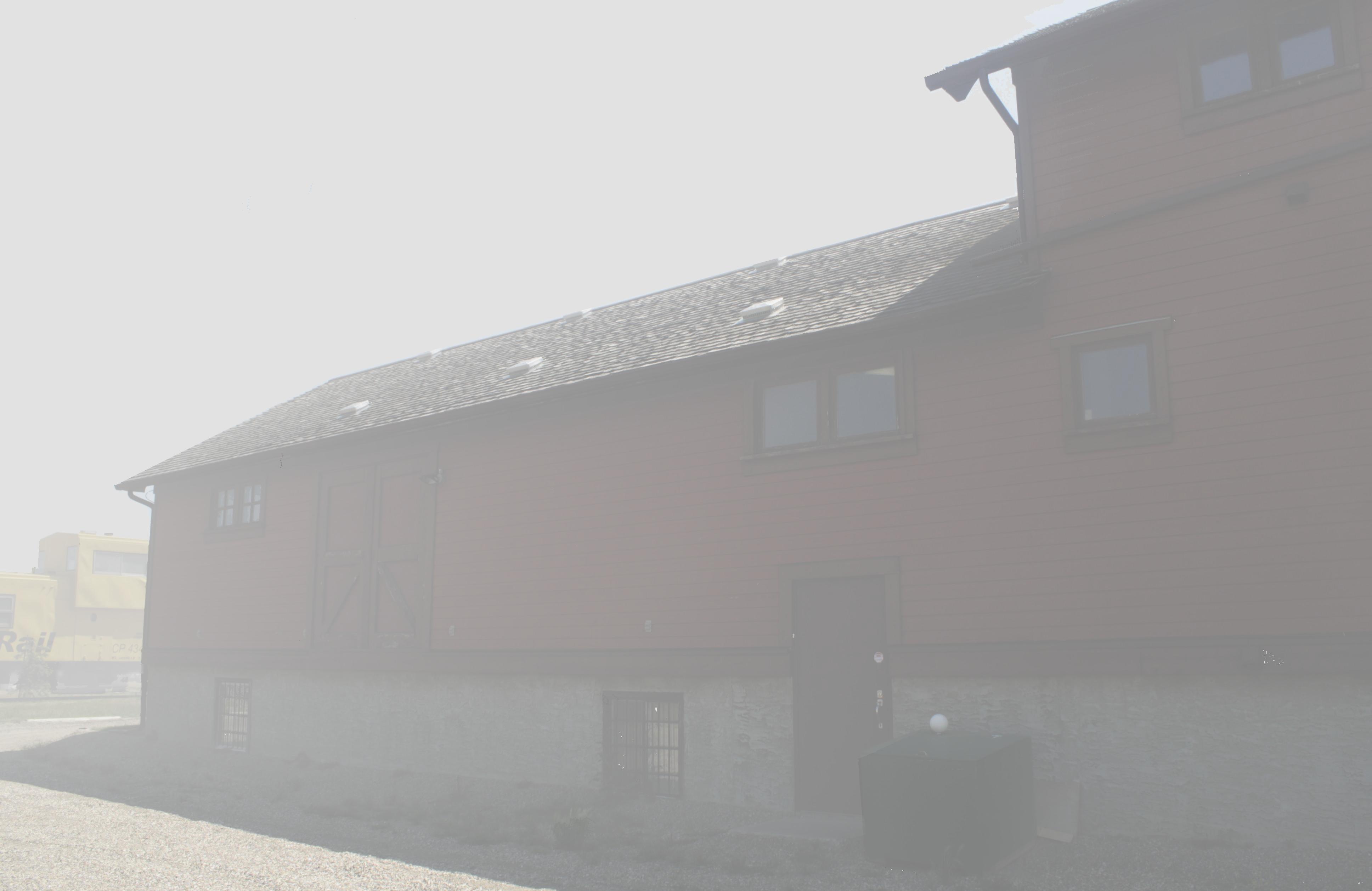} & \hspace{-0.45cm}
				\includegraphics[width = 0.085\textwidth,height=0.045\textheight]{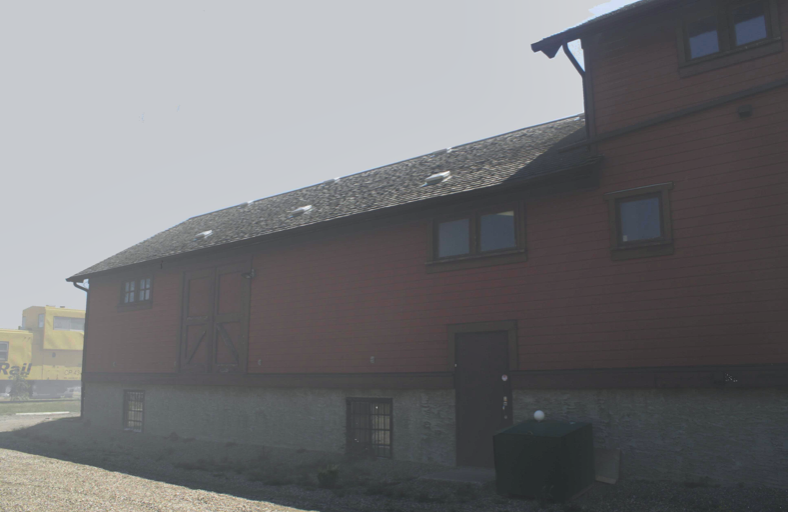} & \hspace{-0.45cm}
				\includegraphics[width = 0.085\textwidth,height=0.045\textheight]{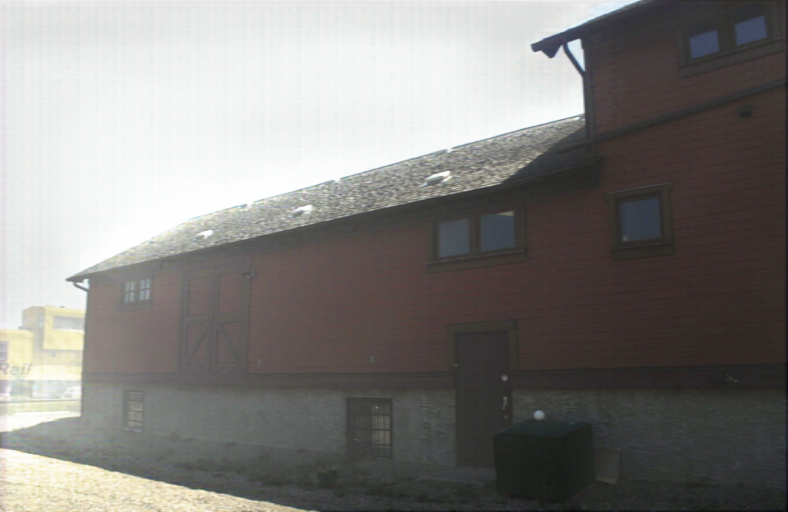} & \hspace{-0.45cm}
				\includegraphics[width = 0.085\textwidth,height=0.045\textheight]{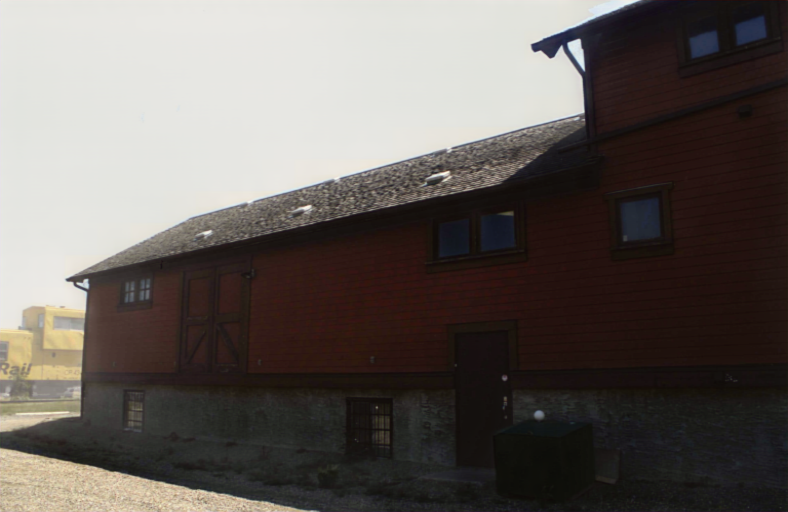} & \hspace{-0.45cm}
				\includegraphics[width = 0.085\textwidth,height=0.045\textheight]{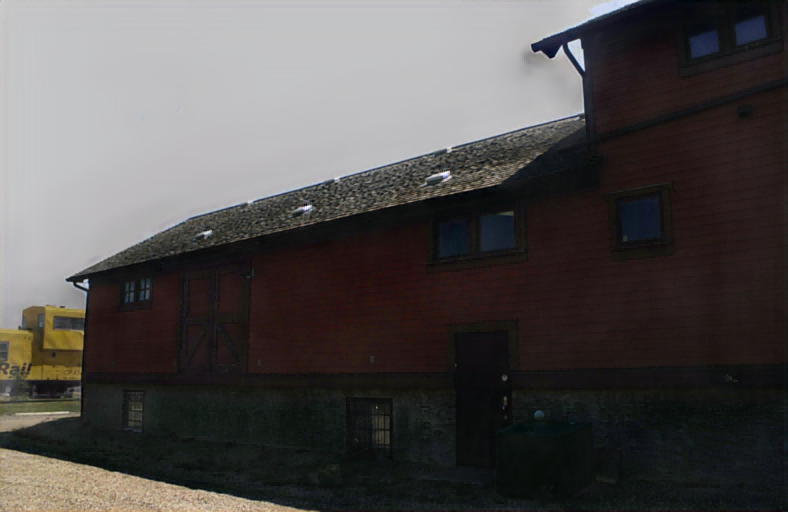} & \hspace{-0.45cm}
				\includegraphics[width = 0.085\textwidth,height=0.045\textheight]{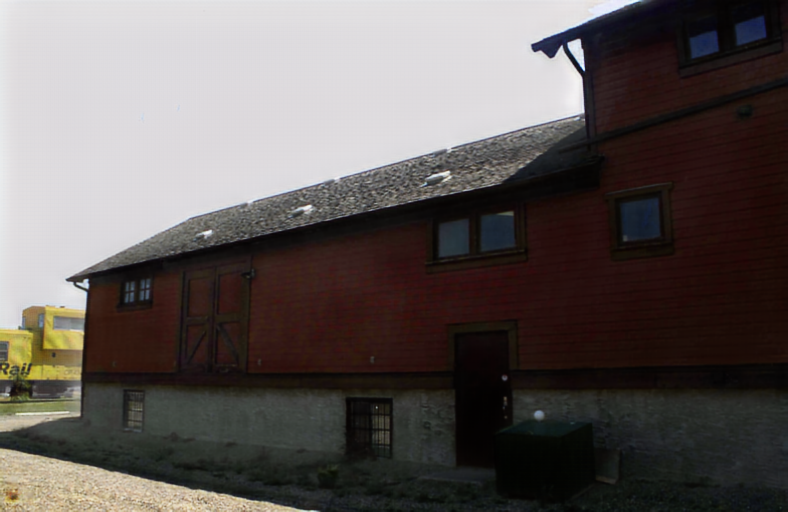} & \hspace{-0.45cm}
				\includegraphics[width = 0.085\textwidth,height=0.045\textheight]{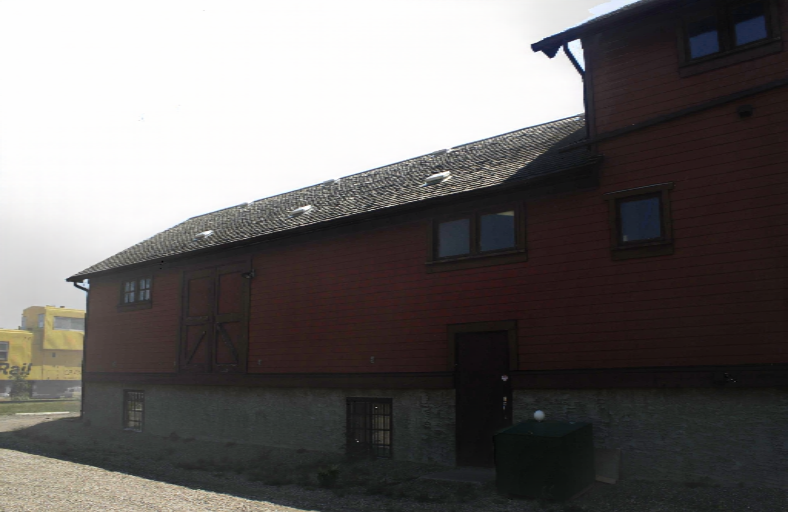} \\
				
				PSNR/SSIM
				& \hspace{-0.45cm} 21.283/0.860
				& \hspace{-0.45cm} 18.166/0.825
				& \hspace{-0.45cm} 17.512/0.712
				& \hspace{-0.45cm} 15.890/0.671
				& \hspace{-0.45cm} 17.473/0.668
				& \hspace{-0.45cm} 17.769/0.799\\
				
				\includegraphics[width = 0.085\textwidth,height=0.045\textheight]{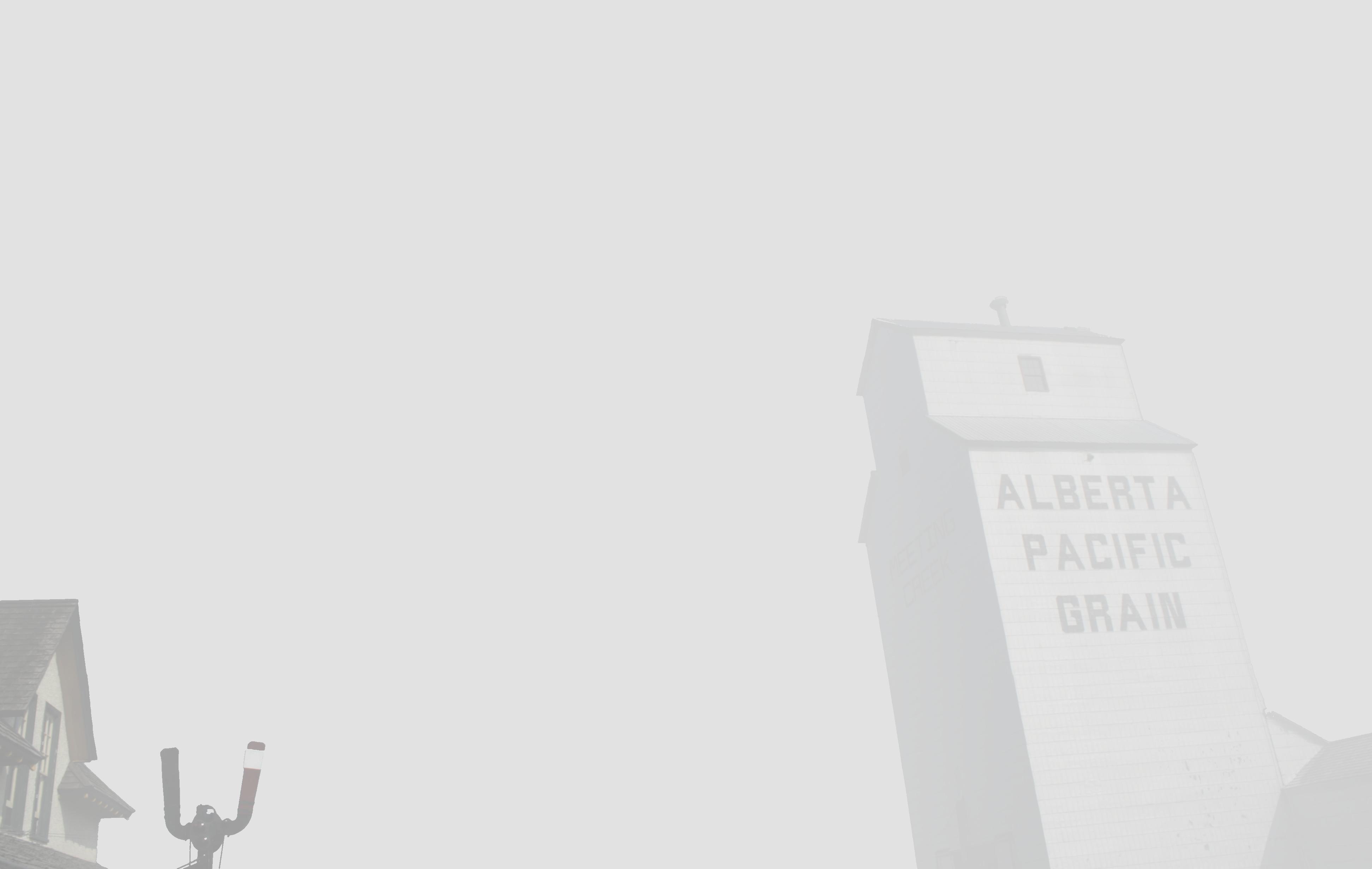} & \hspace{-0.45cm}
				\includegraphics[width = 0.085\textwidth,height=0.045\textheight]{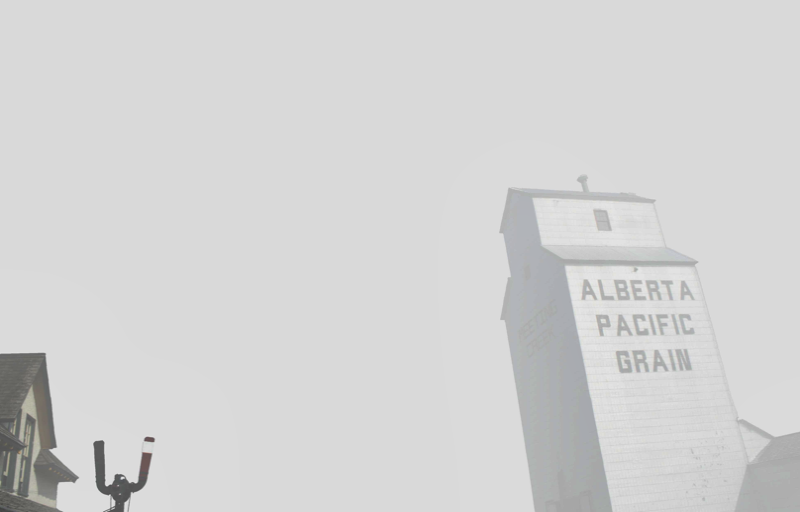} & \hspace{-0.45cm}
				\includegraphics[width = 0.085\textwidth,height=0.045\textheight]{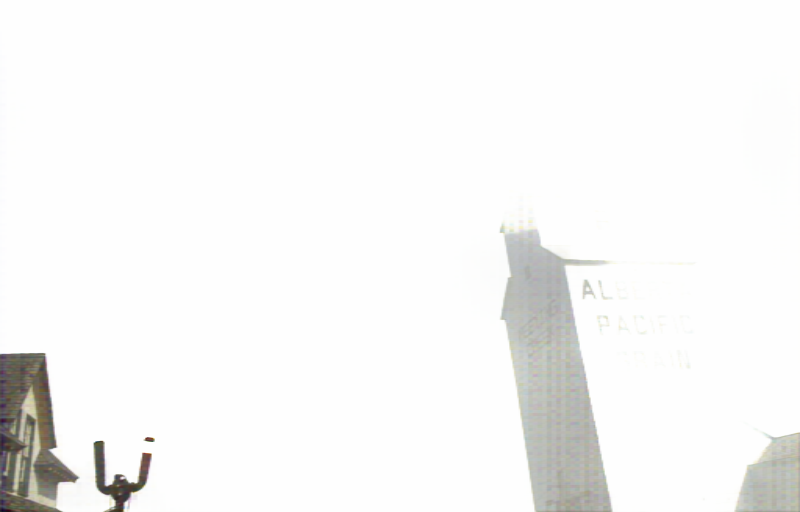} & \hspace{-0.45cm}
				\includegraphics[width = 0.085\textwidth,height=0.045\textheight]{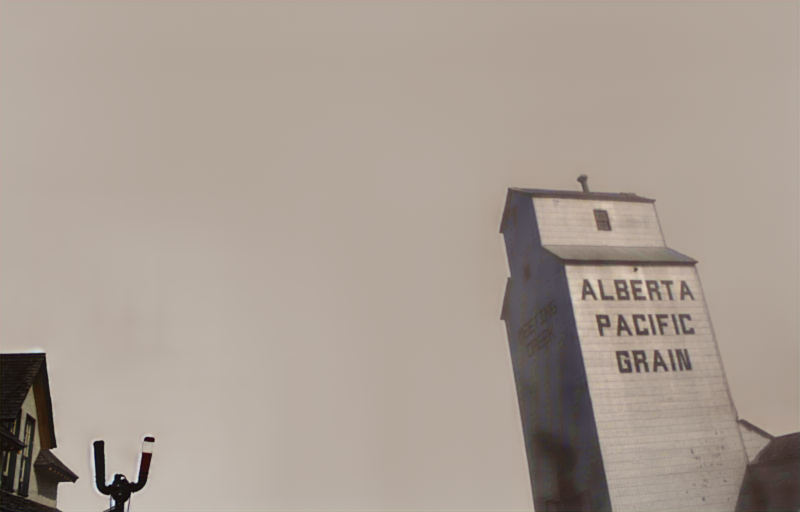} & \hspace{-0.45cm}
				\includegraphics[width = 0.085\textwidth,height=0.045\textheight]{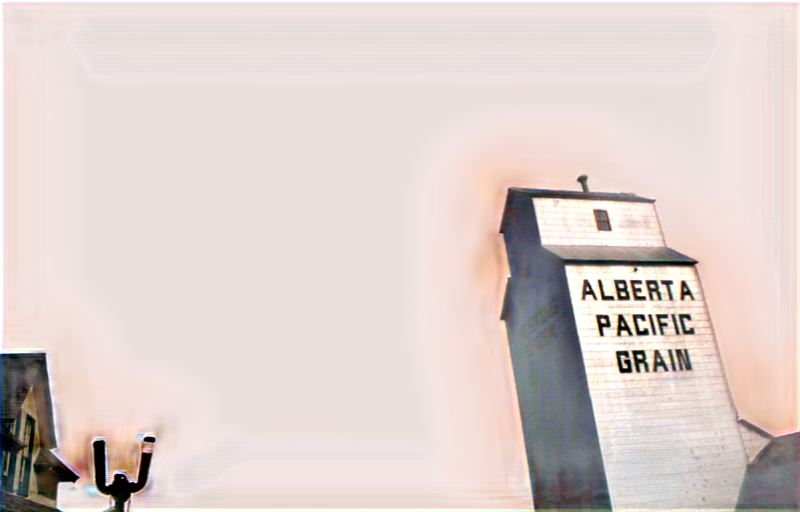} & \hspace{-0.45cm}
				\includegraphics[width = 0.085\textwidth,height=0.045\textheight]{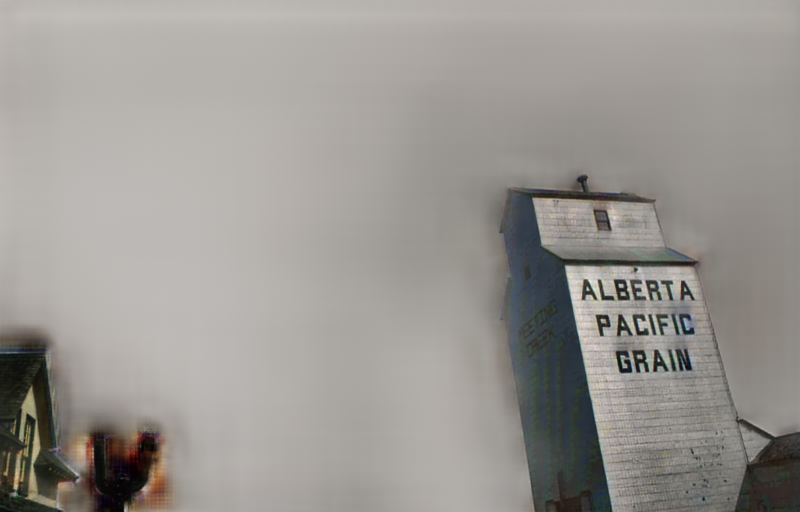} & \hspace{-0.45cm}
				\includegraphics[width = 0.085\textwidth,height=0.045\textheight]{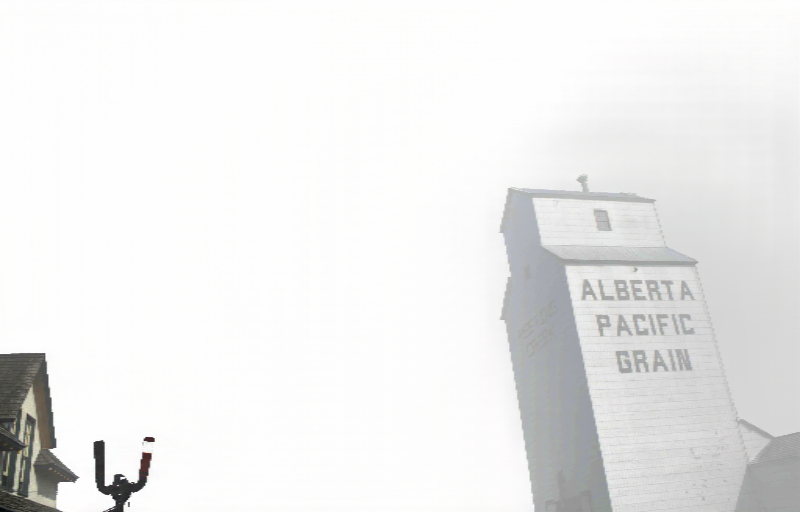} \\
				
				PSNR/SSIM
				& \hspace{-0.45cm} 10.755/0.855
				& \hspace{-0.45cm} 7.562/0.767
				& \hspace{-0.45cm} 15.384/0.902
				& \hspace{-0.45cm} 9.819/0.794
				& \hspace{-0.45cm} 14.595/0.874
				& \hspace{-0.45cm} 7.994/0.818\\
				
				(a) Input & \hspace{-0.45cm}
				(b) DehazeNet \cite{dehazenet2016TIP} & \hspace{-0.45cm}
				(c) DCPDN \cite{zhang2018densely} & \hspace{-0.45cm}
				(d) EPDN \cite{pix2pixdehazing} & \hspace{-0.45cm}
				(e) GCANet \cite{chen2019gated}& \hspace{-0.45cm}
				(f) DA-Dehaze \cite{shao2020domain}& \hspace{-0.45cm}
				(g) MSBDN \cite{dong2020multi}\\

				\includegraphics[width = 0.085\textwidth,height=0.045\textheight]{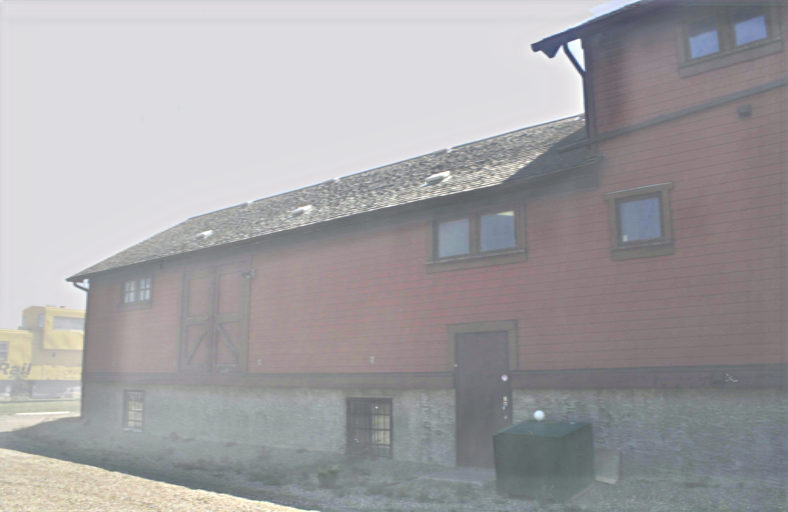} & \hspace{-0.45cm}
				\includegraphics[width = 0.085\textwidth,height=0.045\textheight]{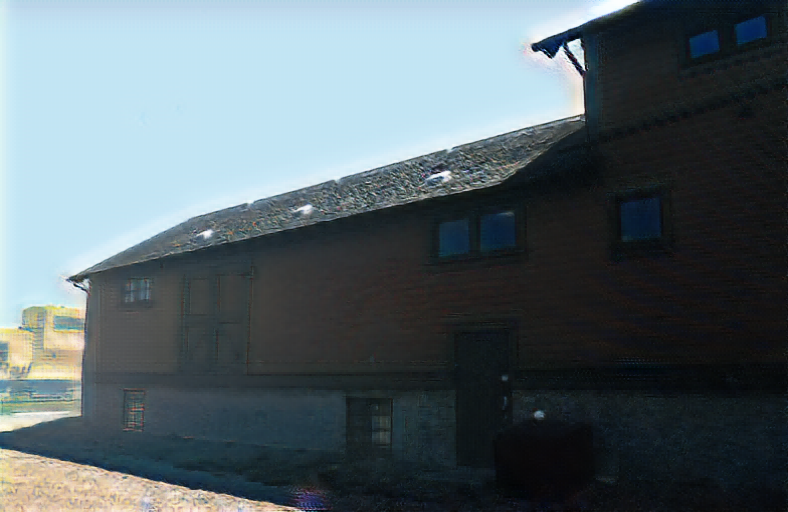} & \hspace{-0.45cm}
				\includegraphics[width = 0.085\textwidth,height=0.045\textheight]{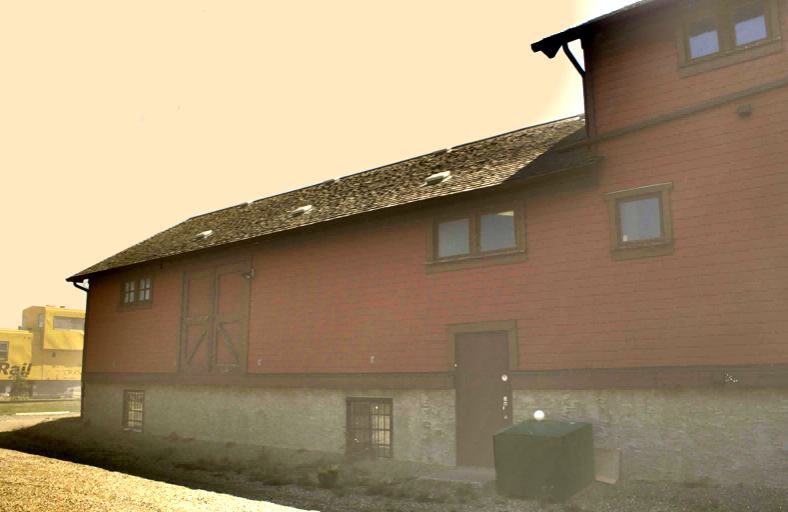} & \hspace{-0.45cm}
				\includegraphics[width = 0.085\textwidth,height=0.045\textheight]{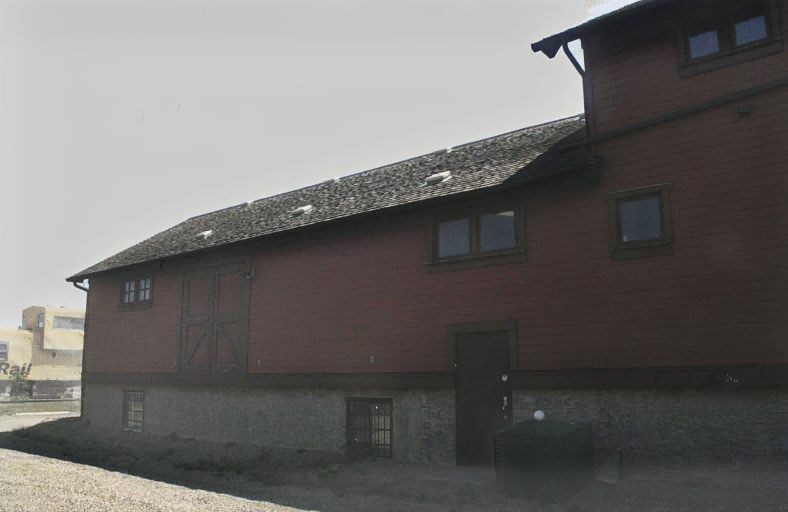} & \hspace{-0.45cm}
				\includegraphics[width = 0.085\textwidth,height=0.045\textheight]{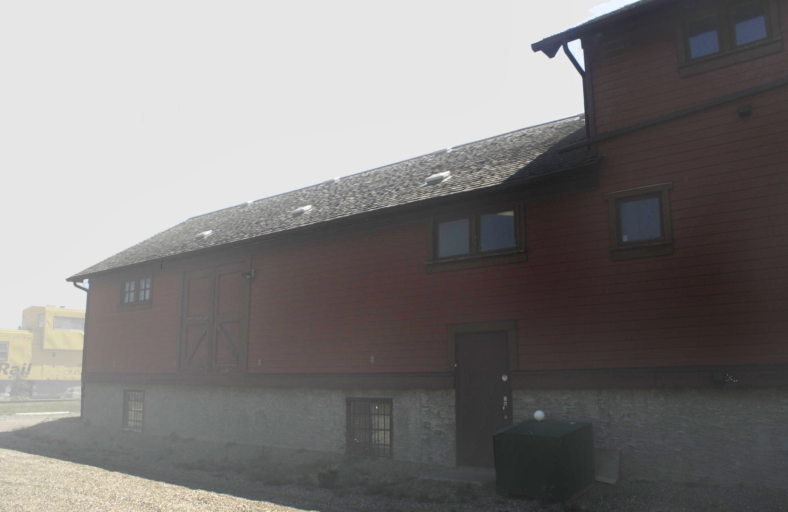} & \hspace{-0.45cm}
				\includegraphics[width = 0.085\textwidth,height=0.045\textheight]{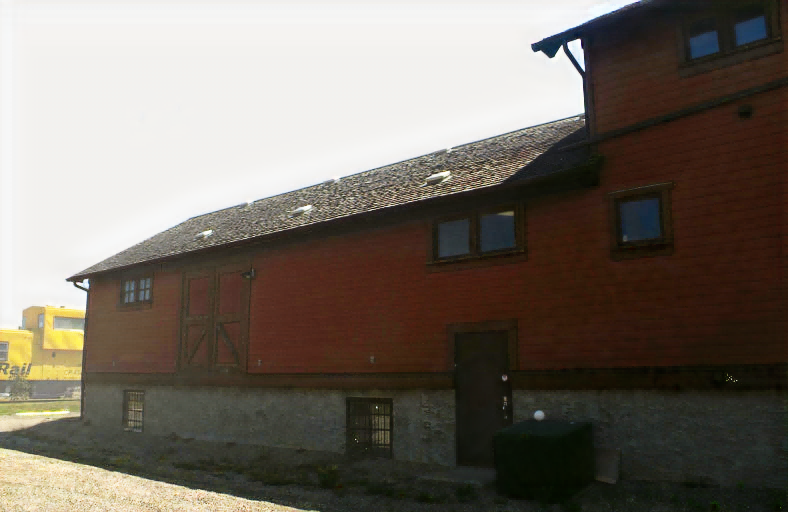} & \hspace{-0.45cm}
				\includegraphics[width = 0.085\textwidth,height=0.045\textheight]{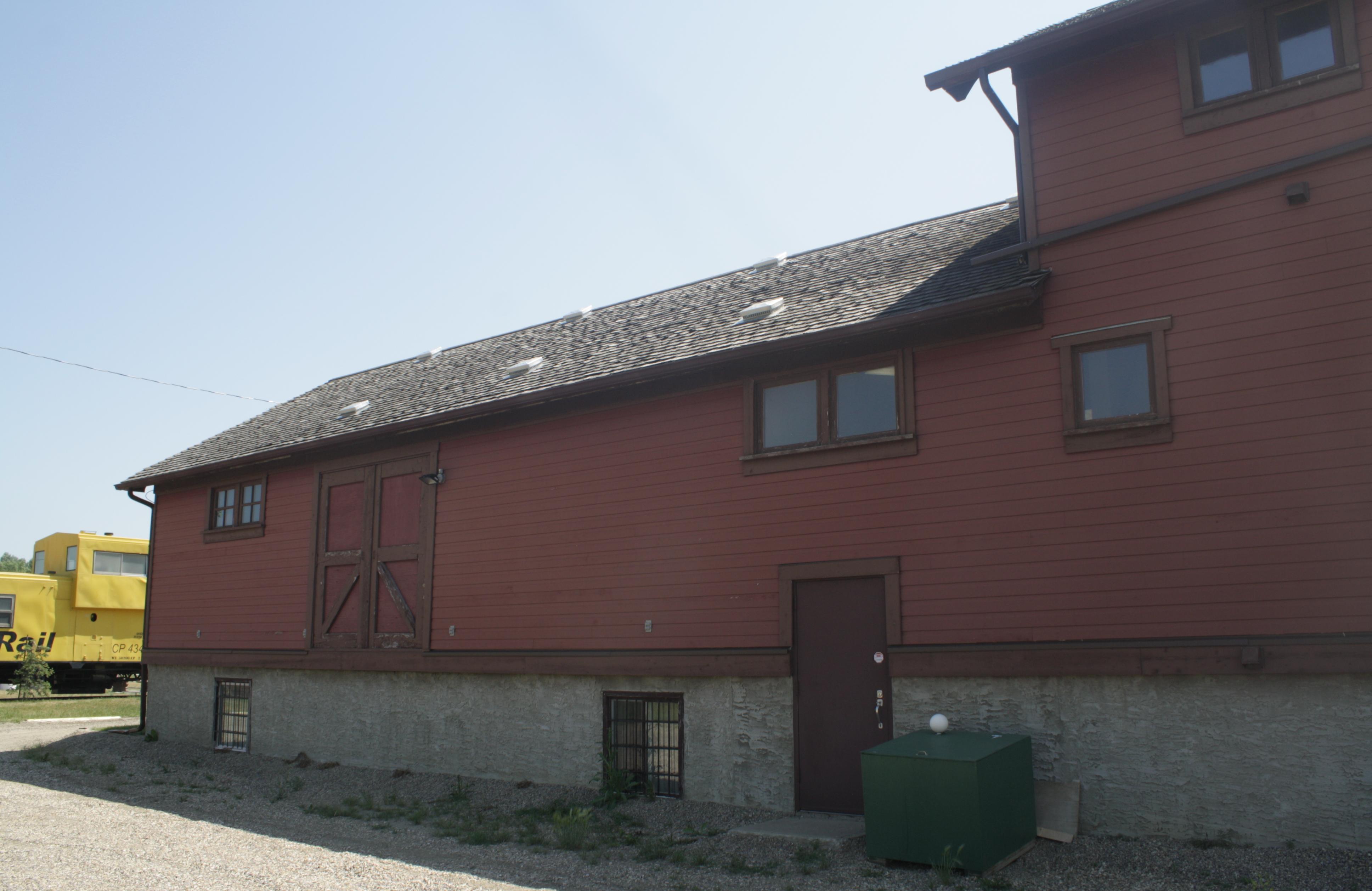} \\
				
				12.717/0.755
				& \hspace{-0.45cm} 17.771/0.655
				& \hspace{-0.45cm} 15.198/0.823
				& \hspace{-0.45cm} 21.295/\textbf{0.876}
				& \hspace{-0.45cm} 19.779/0.849
				& \hspace{-0.45cm} \textbf{21.913}/0.861
				& \hspace{-0.45cm} $+\infty$/1\\
				
				\includegraphics[width = 0.085\textwidth,height=0.045\textheight]{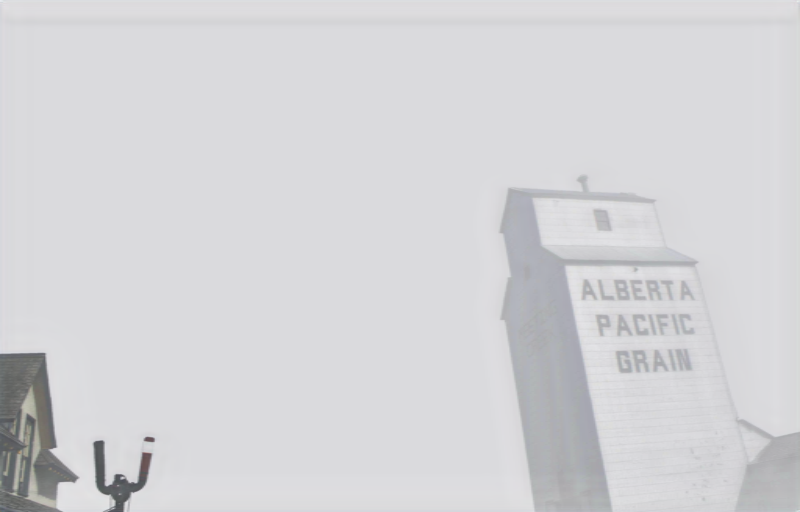} & \hspace{-0.45cm}
				\includegraphics[width = 0.085\textwidth,height=0.045\textheight]{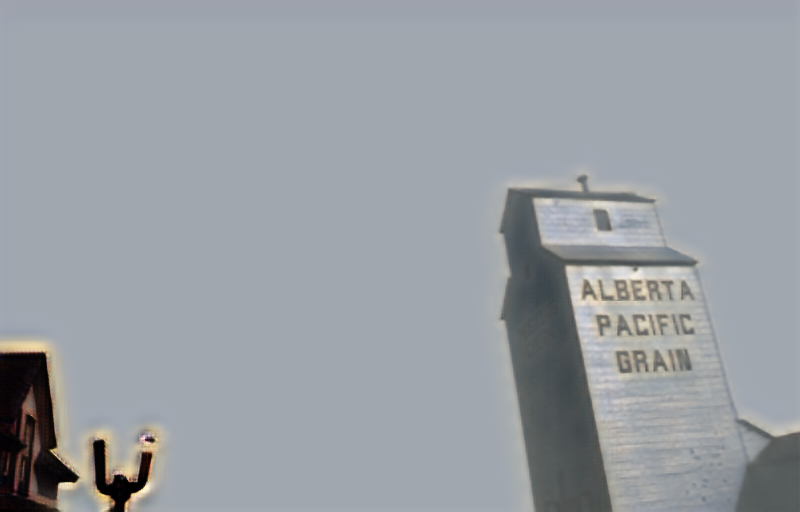} & \hspace{-0.45cm}
				\includegraphics[width = 0.085\textwidth,height=0.045\textheight]{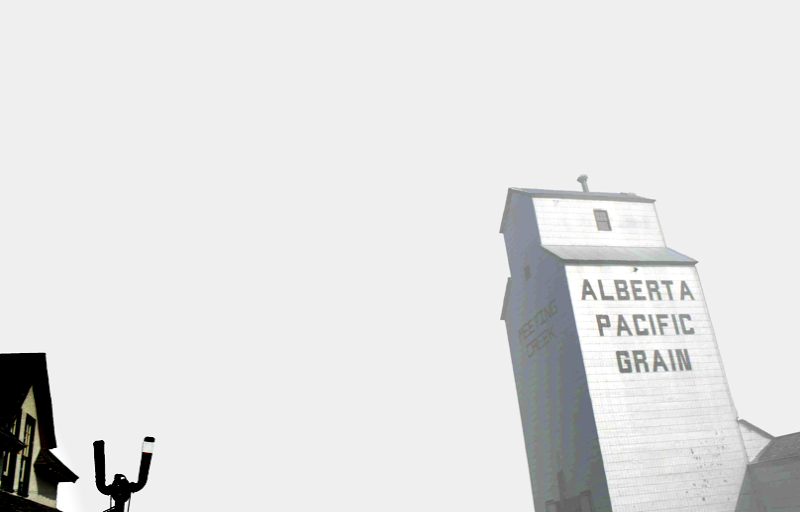} & \hspace{-0.45cm}
				\includegraphics[width = 0.085\textwidth,height=0.045\textheight]{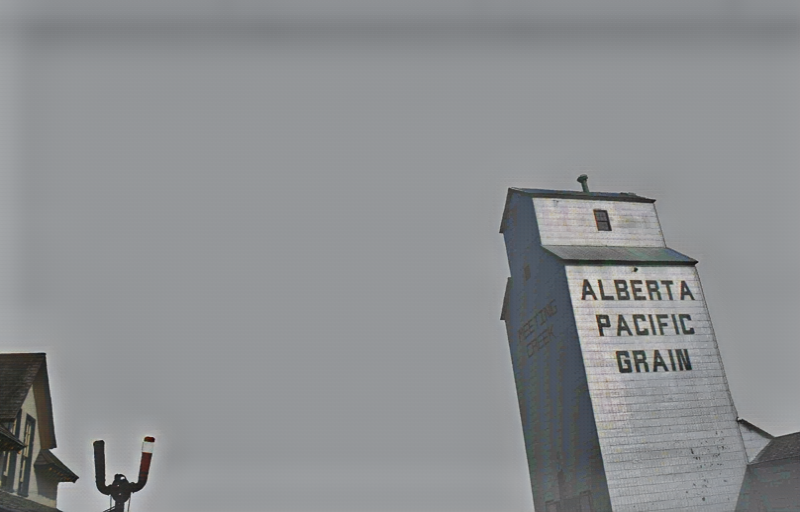} & \hspace{-0.45cm}
				\includegraphics[width = 0.085\textwidth,height=0.045\textheight]{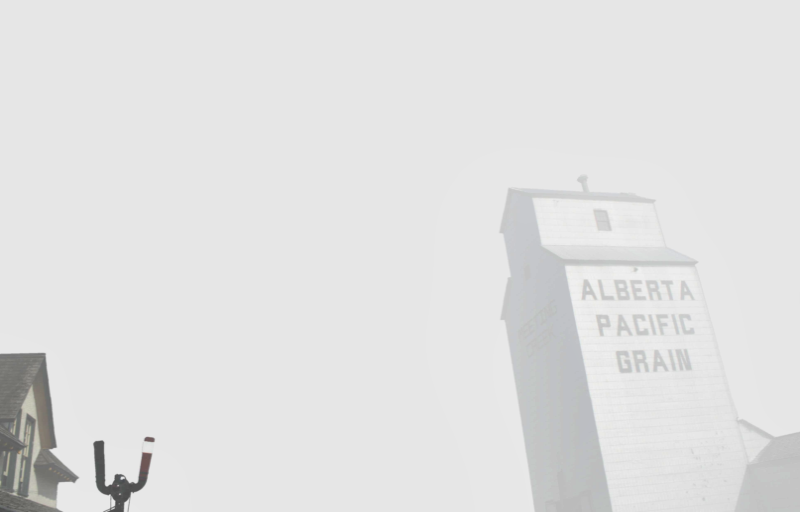} & \hspace{-0.45cm}
				\includegraphics[width = 0.085\textwidth,height=0.045\textheight]{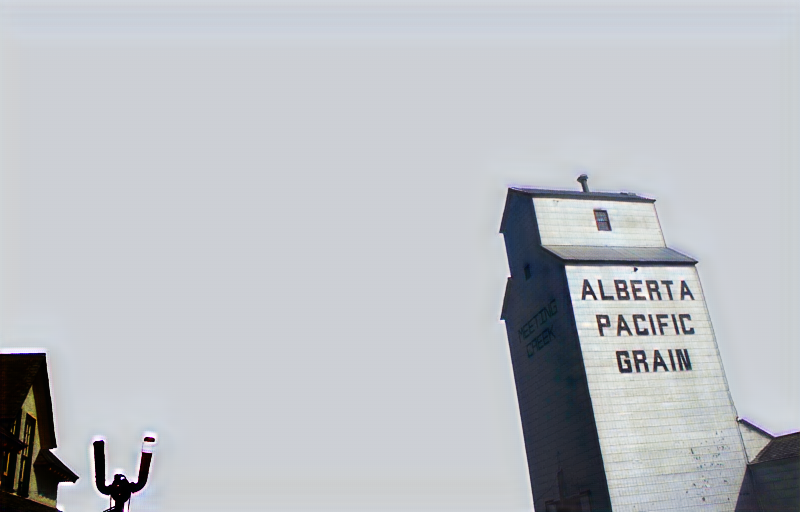} & \hspace{-0.45cm}
				\includegraphics[width = 0.085\textwidth,height=0.045\textheight]{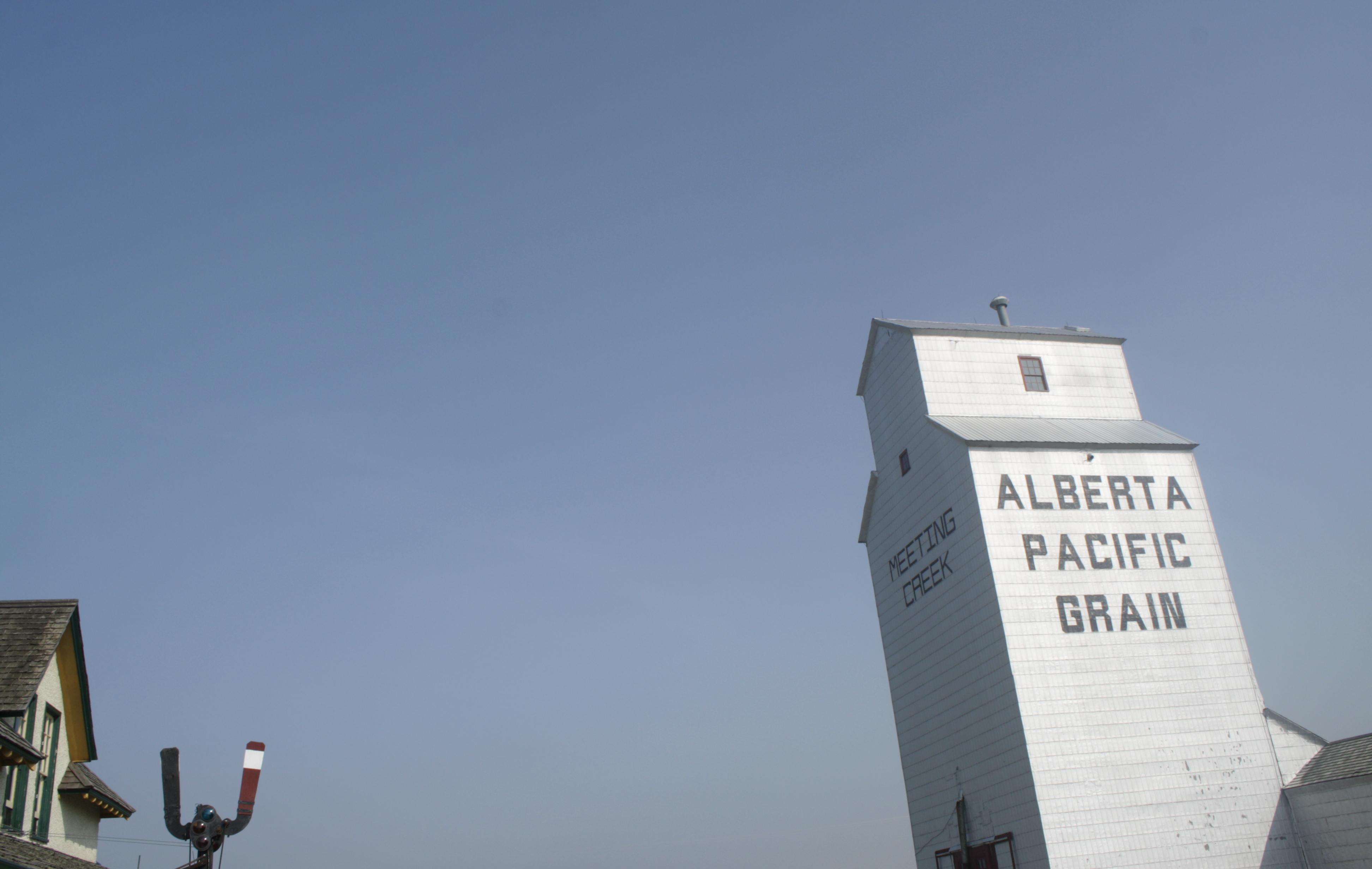} \\
				
				10.574/0.838
				& \hspace{-0.45cm} 17.269/0.846
				& \hspace{-0.45cm} 8.999/0.837
				& \hspace{-0.45cm} 18.466/0.891
				& \hspace{-0.45cm} 9.407/0.825
				& \hspace{-0.45cm} \textbf{19.008/0.901}
				& \hspace{-0.45cm} $+\infty$/1\\
				
				(h) PSD \cite{chen2021psd}& \hspace{-0.45cm}
				(i) Cycle-dehaze \cite{cycle-dehazing}& \hspace{-0.45cm}
				(j) IDE \cite{IDE2021ide}& \hspace{-0.45cm}
				(k) RefineDNet \cite{zhao2021refinednet}& \hspace{-0.45cm}
				(l) $D^{4}$ \cite{yang2022self}& \hspace{-0.45cm}
				(m) QPC-Net& \hspace{-0.45cm}
				(n) GT
				
		\end{tabular}}
	\end{center}
	\vspace{-0.5cm}
	\caption{Qualitative comparison between the proposed QPC-Net and the state-of-the-art methods on the HazeRD dataset.}
	\vspace{-0.4cm}
	\label{synthesis2_dehaze}
\end{figure*}

The qualitative and quantitative results are shown in Fig. \ref{ablation_dehaze} and Table \ref{ablation-tb}, it demonstrates that QPC-Net achieves the best defogging performance in terms of  visual effect and quantitative accuracy. As can be seen in Fig. \ref{ablation_dehaze}(b), the CycleGAN's results have the worst defogging performance, which have the serious color distortion and the texture details are completely lost. Compared to CycleGAN, we can see from Fig. \ref{ablation_dehaze}(c) that the texture information of the Cycle-dehaze's results are restored better due to the VGG perceptual loss can preserve more image content. 
\begin{table*}[hbp]
	\renewcommand\arraystretch{1.2}
	\begin{center}
		\vspace{-0.2cm}
		\captionsetup{justification=centering}
		\caption{\\A{\footnotesize VERAGE} PSNR, SSIM, LPIPS, FADE, BRISQUE, NIQE, $\overline{\gamma}$ {\footnotesize OF} D{\footnotesize EFOGGED} R{\footnotesize ESULTS} {\footnotesize ON} S{\footnotesize YNTHETIC} D{\footnotesize ATASETS} SOTS {\footnotesize AND} HazeRD. T{\footnotesize HE} T{\footnotesize OP} T{\footnotesize WO} P{\footnotesize ERFORMANCE} V{\footnotesize ALUES} A{\footnotesize RE} H{\footnotesize IGHLIGHTED} {\footnotesize IN} R{\footnotesize ED} {\footnotesize AND} B{\footnotesize LUE}}
		\resizebox{\textwidth}{2.3cm}{
			\begin{tabular}{c|c|ccccccc|ccccccc}
				\hline
				&Dataset & \multicolumn{7}{c|}{SOTS \cite{RESIDE}} & \multicolumn{7}{c}{HazeRD \cite{HAZERD2017ICIP}}\\
				
				\cline{2-16}
				&\diagbox{Methods}{Metric} & PSNR \cite{huynh2008scope} & SSIM \cite{wang2004image} & LPIPS \cite{zhang2018unreasonable}  & FADE \cite{referencelessdefogging2015TIP} & BRISQUE \cite{MSCN2012TIP} & NIQE \cite{mittal2012making} & $\overline{\gamma}$ \cite{hautiere2011blind} &  PSNR \cite{huynh2008scope} & SSIM \cite{wang2004image} & LPIPS \cite{zhang2018unreasonable}  & FADE \cite{referencelessdefogging2015TIP} & BRISQUE \cite{MSCN2012TIP} & NIQE \cite{mittal2012making} & $\overline{\gamma}$ \cite{hautiere2011blind} \\
				
				\hline
				\multirow{6}{*}{\rotatebox{90}{Paired}}
				&DehazeNet \cite{dehazenet2016TIP} &  23.6907 & 0.8698 & 0.0415 & 0.6400 & 13.5029 & 3.3586 & 1.4635 &  15.3478 & 0.7956 & 0.1522 & 1.3983 & 28.6569 & 4.4703 & 1.6570 \\
				
				&DCPDN \cite{zhang2018densely} &  21.0540 & 0.8716 & 0.0718 & 0.9340 & 21.8789 & 3.8303 &  1.3136&  14.3647 & 0.7961 & 0.2094  & 1.5799 & 25.9806 & 5.8394 & 2.3011 \\
				
				&EPDN \cite{pix2pixdehazing} & 22.0715 & 0.8376 & 0.0524 & 0.5374 & \textcolor[rgb]{0,0,1}{10.9584} & 3.6732 & 1.5170 &  15.6827 & 0.7925 & 0.1335  & 1.0842 & 19.1310 & 4.1340 & 2.0839\\

				&GCANet \cite{chen2019gated} &  28.1412 & 0.9426 & 0.0623 & 0.6375 & 15.8146 & 3.5688 & 1.6159 &  15.5987 & \textcolor[rgb]{0,0,1}{0.8256} & \textcolor{blue}{0.1325}  & 0.8668 & \textcolor[rgb]{0,0,1}{14.5558} & \textcolor[rgb]{0,0,1}{3.7155} & 2.2867\\
				
				&DA-Dehaze \cite{shao2020domain} & 27.4409 & 0.9457 & 0.0430 & 0.6296 & 11.3146 & 3.3265 & 1.6264 &  16.7240 & 0.8234 & 0.1513 & 0.8533 & 24.6961 & 4.0871 & 2.3394\\
				
				&MSBDN \cite{dong2020multi} &  \textcolor[rgb]{1,0,0}{32.4001} & \textcolor[rgb]{0,0,1}{0.9580} & \textcolor[rgb]{1,0,0}{0.0161} & 0.6438 & 17.3982 & 3.4035 & 1.5959 &  15.1064 & 0.7445 & 0.1707 & 1.2605 & 27.4725 & 5.0339 & 2.2473\\
				
				&PSD \cite{chen2021psd}&  18.1844 & 0.8499 & 0.0921 & 1.1286 & 14.4841 & 3.6480 & \textcolor{red}{3.0831} &  14.4837 & 0.7388 & 0.1963 & 1.7202 & 21.9477 & 4.1625 & 2.5439\\
				
				\hline
				\multirow{5}{*}{\rotatebox{90}{W/o Paired}}
				&Cycle-dehaze \cite{cycle-dehazing} &  20.9274 & 0.7431 & 0.1770 & 0.6031 & 18.6996 & 3.0992 & 1.4073 &  13.6825 & 0.6792 & 0.2277 & 1.535 & 23.3122 & 4.1273 & 1.4616 \\
				
				&IDE \cite{IDE2021ide} &  17.4091 & 0.7853 & 0.9635 & 0.5901 & 13.4620 & 3.0655 & \textcolor{blue}{2.3944} &  14.2400 & 0.8161 & 0.1445 & \textcolor[rgb]{1,0,0}{0.5981} & 30.4148 & 4.3759 & \textcolor[rgb]{1,0,0}{3.2809} \\

				&RefineDNet \cite{zhao2021refinednet} &  24.2836 & 0.9053 & 0.0486 & 0.7179 & 11.3096 & \textcolor{blue}{2.9106} & 1.6721 &  \textcolor{blue}{17.0230} & 0.7891 & \textcolor{red}{0.1262} & 1.1356 & 17.5284 & 3.7186 & 2.3499 \\
				
				&$D^{4}$ \cite{yang2022self} &  25.8069 & 0.9214 & 0.1087 & \textcolor{blue}{0.5365} & 30.8489 & 4.1004 & 1.4108 &  12.8934 & 0.6067 & 0.2245 & 1.5477 & 27.3635 & 5.0726 & 1.6238 \\
				
				&QPC-Net &  \textcolor[rgb]{0,0,1}{29.3991} & \textcolor[rgb]{1,0,0}{0.9646} & \textcolor[rgb]{0,0,1}{0.0392} & \textcolor{red}{0.5179} & \textcolor[rgb]{1,0,0}{10.9396} & \textcolor{red}{2.8984} & 1.6809 &  \textcolor[rgb]{1,0,0}{17.2106} & \textcolor[rgb]{1,0,0}{0.8274} & 0.1413 & \textcolor[rgb]{0,0,1}{0.8215} & \textcolor[rgb]{1,0,0}{14.0798} & \textcolor[rgb]{1,0,0}{3.6820} & \textcolor[rgb]{0,0,1}{2.5805}\\
				
				\hline
		\end{tabular}}
		\label{synthesis-tb}
		\vspace{-0.4cm}
	\end{center}
\end{table*}
However, as shown in the zoom-in regions in Fig. \ref{ablation_dehaze}(c), the color board (1st row) and the sign (2nd row) are blurred. Fig. \ref{ablation_dehaze}(d) shows the results obtained by using QPC-Net without VGG perceptual losses. We can see that there is a significant amount of fog in the defogged results and the sign is also blurred. In addition, without using the proposed DC module, there is a serious chessboard phenomenon in some regions (\textit{e.g.}, the color board and the sign in Fig. \ref{ablation_dehaze}(e)). It demonstrates that the proposed DC module could help the network preserve richer image content and texture details. As shown in Fig. \ref{ablation_dehaze}(f), the defogged results generated by our QPC-Net without ASM module have color distortion phenomenon. The reason is due to the atmospheric scattering model can guide the synthesizing network to generate a more natural foggy image to improve the mapping ability from foggy image domain to fogfree image domain.
\begin{figure*}[ht]\footnotesize
	\vspace{-0.2cm}
	\begin{center}
		\begin{tabular}{@{}ccccccc@{}}
			
			\includegraphics[width = 0.135\textwidth,height=0.075\textheight]{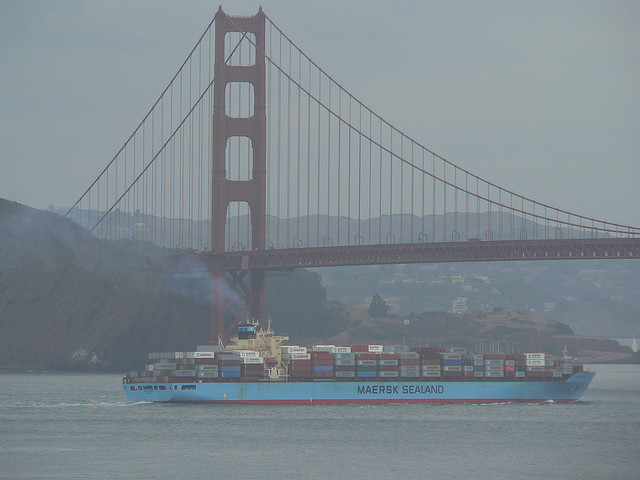}& \hspace{-0.45cm}
			\includegraphics[width = 0.135\textwidth,height=0.075\textheight]{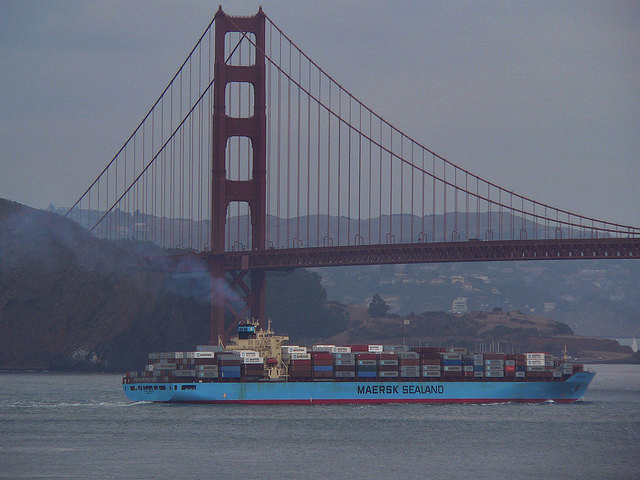} & \hspace{-0.45cm}
			\includegraphics[width = 0.135\textwidth,height=0.075\textheight]{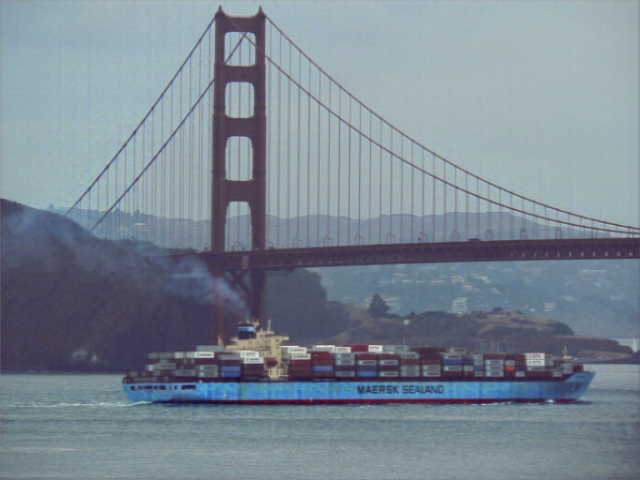} & \hspace{-0.45cm}
			\includegraphics[width = 0.135\textwidth,height=0.075\textheight]{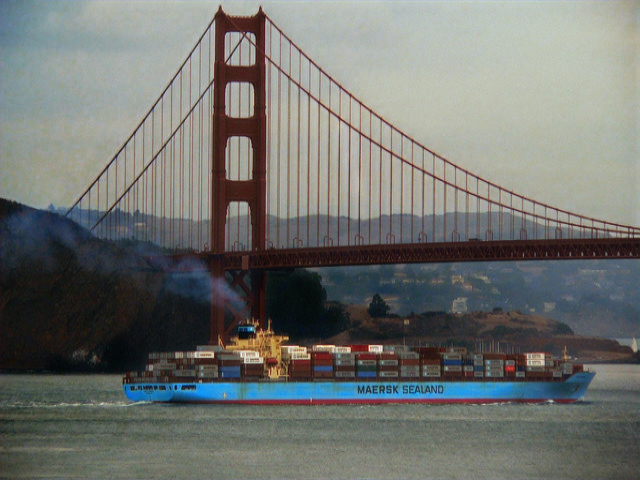} & \hspace{-0.45cm}
			\includegraphics[width = 0.135\textwidth,height=0.075\textheight]{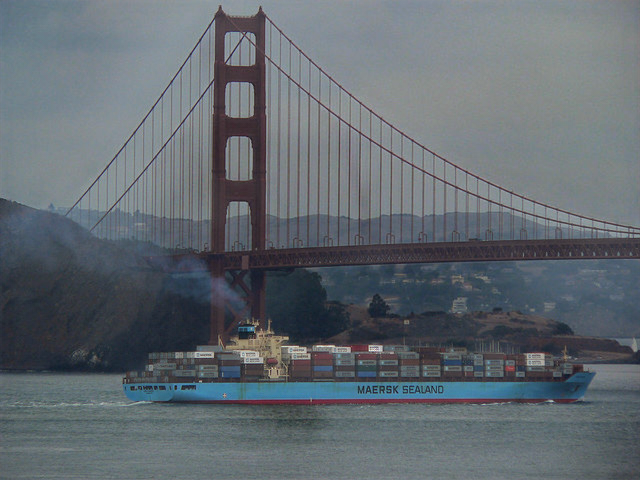} & \hspace{-0.45cm}
			\includegraphics[width = 0.135\textwidth,height=0.075\textheight]{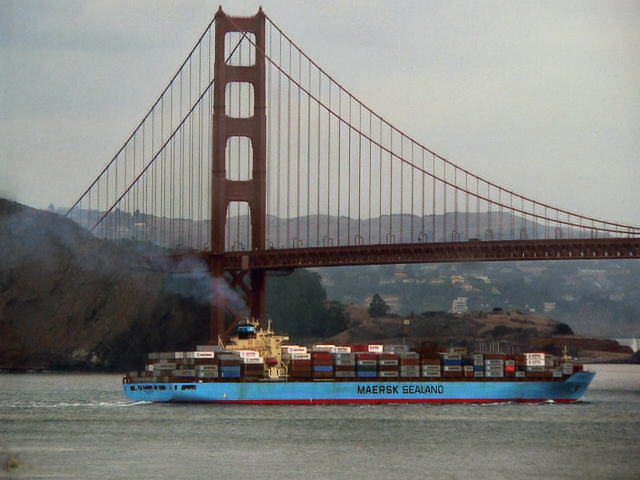} & \hspace{-0.45cm}
			\includegraphics[width = 0.135\textwidth,height=0.075\textheight]{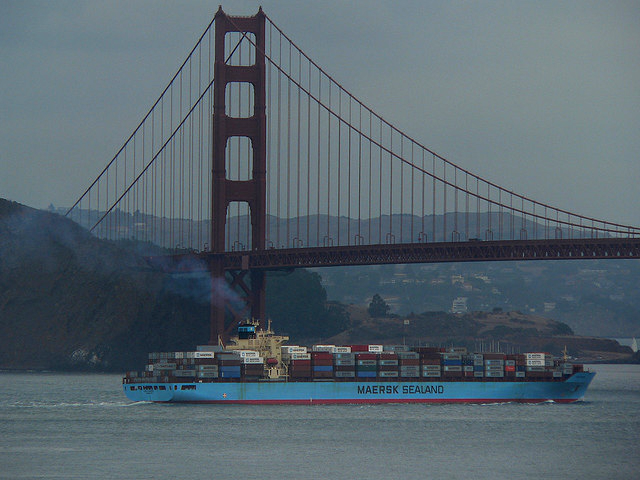} \\
			
			\includegraphics[width = 0.135\textwidth,height=0.075\textheight]{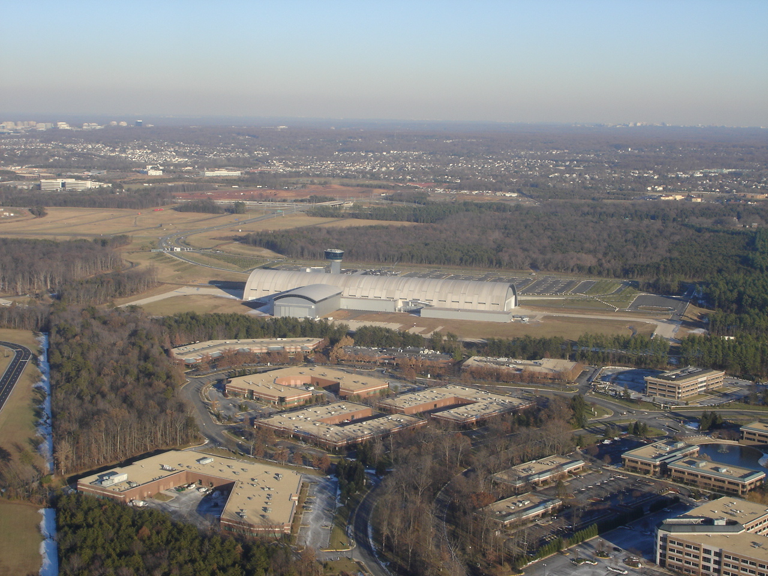} & \hspace{-0.45cm}
			\includegraphics[width = 0.135\textwidth,height=0.075\textheight]{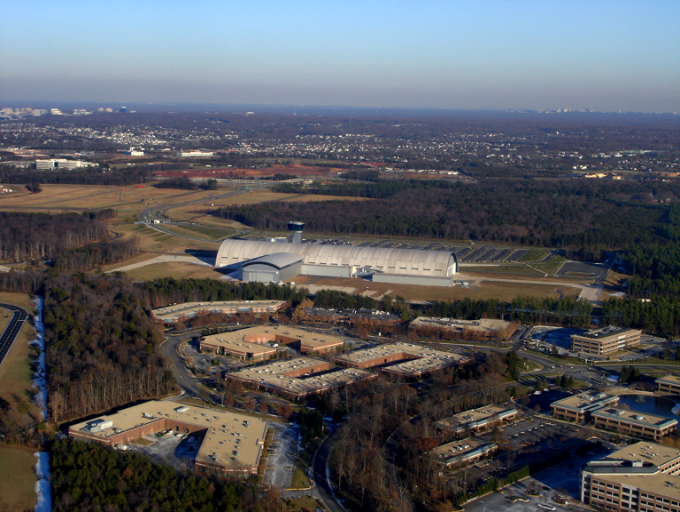} & \hspace{-0.45cm}
			\includegraphics[width = 0.135\textwidth,height=0.075\textheight]{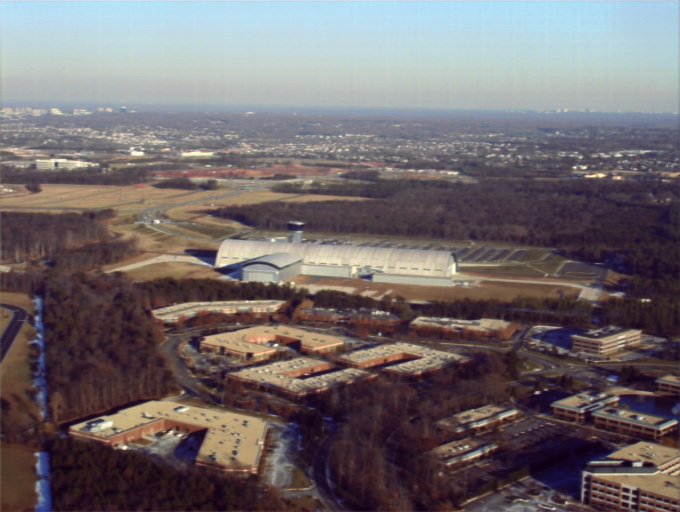} & \hspace{-0.45cm}
			\includegraphics[width = 0.135\textwidth,height=0.075\textheight]{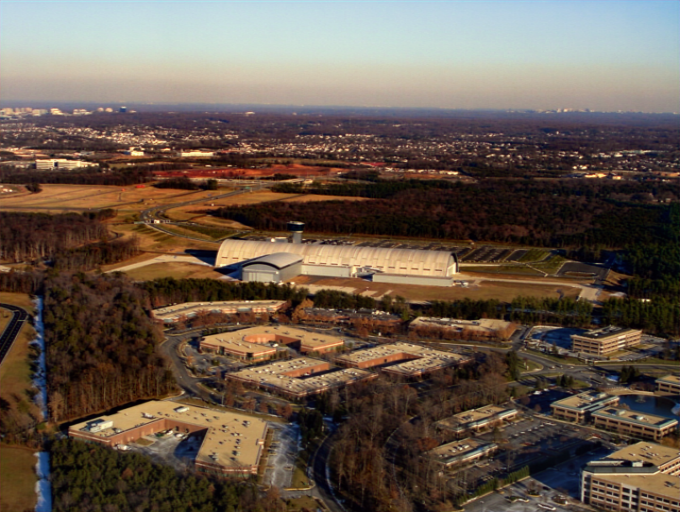} & \hspace{-0.45cm}
			\includegraphics[width = 0.135\textwidth,height=0.075\textheight]{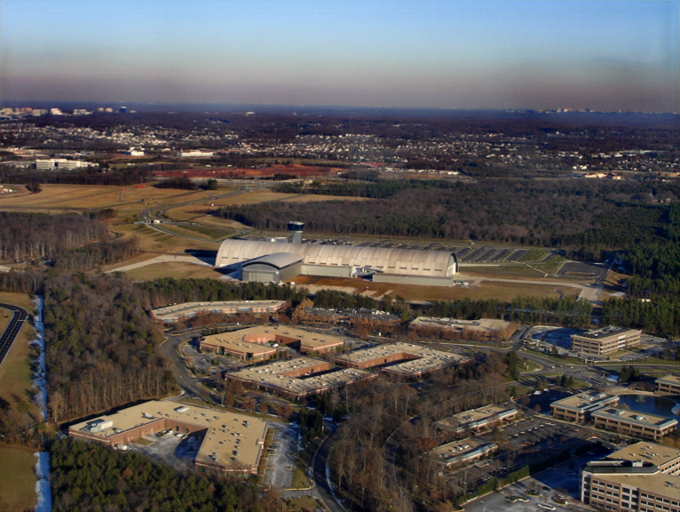} & \hspace{-0.45cm}
			\includegraphics[width = 0.135\textwidth,height=0.075\textheight]{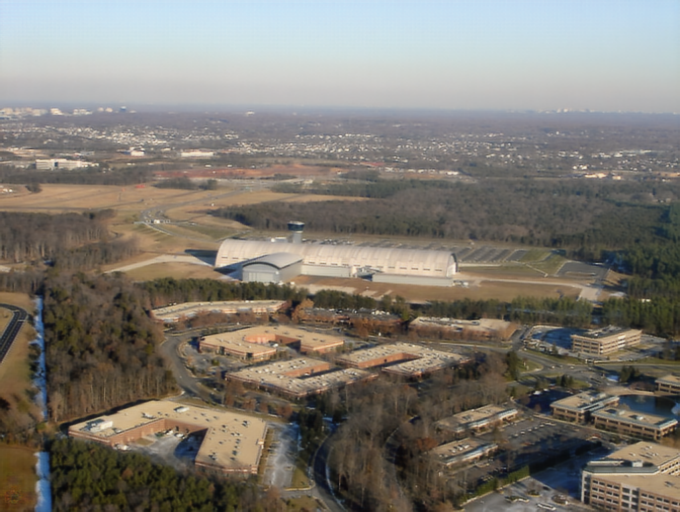} & \hspace{-0.45cm}
			\includegraphics[width = 0.135\textwidth,height=0.075\textheight]{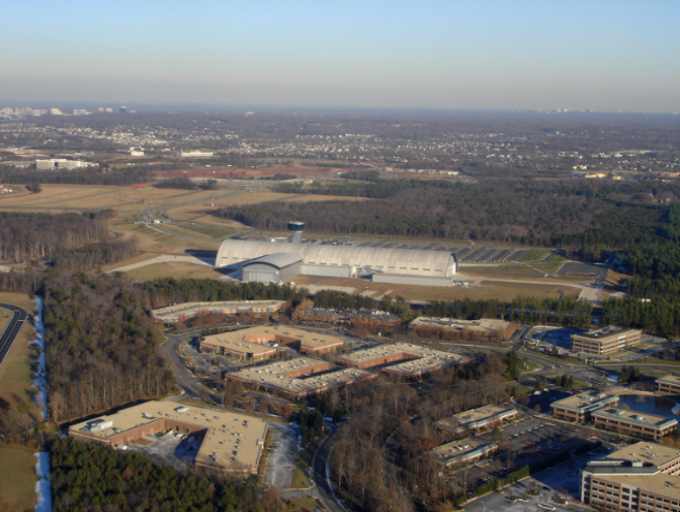} \\
			
			\includegraphics[width = 0.135\textwidth,height=0.06\textheight]{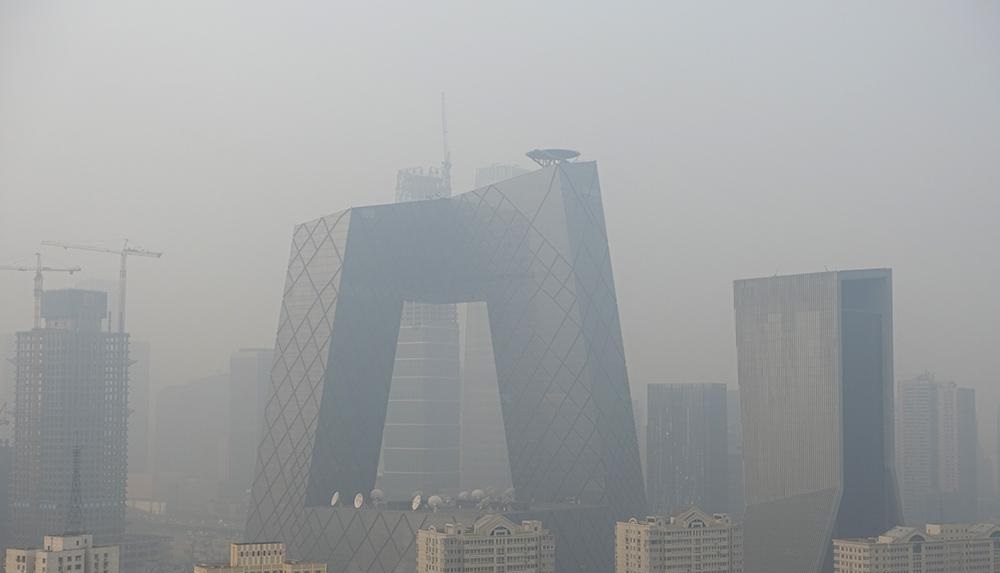} & \hspace{-0.45cm}
			\includegraphics[width = 0.135\textwidth,height=0.06\textheight]{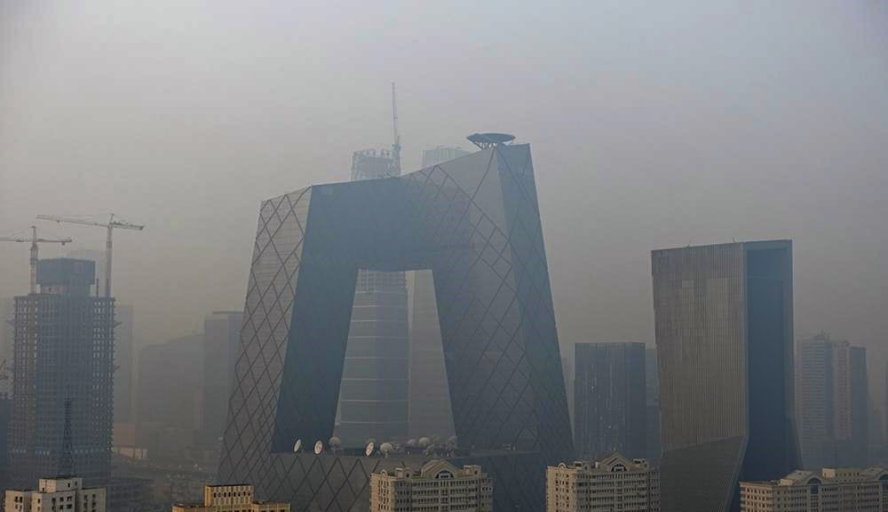} & \hspace{-0.45cm}
			\includegraphics[width = 0.135\textwidth,height=0.06\textheight]{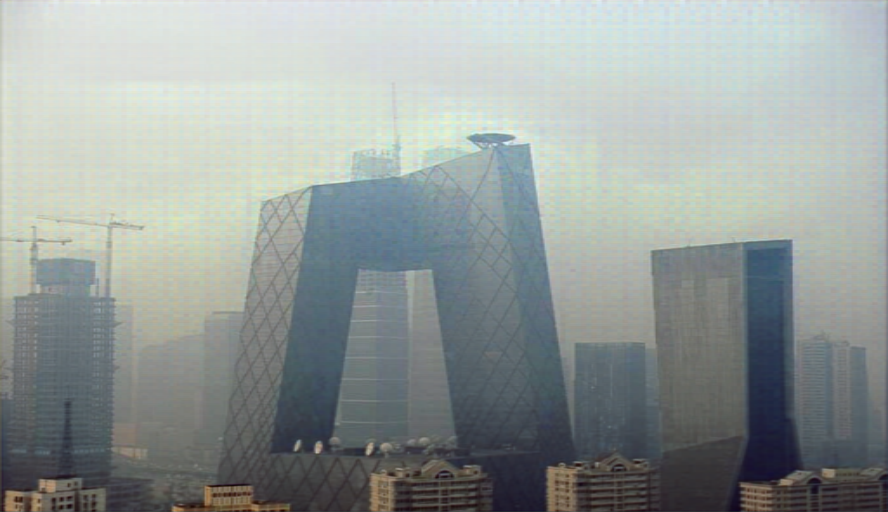} & \hspace{-0.45cm}
			\includegraphics[width = 0.135\textwidth,height=0.06\textheight]{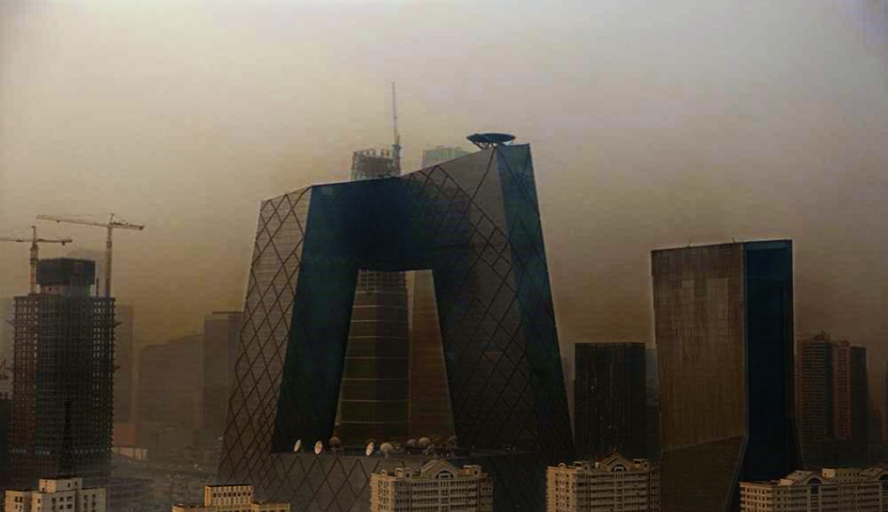} & \hspace{-0.45cm}
			\includegraphics[width = 0.135\textwidth,height=0.06\textheight]{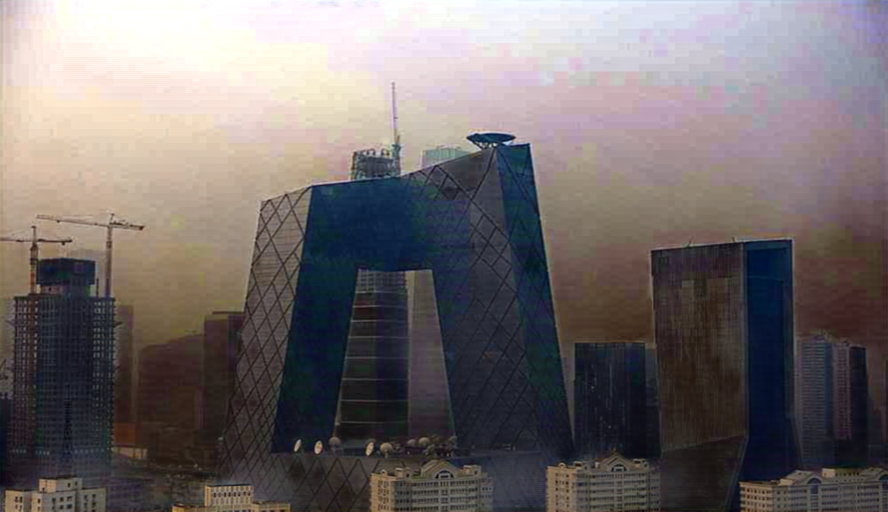} & \hspace{-0.45cm}
			\includegraphics[width = 0.135\textwidth,height=0.06\textheight]{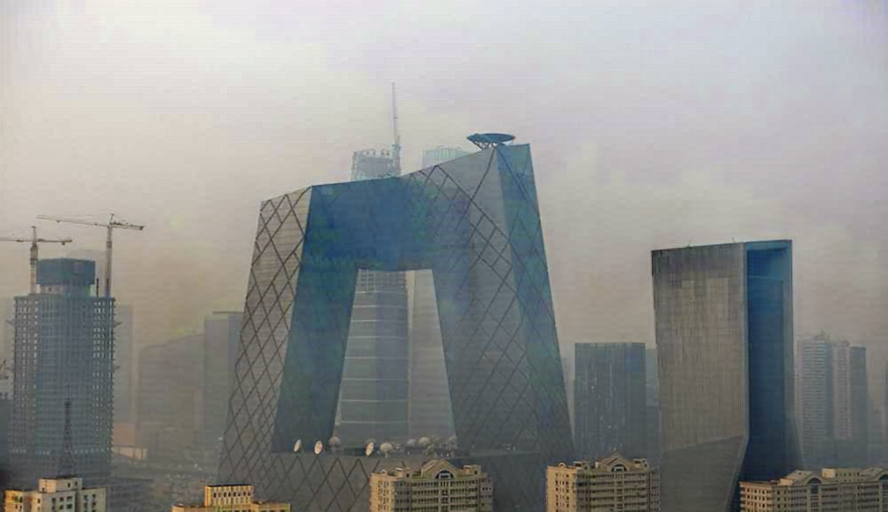} & \hspace{-0.45cm}
			\includegraphics[width = 0.135\textwidth,height=0.06\textheight]{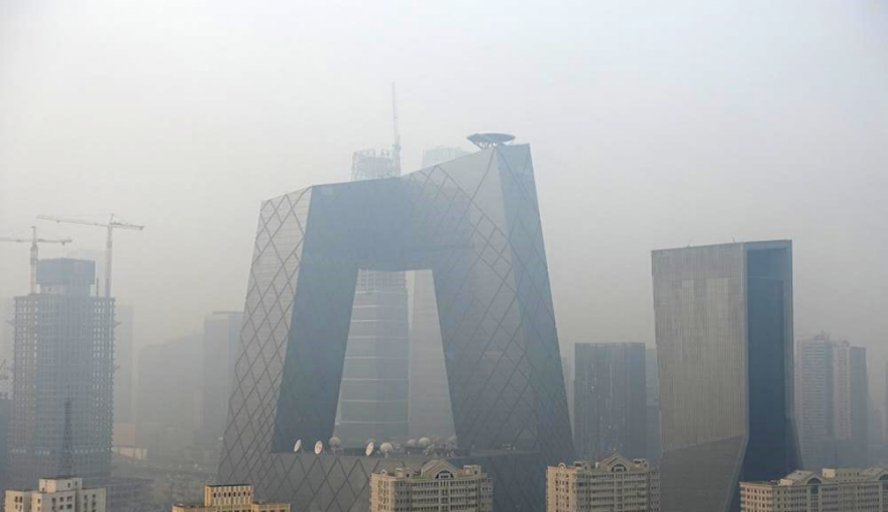} \\
			
			\includegraphics[width = 0.135\textwidth,height=0.075\textheight]{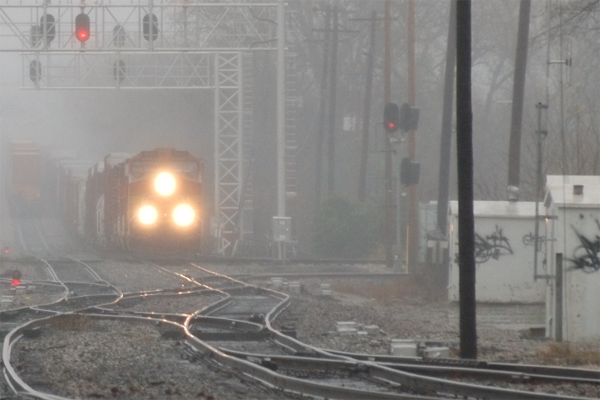} & \hspace{-0.45cm}
			\includegraphics[width = 0.135\textwidth,height=0.075\textheight]{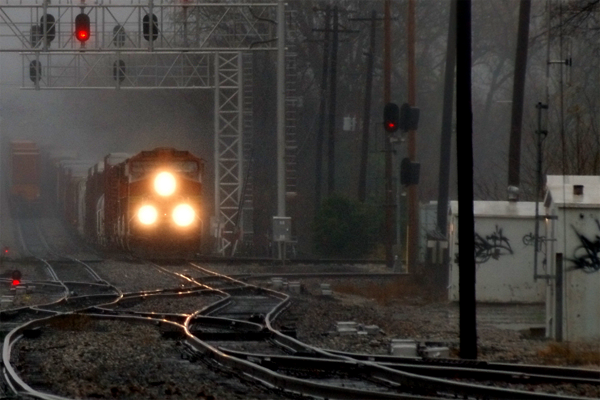} & \hspace{-0.45cm}
			\includegraphics[width = 0.135\textwidth,height=0.075\textheight]{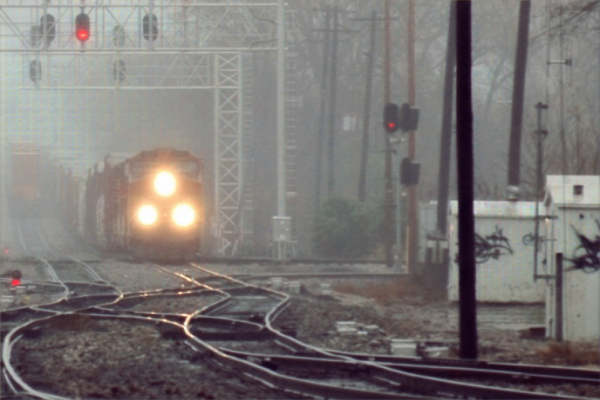} & \hspace{-0.45cm}
			\includegraphics[width = 0.135\textwidth,height=0.075\textheight]{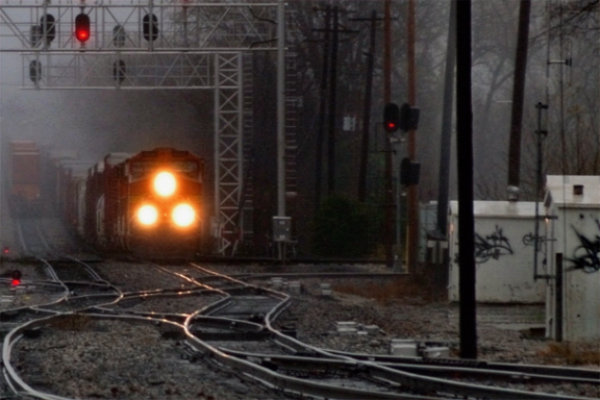} & \hspace{-0.45cm}
			\includegraphics[width = 0.135\textwidth,height=0.075\textheight]{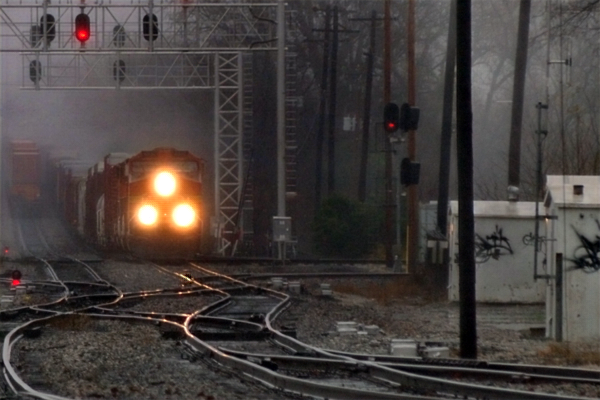} & \hspace{-0.45cm}
			\includegraphics[width = 0.135\textwidth,height=0.075\textheight]{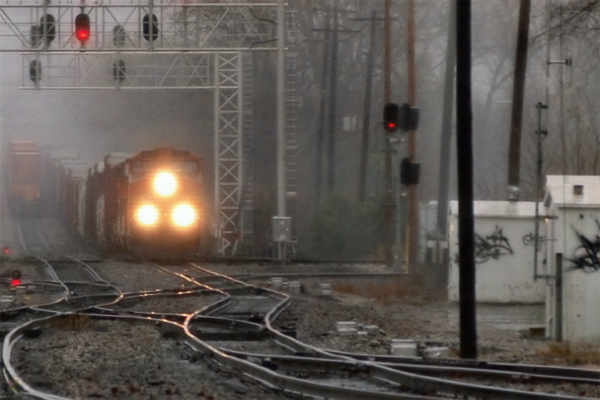} & \hspace{-0.45cm}
			\includegraphics[width = 0.135\textwidth,height=0.075\textheight]{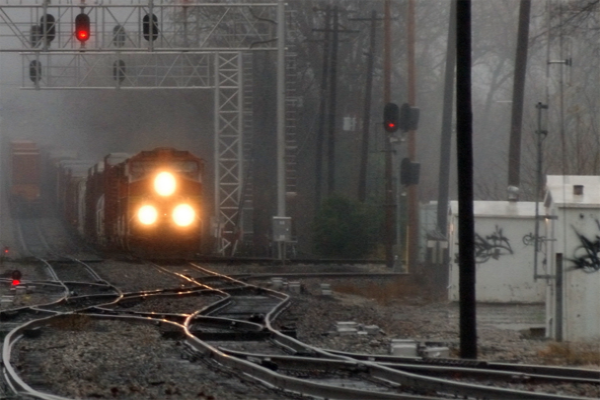} \\
			
			(a) Input & \hspace{-0.45cm}
			(b) DehazeNet \cite{dehazenet2016TIP} & \hspace{-0.45cm}
			(c) DCPDN \cite{zhang2018densely} & \hspace{-0.45cm}
			(d) EPDN \cite{pix2pixdehazing} & \hspace{-0.45cm}
			(e) GCANet \cite{chen2019gated}& \hspace{-0.45cm}
			(f) DA-Dehaze \cite{shao2020domain}& \hspace{-0.45cm}
			(g) MSBDN \cite{dong2020multi}\\

			\includegraphics[width = 0.135\textwidth,height=0.075\textheight]{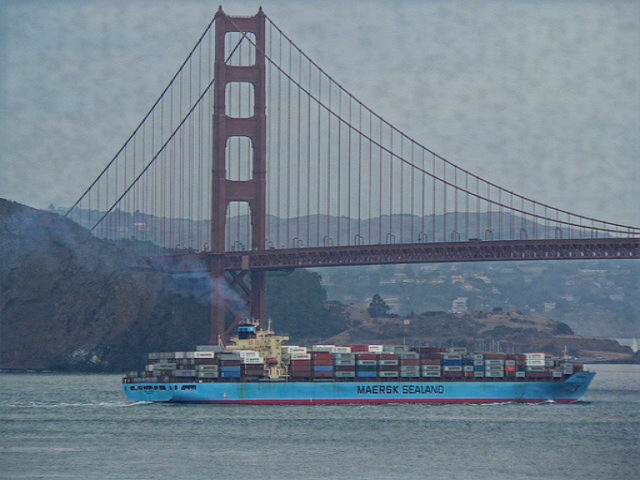} & \hspace{-0.45cm}
			\includegraphics[width = 0.135\textwidth,height=0.075\textheight]{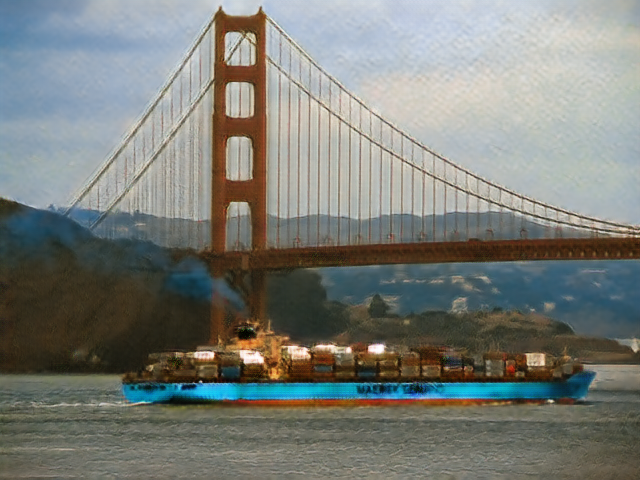} & \hspace{-0.45cm}
			\includegraphics[width = 0.135\textwidth,height=0.075\textheight]{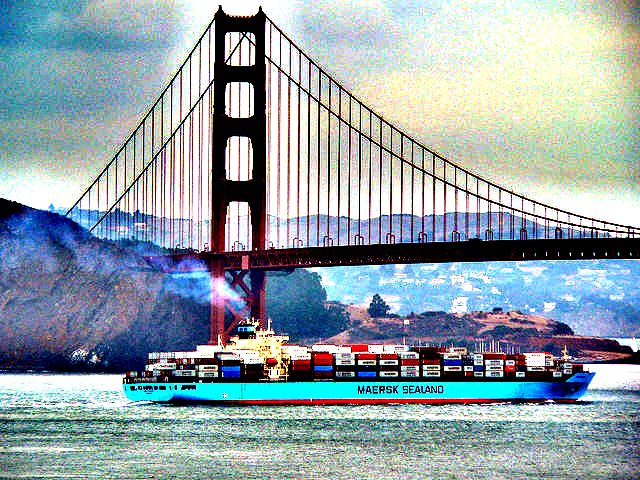} & \hspace{-0.45cm}
			\includegraphics[width = 0.135\textwidth,height=0.075\textheight]{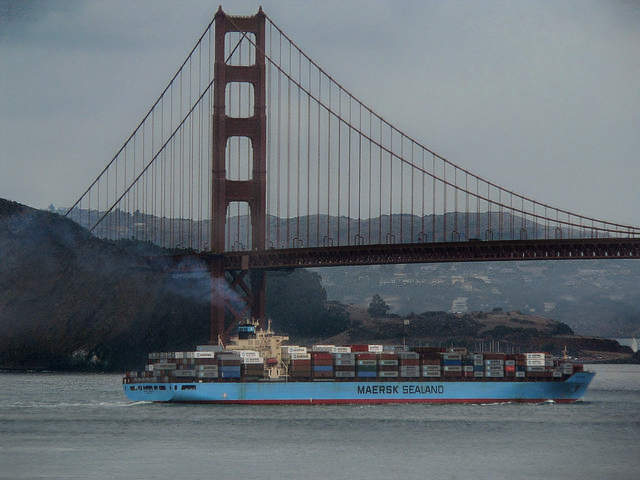} & \hspace{-0.45cm}
			\includegraphics[width = 0.135\textwidth,height=0.075\textheight]{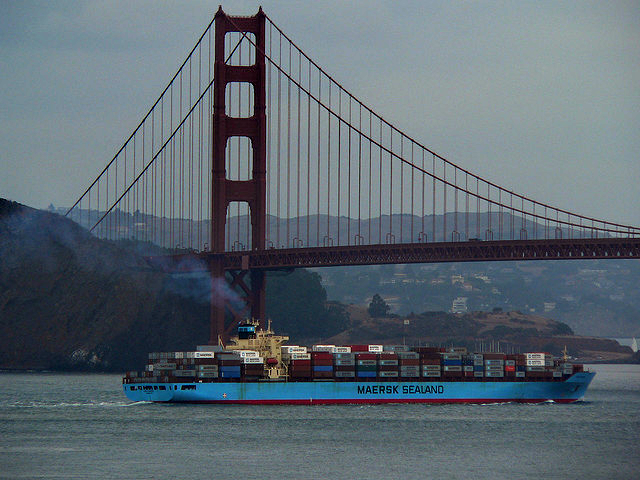} & \hspace{-0.45cm}
			\includegraphics[width = 0.135\textwidth,height=0.075\textheight]{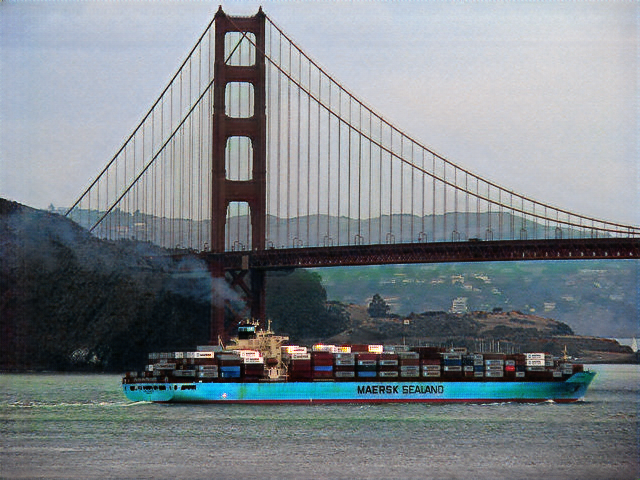} \\
			
			\includegraphics[width = 0.135\textwidth,height=0.075\textheight]{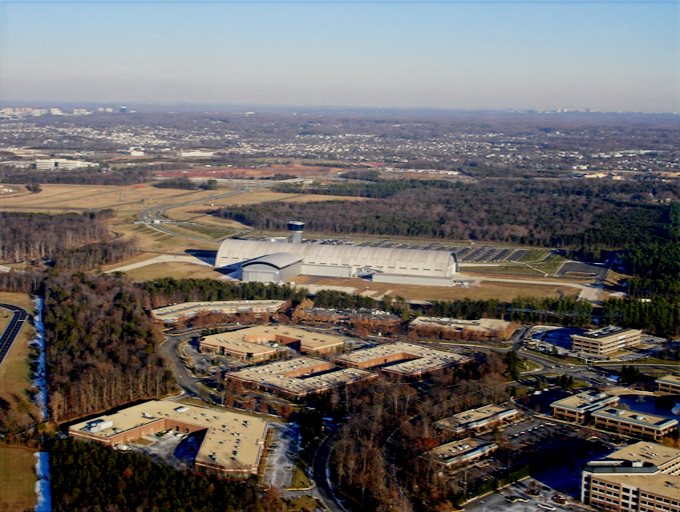} & \hspace{-0.45cm}
			\includegraphics[width = 0.135\textwidth,height=0.075\textheight]{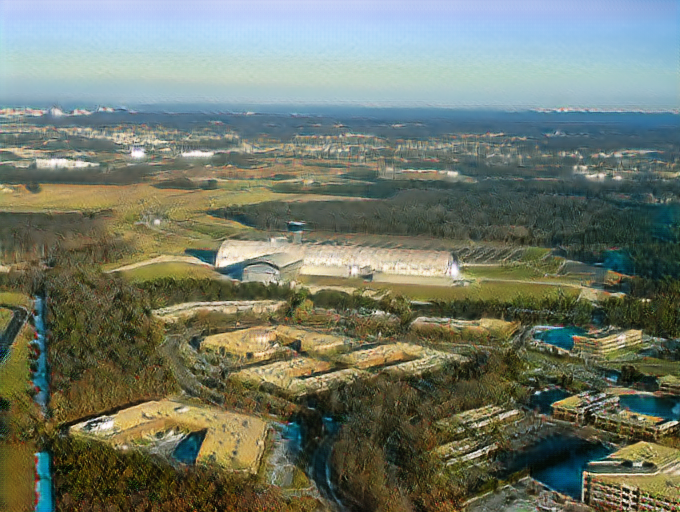} & \hspace{-0.45cm}
			\includegraphics[width = 0.135\textwidth,height=0.075\textheight]{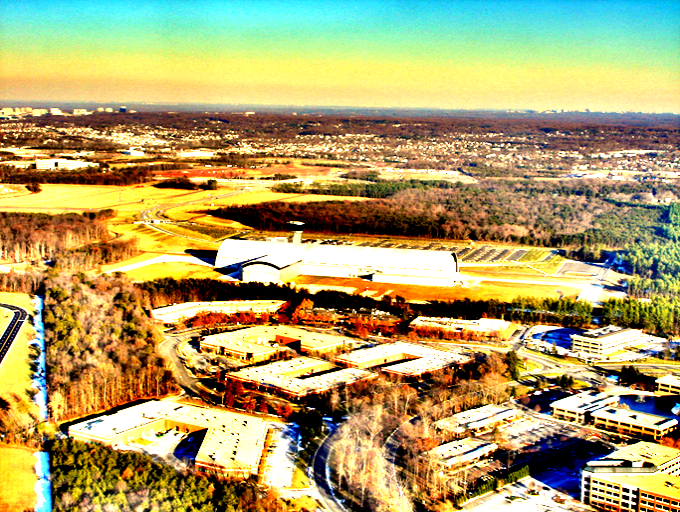} & \hspace{-0.45cm}
			\includegraphics[width = 0.135\textwidth,height=0.075\textheight]{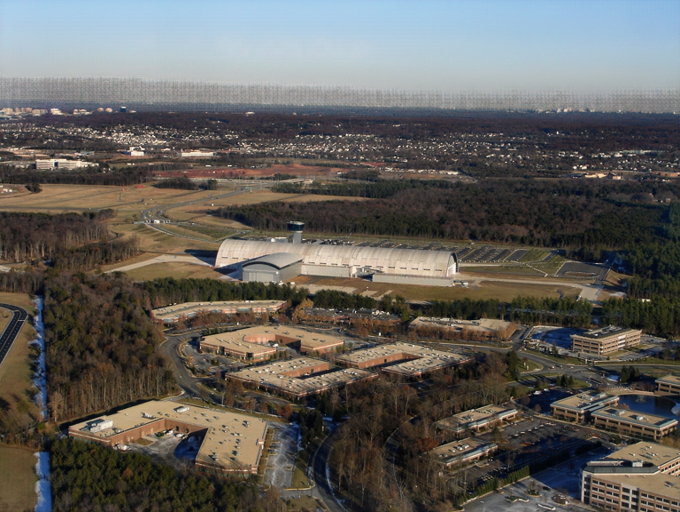} & \hspace{-0.45cm}
			\includegraphics[width = 0.135\textwidth,height=0.075\textheight]{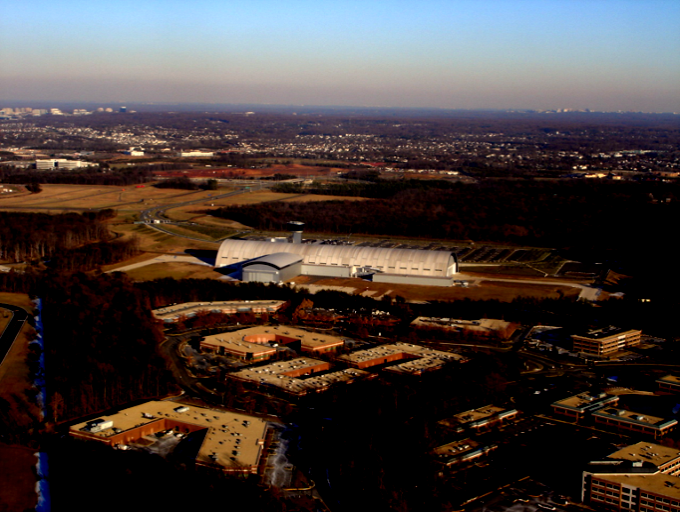} & \hspace{-0.45cm}
			\includegraphics[width = 0.135\textwidth,height=0.075\textheight]{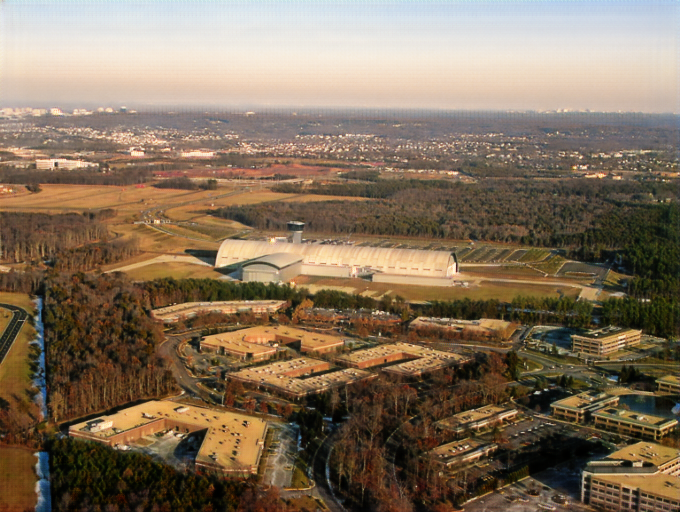} \\
			
			\includegraphics[width = 0.135\textwidth,height=0.06\textheight]{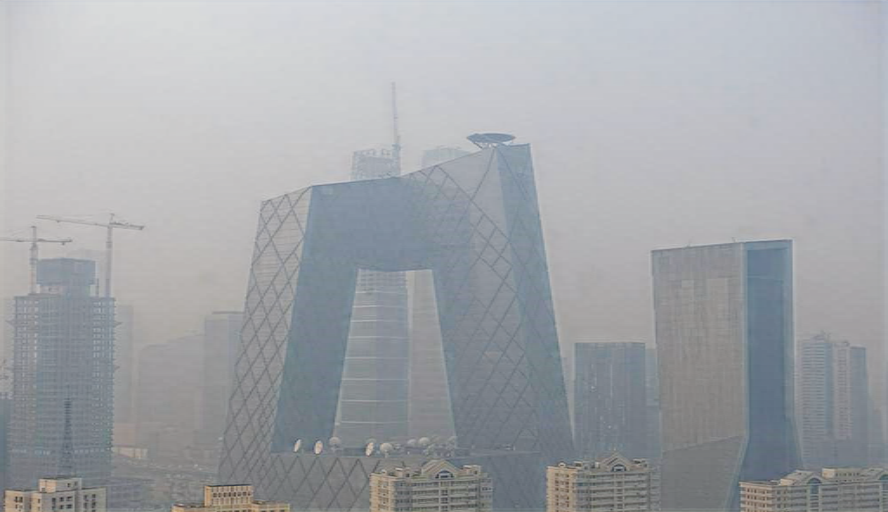} & \hspace{-0.45cm}
			\includegraphics[width = 0.135\textwidth,height=0.06\textheight]{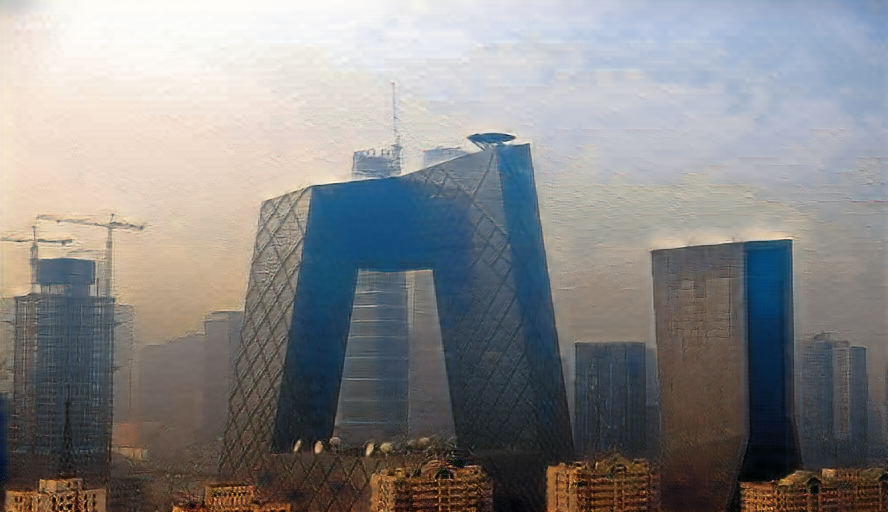} & \hspace{-0.45cm}
			\includegraphics[width = 0.135\textwidth,height=0.06\textheight]{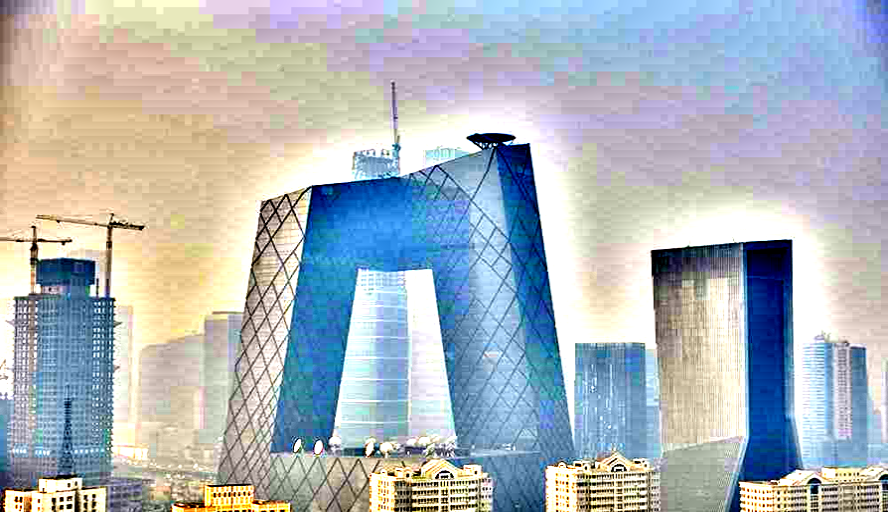} & \hspace{-0.45cm}
			\includegraphics[width = 0.135\textwidth,height=0.06\textheight]{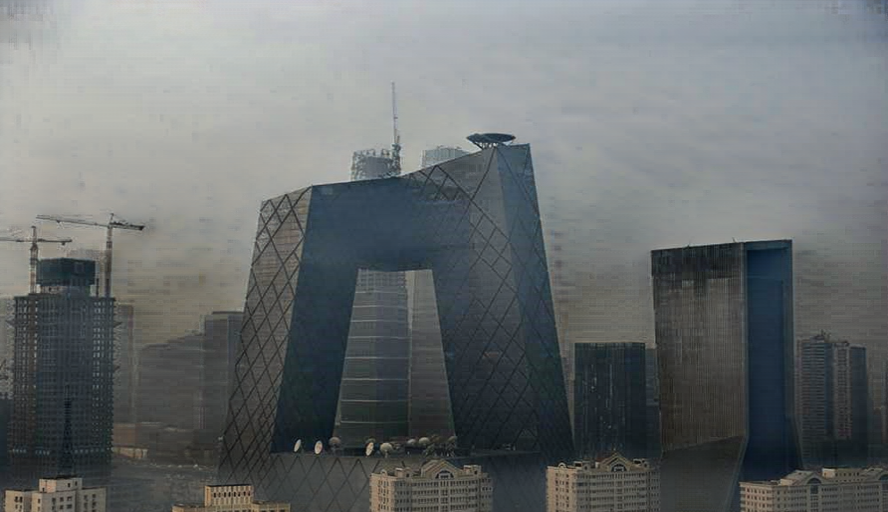} & \hspace{-0.45cm}
			\includegraphics[width = 0.135\textwidth,height=0.06\textheight]{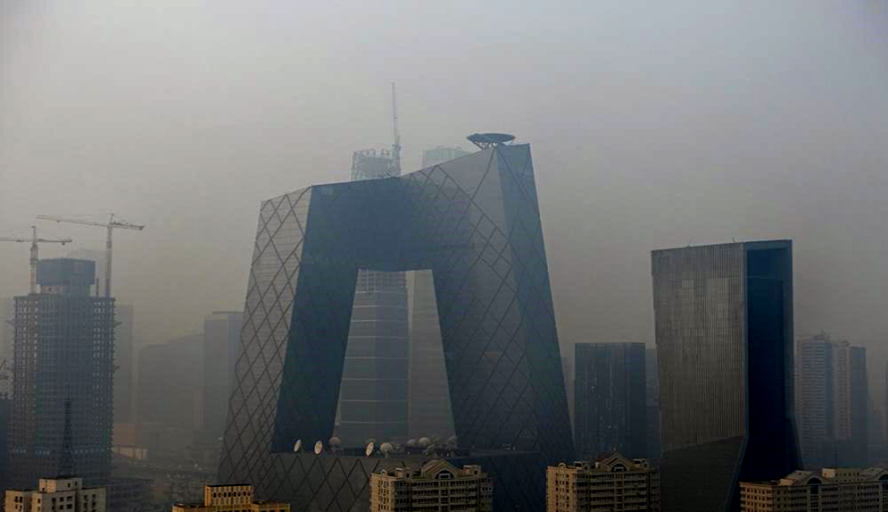} & \hspace{-0.45cm}
			\includegraphics[width = 0.135\textwidth,height=0.06\textheight]{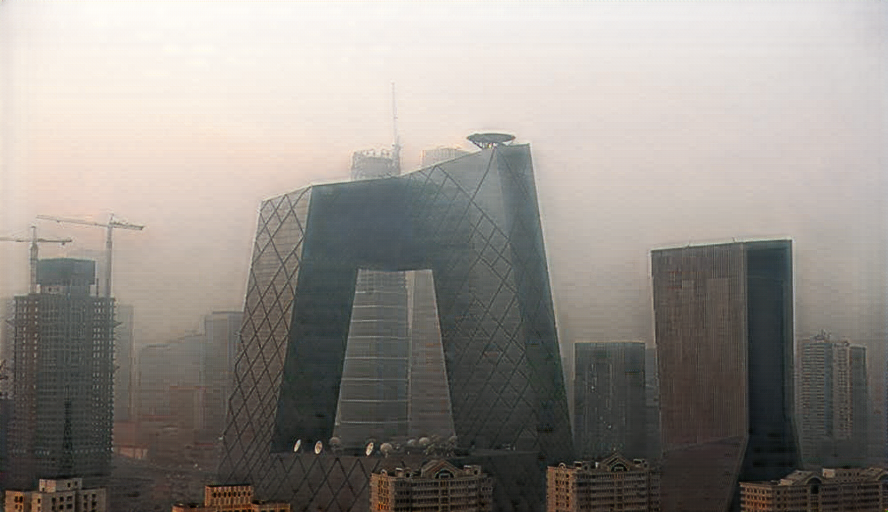}\\
			
			\includegraphics[width = 0.135\textwidth,height=0.075\textheight]{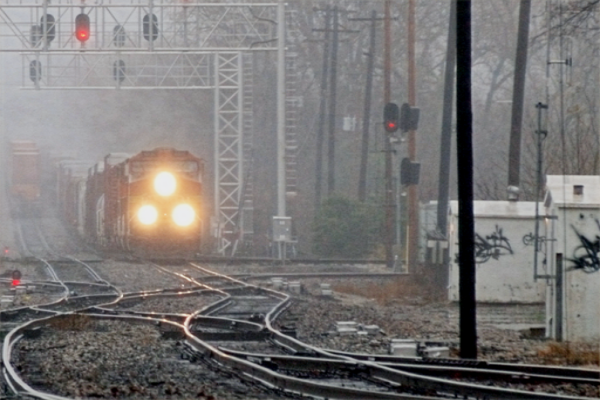} & \hspace{-0.45cm}
			\includegraphics[width = 0.135\textwidth,height=0.075\textheight]{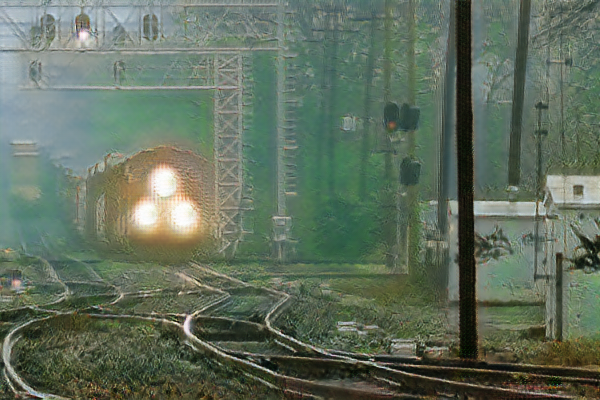} & \hspace{-0.45cm}
			\includegraphics[width = 0.135\textwidth,height=0.075\textheight]{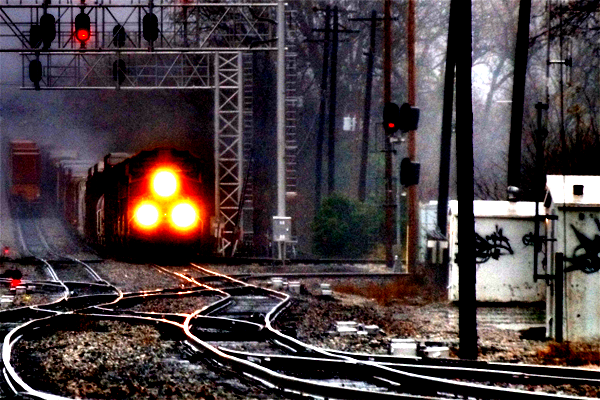} & \hspace{-0.45cm}
			\includegraphics[width = 0.135\textwidth,height=0.075\textheight]{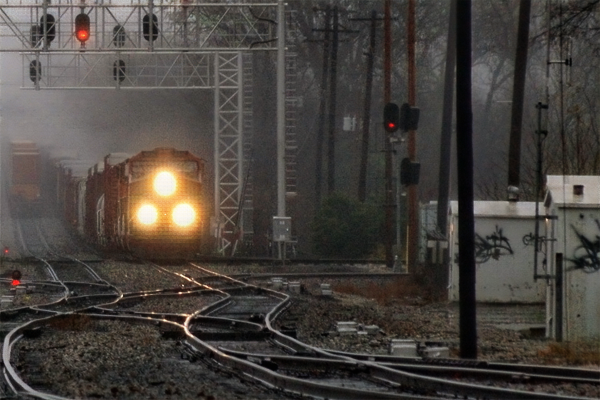} & \hspace{-0.45cm}
			\includegraphics[width = 0.135\textwidth,height=0.075\textheight]{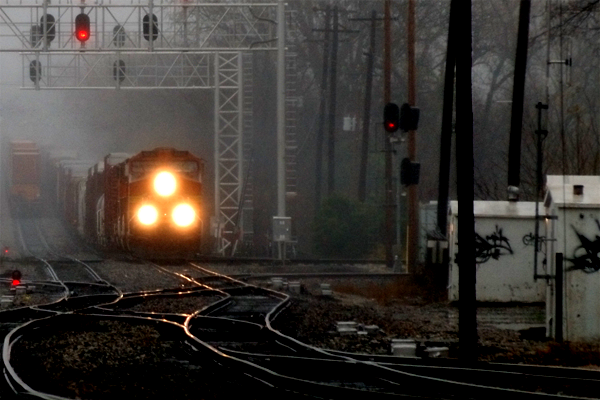} & \hspace{-0.45cm}
			\includegraphics[width = 0.135\textwidth,height=0.075\textheight]{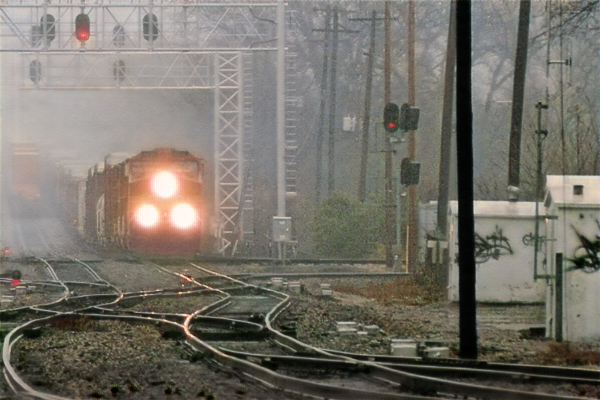}\\
			
			(h) PSD \cite{chen2021psd}& \hspace{-0.45cm}
			(i) Cycle-dehaze \cite{cycle-dehazing}& \hspace{-0.45cm}
			(j) IDE \cite{IDE2021ide}& \hspace{-0.45cm}
			(k) RefineDNet \cite{zhao2021refinednet}& \hspace{-0.45cm}
			(l) $D^{4}$ \cite{yang2022self}& \hspace{-0.45cm}
			(m) QPC-Net
			
		\end{tabular}
	\end{center}
	\vspace{-0.5cm}
	\caption{Qualitative comparison between the proposed QPC-Net and the state-of-the-art methods on the real hazy images.}
	\vspace{-3mm}
	\label{real1_dehaze}
\end{figure*}
Finally, compared to the defogged results generated by our QPC-Net without HAG module (as shown in Fig. \ref{ablation_dehaze}(g)), our QPC-Net's results (as shown in Fig. \ref{ablation_dehaze}(h)) are clearer and more pleasing in visual effect. It demonstrates that HAG can fuse more contextual information from the layers to better recover the color-texture information of the defogged results. Moreover, the above analysis is also reflected in indicator NIQE. It can be seen from Fig. \ref{ablation_dehaze}, the NIQE values of our method are the smallest, followed by w/o VGG, w/o ASM, w/o DC, w/o HAG, Cycle-dehaze \cite{cycle-dehazing} and  CycleGAN \cite{zhu2017unpaired}. This indicates that our method can achieve the best defogging performance. To further demonstrate the effectiveness of our QPC-Net, we quantitative compare on SOTS \cite{RESIDE} outdoor dataset for different configurations. As shown in Table \ref{ablation-tb}, our QPC-Net has the best values of PSNR and SSIM.

\subsection{Performance Comparison with State-of-the-arts}
In this subsection, to verify the effectiveness of our proposed method, we qualitatively and quantitatively compared against the state-of-the-arts methods on both synthetic and real-world datasets.

1) \textbf{\textit{Comparison on synthesized foggy images}}

Fig. \ref{synthesis1_dehaze} and Fig. \ref{synthesis2_dehaze} illustrate the defogging results on SOTS \cite{RESIDE} and HazeRD \cite{HAZERD2017ICIP} by our method and other methods, respectively. In Fig. \ref{synthesis1_dehaze}(b)-(c),(h)-(j),(l), there is a significant amount of fog in the defogged results of DehazeNet \cite{dehazenet2016TIP}, DCPDN \cite{zhang2018densely}, PSD \cite{chen2021psd}, Cycle-dehaze \cite{cycle-dehazing}, IDE \cite{IDE2021ide} and $D^4$  \cite{yang2022self}. As can be seen in Fig. \ref{synthesis1_dehaze}(h)-(j), although PSD \cite{chen2021psd}, Cycle-dehaze \cite{cycle-dehazing} and IDE \cite{IDE2021ide} are able to improve image brightness, some results appeared varying degrees of discoloration and over-enhancement. Furthermore, the defogged results generated by DCPDN \cite{zhang2018densely} (as shown in the second row of Fig. \ref{synthesis2_dehaze}(c)) have completely lost most of the structural information when processing images with dense fog. We can observe that the defogged results of EPDN \cite{pix2pixdehazing}, GCANet \cite{chen2019gated}, DA-Dehaze \cite{shao2020domain} and RefineDNet \cite{zhao2021refinednet} (as shown in Fig. \ref{synthesis1_dehaze}(d)-(f),(k) and Fig. \ref{synthesis2_dehaze}(d)-(f),(k)) exhibit low brightness and artifacts, which is caused by excessive defogging. For the defogged results in Fig. \ref{synthesis1_dehaze}(g) and Fig. \ref{synthesis1_dehaze}(m), both MSBDN \cite{dong2020multi} and our QPC-Net can thoroughly remove the fog while retaining a clear texture. However, the defogged results of MSBDN \cite{dong2020multi} (as shown in \ref{synthesis2_dehaze}(g)) appear over-enhancement in the sky region. In comparison, our model can remain sharper edge contours and the results (as shown in Fig. \ref{synthesis2_dehaze}(m)) are closer to the ground truth images visually.

To quantitatively compare the recovery quality of the different techniques, Table \ref{synthesis-tb} summarizes the mean value of the seven metrics on the outdoor datasets of SOTS \cite{RESIDE} and HazeRD \cite{HAZERD2017ICIP}. Although Our method is trained unpaired, it still outperforms the competitors including supervised ones, in terms of most reference and non-reference metrics. Especially, on both synthetic datasets, the proposed method achieves the best values in terms of PSNR, SSIM and BRISQUE, which demonstrates that the superiority of our method in terms of color recovery and generation quality of defogging results.
\begin{table*}[htp]
	\renewcommand\arraystretch{1.2}
	\begin{center}
		\captionsetup{justification=centering}
		\caption{\\A{\footnotesize VERAGE} PSNR, SSIM, LPIPS, FADE, BRISQUE, NIQE, $\overline{\gamma}$ {\footnotesize OF} D{\footnotesize EFOGGED} R{\footnotesize ESULTS} {\footnotesize ON} R{\footnotesize EAL-WORLD} D{\footnotesize ATASETS} O-HAZE {\footnotesize AND} LIVE. T{\footnotesize HE} T{\footnotesize OP} T{\footnotesize WO} P{\footnotesize ERFORMANCE} V{\footnotesize ALUES} A{\footnotesize RE} H{\footnotesize IGHLIGHTED} {\footnotesize IN} R{\footnotesize ED} {\footnotesize AND} B{\footnotesize LUE}}
		\resizebox{\textwidth}{2.3cm}{
			\begin{tabular}{c|c|ccccccc|ccccccc}
				\hline
				&Dataset & \multicolumn{7}{c|}{O-HAZE \cite{O-HAZE}} & \multicolumn{7}{c}{LIVE \cite{referencelessdefogging2015TIP}}\\
				
				\cline{2-16}
				&\diagbox{Methods}{Metric} & PSNR \cite{huynh2008scope} & SSIM \cite{wang2004image} & LPIPS \cite{zhang2018unreasonable}  & FADE \cite{referencelessdefogging2015TIP} & BRISQUE \cite{MSCN2012TIP} & NIQE \cite{mittal2012making} & $\overline{\gamma}$ \cite{hautiere2011blind} &  PSNR \cite{huynh2008scope} & SSIM \cite{wang2004image} & LPIPS \cite{zhang2018unreasonable}  & FADE \cite{referencelessdefogging2015TIP} & BRISQUE \cite{MSCN2012TIP} & NIQE \cite{mittal2012making} & $\overline{\gamma}$ \cite{hautiere2011blind} \\
				
				\hline
				\multirow{6}{*}{\rotatebox{90}{Paired}}
				&DehazeNet \cite{dehazenet2016TIP} & 18.5655  & 0.7304 & 0.2566 & 0.8119 & 11.6304 & 2.6174 & 1.2240 &  -  & - & - & 0.7587 & 16.8184 & 4.1288 & 1.0860\\
				
				&DCPDN \cite{zhang2018densely} &  18.6372 & 0.7268 & 0.2754 & 0.7051 & 21.1391 & 3.4793 & 1.4379 &  - & - & - & 1.1490 & 19.3902 & 4.3371 & 1.0150\\
				
				&EPDN \cite{pix2pixdehazing} &  19.0154 & 0.7822 & 0.1989 & 0.3505 & 18.1324 & 3.1017 &  1.7658 & - & - & - & 0.6107 & 9.7798 & 4.1584 & 1.4640\\

				&GCANet \cite{chen2019gated} &  17.9407 & 0.7443 & 0.1917 & \textcolor[rgb]{0,0,1}{0.2714}  & 21.7128 & 2.8813 & 1.9534 & - & - & - & 0.6791 & 10.5648 & 3.8063 & 1.4816\\
				
				&DA-Dehaze \cite{shao2020domain} &  \textcolor{blue}{19.5841} & 0.7279 & 0.2505 & 0.3527 & 15.9735 & 2.7854 & 2.1407  & - & - & - & 0.6173 & 12.6701 & 3.8717 &  1.3834\\
				
				&MSBDN \cite{dong2020multi} &  15.3640 & 0.5733 & 0.2656 & 0.5195 & 18.2713 & 4.7255 & 1.5547  & - & - & - & 1.0433 & 9.4844 & 4.1451 & 1.2242\\
				
				&PSD \cite{chen2021psd}&  13.4581 & 0.5520 & 0.2523  & 0.6392 & 14.5644 & 2.4693 & 3.222  & - & - & - & 0.8522 & 25.4683 & 4.1109  & 2.0147\\
				
				\hline
				\multirow{5}{*}{\rotatebox{90}{W/o Paired}}
				&Cycle-dehaze \cite{cycle-dehazing} &  13.8378 & 0.4867 & 0.3435 & 0.3816 & 12.9541 & 2.7142 & 2.7609  & - & - & - & 0.9243 & 17.0631 & 3.7915 & 1.4897 \\
				
				&IDE \cite{IDE2021ide} &  17.2068 & \textcolor[rgb]{0,0,1}{0.7997} & 0.2446  & 0.3261 & 14.6594 & 2.3629 & \textcolor[rgb]{1,0,0}{3.6548} &  - & - & - & \textcolor[rgb]{0,0,1}{0.5632} & 21.4205 & 3.7041 & \textcolor[rgb]{1,0,0}{2.5919}\\

				&RefineDNet \cite{zhao2021refinednet} &  16.7359 & 0.5844 & \textcolor{red}{0.1903} & 0.3906 & \textcolor{red}{9.7552} & \textcolor{blue}{2.2746} & 1.8385 & - & - & - & 0.6287 & \textcolor{blue}{9.4159} & \textcolor{red}{3.4355} & 1.4865 \\
				
				&$D^{4}$ \cite{yang2022self} &  14.2810 & 0.4457 & 0.2838 & 0.4085 & 19.4519 & 2.6633 & 1.4652 & - & - & - & 0.5454 & 35.9168 & 5.1459 & 1.1705 \\
				
				&QPC-Net &  \textcolor[rgb]{1,0,0}{20.2749} & \textcolor[rgb]{1,0,0}{0.8009} & \textcolor{blue}{0.1914} & \textcolor{red}{0.2677} & \textcolor[rgb]{0,0,1}{11.0563} & \textcolor{red}{2.2636} & \textcolor[rgb]{0,0,1}{3.2397}  &  - & - & - & \textcolor[rgb]{1,0,0}{0.5156} & \textcolor[rgb]{1,0,0}{9.2760} & \textcolor[rgb]{0,0,1}{3.5043} & \textcolor[rgb]{0,0,1}{2.5796}\\
				
				\hline
		\end{tabular}}
		\label{real-tb}
		\vspace{-0.8cm}
	\end{center}
\end{table*}

2)  \textbf{\textit{Comparison on real-world foggy images}}

We evaluate our method on real foggy images, which are provided by the LIVE \cite{referencelessdefogging2015TIP} and previous methods. Fig. \ref{real1_dehaze} shows the defogged results of qualitative comparison between our QPC-Net and the other methods on the real foggy images. As can be seen in Fig. \ref{real1_dehaze}(b),(h), although DehazeNet \cite{dehazenet2016TIP} and PSD \cite{chen2021psd} are able to remove fog from misty images, the defogged results still have some remaining fog artifacts. As shown in Fig. \ref{real1_dehaze}(c), most of the fog is unremoved in the DCPDN's results, and some results appeared serious checkerboard effect. The results of EPDN \cite{pix2pixdehazing} and GCANet \cite{chen2019gated} have no remaining fog, however, some defogged images significantly suffer from over-dehaze resulting in regional darkening (as shown in the last row of Fig. \ref{real1_dehaze}(d) and Fig. \ref{real1_dehaze}(e)). The defogged results of RefineDNet \cite{zhao2021refinednet} and $D^4$  \cite{yang2022self} (as shown in Fig. \ref{real1_dehaze}(k)-(l)) also showed low contrast. As can be seen in Fig. \ref{real1_dehaze}(f)-(g), there is some fog remain in the defogged results of DA-Dehaze \cite{shao2020domain} and MSBDN \cite{dong2020multi}. The defogged images generated by Cycle-dehaze \cite{cycle-dehazing} are distorted (as shown in Fig. \ref{real1_dehaze}(i)). Similar to the defogged results from synthetic images, as shown in Fig. \ref{real1_dehaze}(j), the color distortion phenomenon still exists in the results by using IDE \cite{IDE2021ide} to process the real-world foggy images. In contrast, the results of our method QPC-Net (as shown in Fig. \ref{real1_dehaze}(m)) are more natural and clearer than other methods'. The reason is that our method uses unpaired real-world foggy dataset for training, and the proposed attention mechanism network can pay more attention to fog relevant features to preserve richer image content and texture details for defogged results. We performed the quantitative comparison of defogged results on the two real-world datasets of O-HAZE \cite{O-HAZE} and LIVE \cite{referencelessdefogging2015TIP}. As shown in Table \ref{real-tb}, it is clear that QPC-Net outperforms state-of-the-art techniques in terms of PSNR and SSIM, which proves our method has quite competitive in single image defogging. Our method has higher $\overline{\gamma}$ than others, which indicates that our method well restores texture details. Meanwhile, the BRISQUE scores and NIQE values of our results are lower than other methods, which means that our defogged results have higher fidelity. Moreover, our method has a smaller FADE than others, which demonstrates that our defogged results are clearer.

\section{Concluding Remarks}
\label{concluding}
In this work, we presented QPC-Net to exploit the multiple constraint criterion to improve the quality of defogged images based on a cyclic generative adversarial framework.We devised three basic modules (FRM, FSM, and CTRM) to constrain each other to generate high-quality images from fog/fogfree domain to fogfree/fog domain by using quad mapping paths. We also explored several derived images of foggy inputs to guide our channel-spatial attention to cooperatively retrieve more foggy related contextual information to enhance details in defogged images. Finally, we used a novel sky segmentation network to optimize the atmospheric light to guide the FSM to generate more natural foggy images to improve the transformation mapping ability from foggy to fog-free images by FRM. Extensive qualitative and quantitative comparisons demonstrate that QPC-Net delivers superior defogging performance over the state of the art.





\bibliographystyle{elsarticle-num}
\bibliography{UQCA_JabRef}







\end{document}